\documentclass[smallextended,final]{svjour3}
\makeatletter
\usepackage[12pt]{extsizes}
\newcommand\subparagraph{%
	\@startsection{subparagraph}{5}
	{\parindent}
	{3.25ex \@plus 1ex \@minus .2ex}
	{-1em}
	{\normalfont\normalsize\bfseries}}
\makeatother
\let\subparagraph\relax
\usepackage{amsmath,amssymb,amsbsy,amsfonts,latexsym,amsopn,amstext,amsxtra,euscript,amscd,hyperref}
\usepackage{mathptmx}      
\usepackage{graphicx}
\usepackage{algorithm}
\usepackage{algorithmic}
\usepackage[numbers]{natbib}
\usepackage{notoccite}
\usepackage{graphics,graphicx}
\usepackage{epsfig,epstopdf}
\usepackage{array,tikz}
\usepackage{caption}
\usepackage{slashbox}
\usepackage{wrapfig}
\usepackage{setspace}
\usepackage{titlesec}
\newcolumntype{C}{@{}c@{}}

\renewcommand{\arraystretch}{1.25}
\usepackage{float}
\usepackage[top=1.5cm, left=1.5cm,right=1.6cm,bottom=1.28cm]{geometry}
\newcounter{myeqno}
\usepackage{chngcntr}

\usepackage{tikz}
\usetikzlibrary{shapes.geometric,arrows}
\usetikzlibrary{shapes.geometric,arrows,fit,matrix,positioning,shapes.multipart}
\tikzstyle{startstop}=[rectangle, rounded corners, minimum width=3cm, minimum height=1cm, draw=black]
\tikzstyle{startstop1}=[rectangle, rounded corners, minimum width=8cm, minimum height=4cm, draw=black]
\tikzstyle{startstop2}=[square, rounded corners, minimum width=0.5cm, minimum height=1cm, draw=black]
\tikzstyle{startstop3}=[square, rounded corners, minimum width=5cm, minimum height=1cm, draw=black]
\tikzstyle{startstop4}=[square, rounded corners, minimum width=1cm, minimum height=5cm, draw=black]
\tikzstyle{round} = [ellipse, minimum width=3cm, minimum height=1cm, draw= black]
\tikzstyle{startstop2}=[rectangle, rounded corners, minimum width=2cm, minimum height=1cm, draw=black]
\tikzstyle{arrow} =[draw, -latex']
\renewcommand{\arraystretch}{1.5}
\usepackage{longtable,lscape}
\usepackage{rotating}
\usepackage{multirow,multicol}
\usepackage{enumerate}
\usepackage{subcaption}
\usepackage{makecell}

\usepackage{stfloats}

\usepackage{pdflscape}
\usepackage{xcoffins}
\NewCoffin\tablecoffin

\usepackage{color,soul}
\usepackage{xcolor}
\usepackage{colortbl}
\definecolor{mygrey}{RGB}{220,220,220}
\graphicspath{{images/}}
\usepackage{amsmath}
\usepackage{multirow}
\usetikzlibrary{arrows.meta, decorations.pathreplacing}
\begin{document}
    \title{Learning Strategies in Particle Swarm Optimizer: A Critical Review and Performance Analysis}
	 \author{Dikshit Chauhan$^{*,1}$ \and Shivani$^{*,2}$ \and P. N. Suganthan$^{3}$}
	\thanks{corresponding author}
	\institute{$^*$Corresponding author\\
    $^1$Department of Electrical and Computer Engineering, National University of Singapore, 119077, Singapore\\
   $^2$Department of Mathematics and Computing, Dr. B.R. Ambedkar National Institute of Technology Jalandhar, Jalandhar-144008, Punjab, India.\\
   $^3$KINDI Computing Research Center, College of Engineering, Qatar University, Doha, Qatar\\
\email{dikshitchauhan608@gmail.com (Dikshit Chauhan), sainishivani2310@gmail.com (Shivani), p.n.suganthan@qu.edu.qa (P. N. Suganthan)}
    }
	\maketitle
	\begin{abstract}
Nature has long inspired the development of swarm intelligence (SI), a key branch of artificial intelligence that models collective behaviors observed in biological systems for solving complex optimization problems. Particle swarm optimization (PSO) is widely adopted among SI algorithms due to its simplicity and efficiency. Despite numerous learning strategies proposed to enhance PSO's performance in terms of convergence speed, robustness, and adaptability, no comprehensive and systematic analysis of these strategies exists. We review and classify various learning strategies to address this gap, assessing their impact on optimization performance. Additionally, a comparative experimental evaluation is conducted to examine how these strategies influence PSO's search dynamics. Finally, we discuss open challenges and future directions, emphasizing the need for self-adaptive, intelligent PSO variants capable of addressing increasingly complex real-world problems.


	\end{abstract}
	\keywords{Optimization, Evolutionary computation, Particle swarm optimizer, Learning strategies, Analysis
		 }







\section{Introduction}\label{Sec: Introduction}
Over the past few decades, optimization problems have become increasingly prevalent across various industrial and scientific domains, including mechanical engineering \cite{wang2023new}, nuclear energy \cite{yang2024ua}, vehicle engineering \cite{wu2024neighborhood}, reliability analysis \cite{meng2021comparative}, aerospace engineering \cite{meng2024optimum}, and topology optimization \cite{xing2023topology} The growing complexity of real-world problems, such as nonlinear optimal control, text clustering, DNA sequence compression, and distribution network design, has intensified the demand for efficient optimization algorithms. As a result, research on optimization techniques holds significant practical value and broad applicability \cite{zhang2017vector}.

Optimization problems are typically categorized into constrained and unconstrained problems. A constrained optimization problem can be expressed as: $\min \mathbf{F} = f(\mathbf{X}),~\mathbf{X} \in \mathbf{\mathbb{S}},~ \mathbb{S} = \{\mathbf{X} \mid \mathbf{g}_i(\mathbf{X}) \leq 0,~ i = 1, \dots, m_c\},$ where \( \mathbf{F} = f(\mathbf{X}) \) is the objective function, \( \mathbf{g}_i(\mathbf{X}) \) represents the constraint functions, \( m \) is the number of constraints, and \( \mathbf{X} \) is a \( D \)-dimensional vector. In contrast, unconstrained optimization problems, formulated in Euclidean \( n \)-space, take the form: $\min \mathbf{F} = f(\mathbf{X}),~ \mathbf{X} \in \mathbb{S},~ \mathbf{X} = \{x^1, x^2, \dots, x^D\}, \quad \mathbb{S} \subset \mathbb{R}^n$. Efficient optimization methods are essential for solving both types, leading to improved solutions for complex real-world challenges. These methods can be broadly divided into gradient-based techniques and metaheuristic algorithms. While gradient-based methods rely on mathematical derivatives, metaheuristic algorithms are favored for their flexibility, derivative-free nature, and ease of implementation \cite{zhao2025new, sorensen2015metaheuristics}. Among them, particle swarm optimizer (PSO) \cite{kennedy1995particle} has gained considerable attention due to its simplicity and rapid convergence.

PSO is a global stochastic optimization algorithm inspired by the collective behavior of bird flocks and biological swarms. Each particle in the swarm represents a potential solution, adjusting its position based on both personal experience and the knowledge of neighboring particles. This cooperative mechanism enables efficient search space exploration and convergence toward optimal solutions \cite{houssein2021major}.

One of PSO's primary advantages is its simplicity and ease of implementation. With a straightforward mathematical structure and minimal parameter tuning, it is accessible for a wide range of optimization problems. Unlike gradient-based methods, PSO does not require derivative information, making it well-suited for non-differentiable, multimodal, and complex optimization landscapes \cite{zeng2014novel,shami2022particle}. Additionally, PSO exhibits fast convergence, particularly in lower-dimensional problems, where particles quickly identify near-optimal solutions through collaborative learning \cite{gad2022particle,chauhan2024competitive}.

Despite its strengths, PSO faces several challenges, particularly in complex and high-dimensional optimization problems. One of the most significant limitations is premature convergence, where particles become trapped in local optima, leading to suboptimal solutions \cite{eslami2012survey}. This issue is particularly pronounced in multimodal optimization landscapes, where the algorithm may struggle to escape local attraction basins. Moreover, as the number of dimensions increases, PSO's exploration and exploitation capabilities decline, reducing its efficiency, an effect commonly referred to as the curse of dimensionality \cite{houssein2021major,chauhan2024competitive}.

Another limitation of PSO is its sensitivity to parameter settings. The performance of the algorithm heavily depends on selecting appropriate values for key parameters such as inertia weight, cognitive and social coefficients, and velocity limits \cite{nayak202325,li2022ranking}. Improper tuning of these parameters can lead to inefficient exploration, poor convergence behavior, or excessive computational costs \cite{piotrowski2020population}. While researchers have introduced various adaptive techniques to mitigate these issues, finding optimal parameter settings remains a challenge. Furthermore, PSO's full-dimensional learning strategy can sometimes lead to poor solutions, as it does not account for correlations between different variables, reducing its effectiveness in problems requiring coordinated learning across multiple dimensions \cite{piotrowski2020population}.

To address these challenges, researchers have introduced various learning strategies aimed at improving the universality and robustness of PSO by dynamically balancing exploration and exploitation. One approach involves adopting multiple learning strategies, allowing particles to update their positions using different mechanisms. For example, adaptive PSO (APSO) \cite{zhan2009adaptive} classifies the search process into four states: exploration, exploitation, convergence, and jumping out, adjusting inertia weights and acceleration coefficients accordingly. Similarly, dynamic neighbor learning PSO (DNLPSO) \cite{nasir2012dynamic} selects an exemplar from the best positions of neighboring particles to enhance information sharing and adaptability.

Beyond neighborhood-based improvements, learning strategies have played a crucial role in advancing PSO performance. Comprehensive learning PSO (CLPSO) \cite{liang2006comprehensive} enhances multimodal problem-solving by utilizing personal best experiences of all particles, improving diversity and global search capabilities. Extensions of CLPSO, such as heterogeneous CLPSO (HCLPSO) \cite{lynn2015heterogeneous}, further refine this approach by dividing the swarm into exploration and exploitation subpopulations. Other notable strategies include self-learning PSO (SLPSO) \cite{wang2011self}, which allows particles to independently adopt different learning behaviors, and genetic learning PSO (GL-PSO) \cite{wang2011enhancing}, which integrates genetic operators (selection, mutation, and crossover) to guide particle movement more effectively.

In addition, several hybrid strategies have been developed to mitigate premature convergence and improve search efficiency. Generalized opposition-based PSO (GOPSO) \cite{wang2011enhancing} leverages opposition-based learning and Cauchy mutation to escape local optima, while orthogonal learning PSO (OLPSO) \cite{zhan2009orthogonal} applies orthogonal experiment design to construct learning exemplars that enhance both personal and global best experiences. Ensemble PSO (EPSO) \cite{lynn2017ensemble} combines multiple learning strategies to optimize real-parameter problems by leveraging their complementary strengths.

Existing studies have highlighted the effectiveness of single- and multi-strategy-based PSO variants in solving optimization problems, as different strategies enhance various aspects of the algorithm. Given the vast number of learning strategies developed over the past decades, a key question arises: Which strategy or combination of strategies can significantly improve PSO's performance? While diverse learning mechanisms provide a valuable foundation, their strategic integration could lead to more robust and efficient optimization techniques. However, a critical challenge remains: how can these strategies be systematically combined to maximize their effectiveness across different problem landscapes?

While several review papers have explored PSO enhancements and variants, as per the author's knowledge, no existing study has specifically focused on the role of learning mechanisms in PSO. Many of these mechanisms have been theoretically proposed, yet their empirical validation and comparative study remain largely underexplored. Despite the introduction of novel strategies, there is a lack of critical assessments evaluating their effectiveness, limitations, and future research directions.

To address this gap, this review provides a structured and comprehensive analysis of PSO learning mechanisms. Through a detailed examination of various strategies, ranging from adaptive and multi-swarm learning to hybrid and ensemble approaches, this study systematically evaluates their impact on PSO's exploration-exploitation balance, convergence behavior, and overall robustness. Additionally, an experimental evaluation is conducted to assess the comparative performance of different learning mechanisms, offering deeper insights into their practical applicability.

This review aims to advance the understanding of how learning strategies can be leveraged to develop next-generation PSO algorithms capable of solving increasingly complex optimization problems. By addressing existing gaps and identifying key research directions, this study contributes to the ongoing development of adaptive, intelligent, and high-performance PSO variants. A list of main contributions is as follows: \begin{enumerate}[(i)]
    \item We systematically categorize existing learning strategies in PSO, including adaptive learning, comprehensive learning, and hybrid learning mechanisms, highlighting their impact on search efficiency and optimization performance.
    \item A structured evaluation of various PSO learning strategies is conducted, analyzing their effectiveness in improving convergence speed, robustness, and diversity across different optimization landscapes.
    \item The review identifies critical gaps in the current literature, such as the lack of unified evaluation frameworks and adaptive mechanisms, providing insights into promising future research directions for enhancing PSO.
\end{enumerate}

The structure of this article is as follows: Section \ref{Sec: PSO} introduces the fundamental theory of PSO. Section \ref{Sec: Single-swarm} reviews single-swarm-based learning strategies, while Sections \ref{sec: two-swarm} and \ref{sec: multiple-swarm} explore two-swarm-based and multiple-swarm-based learning strategies, respectively. Theoretical developments in PSO learning mechanisms are discussed in Section \ref{sec: theoretical}. Section \ref{sec: experimental analysis} presents a comparative experimental study of different learning strategies. Open research questions and future directions are outlined in Section \ref{sec: future directions}, followed by the conclusion in Section \ref{sec: conclusion}.

\section{PSO}\label{Sec: PSO}
PSO is a population-based optimization technique introduced by Kennedy and Eberhart~\cite{kennedy1995particle}. Inspired by the social behavior of birds and fish, PSO has proven effective in solving various optimization problems~\cite{houssein2021major}. In PSO, a swarm of particles navigates through a $D$-dimensional search space, where each particle represents a potential solution. The movement of particles is influenced by both their personal best-known position ($\mathbf{p}_{Best}$) and the best-known position in the swarm ($\mathbf{g}_{Best}$). This allows the swarm to balance exploration and exploitation in the search space. Each particle updates its velocity and position according to the following equations:

\begin{align}\label{eq: pso_velocity}
    \mathbf{vel}_{i}^d = \omega_g \cdot \mathbf{vel}_{i}^d + \overset{\text{cognitive component}}{\overbrace{c_{e_1} \cdot rand_1 \cdot (\mathbf{p}_{Best_i}^d - \mathbf{x}_{i}^d)}} + \overset{\text{social component}}{\overbrace{c_{e_2} \cdot rand_2 \cdot (\mathbf{g}_{Best}^d - \mathbf{x}_{i}^d)}}, 
\end{align}

where $\mathbf{x}_{i} = \{x_{i}^1, x_{i}^2, \dots, x_{i}^d\}$ represents the position of particle $i$ in a swarm of $N_s$ particles. $\mathbf{vel}_{i} = \{vel_{i}^1, vel_{i}^2, \dots, vel_{i}^d\}$ is the velocity of the particle. $\omega_g$ and $c_{e_1}$, $c_{e_2}$ are the inertia weight and the acceleration coefficients, which balance global and local search. $rand_k$ ($k=$1, 2) is a random number drawn from a uniform distribution in $[0,1]$. $\mathbf{p}_{Best_i}$ is the best position found so far by particle $i$. $\mathbf{g}_{Best}$ is the best position discovered by the entire swarm or a neighborhood of particles. The inertia weight $\omega_g$ plays a crucial role in PSO's performance. A larger $\omega_g$ promotes exploration, while a smaller $\omega_g$ enhances exploitation~\cite{clerc2002particle}. PSO's effectiveness has led to its widespread application across various domains \cite{houssein2021major}. However, it may struggle with high-dimensional or multimodal problems due to premature convergence~\cite{chen2012particle}. PSO searches for optimal solutions based on each particle’s previous best position, the best position of a neighboring particle, its current position, and velocity \cite{mendes2004fully,qu2012distance}. However, relying on a single best neighbor may overlook valuable information from other neighbors, potentially slowing convergence or weakening local search \cite{mendes2004fully,qu2012distance,lynn2018population}. 

Various learning strategies and modifications have been proposed to address these challenges. These approaches aim to enhance PSO's global search capability while maintaining convergence efficiency.

\section{Single swarm-based leaning}\label{Sec: Single-swarm}
This section discusses learning strategies based on a single-population (swarm) framework for PSO. A single-swarm approach means that the entire population functions as a unified entity without being divided into multiple subpopulations. In this setup, the learning strategy is applied homogeneously across all particles, ensuring that each particle follows the same adaptation rules and information-sharing mechanisms. Unlike multi-subpopulation methods, where distinct groups evolve independently or interact selectively, single-swarm strategies maintain a collective learning process, allowing global knowledge propagation without structural segmentation. These strategies often focus on refining particle interactions, enhancing convergence behavior, and maintaining a balance between exploration and exploitation within a unified search space. Various approaches, such as comprehensive learning, orthogonal learning, dimensional learning, exemplar-based learning, etc., have been proposed to improve the efficiency and robustness of single-swarm PSO. In the following subsections, we explore different single-swarm learning strategies and their impact on search performance, stability, and solution quality.
\subsection{Comprehensive learning mechanism}
In the original PSO, a particle's flying direction is guided by its individual best position (\(\mathbf{p}_{Best}\)) and the global best position (\(\mathbf{g}_{Best}\)). However, \(\mathbf{g}_{Best}\) often represents a suboptimal local optimum in multimodal optimization problems, leading to premature convergence and reduced diversity in the population \cite{shi1999empirical}. To address these challenges, a \textit{comprehensive learning mechanism} for PSO (CLPSO) strategy was proposed \cite{liang2006comprehensive}.

 CLPSO \cite{liang2006comprehensive} introduces a learning mechanism where a particle does not rely solely on \(\mathbf{g}_{Best}\) or its own \(\mathbf{p}_{Best}\). Instead, it learns from the \(\mathbf{p}_{Best}\) positions of all particles in the population. Each dimension of a particle is updated independently, learning from potentially different exemplars for different dimensions. This mechanism increases population diversity and prevents stagnation in local optima, especially in multimodal landscapes.

 The velocity of the \(i\)th particle in CLPSO is updated using the equation:
\begin{equation}\label{eq: CL velocity}
\mathbf{vel}_i^d = w_g \cdot \mathbf{vel}_i^d + c_e \cdot rand_i^d \cdot \left( \mathbf{p}_{Best_{f_i^d}}^d - \mathbf{x}_i^d \right),
\end{equation}
Where \(w_g\) is the inertia weight balancing exploration and exploitation, \(c_e\) is the acceleration coefficient determining the influence of the exemplar, \(rand_i^d\) is a random number in \([0, 1]\) introducing stochasticity, and \(\mathbf{p}_{Best_{f_i^d}}^d\) represents the exemplar position for dimension \(d\), which can be the particle's own \(\mathbf{p}_{Best}\) or another particle's \(\mathbf{p}_{Best}\), determined by the learning probability \(P_{r_c}\).

The decision of whether to follow its own \(\mathbf{p}_{Best}\) or another particle's \(\mathbf{p}_{Best}\) for each dimension is based on a learning probability \(P_{r_{c,~i}}\), calculated as:
\begin{equation}\label{eq: CL probability}
P_{r_{c, i}} = \alpha + \beta \cdot \exp \left( \frac{10\cdot (i - 1)}{N_s - 1} \right),
\end{equation}
where \(\alpha = 0.05\), \(\beta = 0.45\) are predefined constants controlling the range of \(P_{r_{c, i}}\), and \(N_s\) represents the population size. For each dimension of a particle, a random number is generated and compared with \(P_{r_{c, i}}\). If the random number is smaller than \(P_{r_{c, i}}\), the particle follows another particle's \(\mathbf{p}_{Best}\). Otherwise, it learns from its \(\mathbf{p}_{Best}\). The exemplar \(\mathbf{p}_{Best_{f_i^d}}^d\) is determined as follows:
\begin{enumerate}[(i)]
    \item \textit{Tournament selection}: Two particles are randomly chosen from the population (excluding the particle being updated).
    \item \textit{Fitness comparison}: The fitness values of the two particles' \(\mathbf{p}_{Best}\) are compared, and the better \(\mathbf{p}_{Best}\) is chosen as the exemplar for the corresponding dimension.
\end{enumerate}

This selection process ensures that particles are guided by high-quality exemplars, balancing exploration and exploitation. The step-wise procedure of the CL mechanism is given in Algorithm~\ref{algo: CL}.

\begin{algorithm}
     \caption{Pseudo code of CL strategy.}\label{algo: CL}
     \begin{algorithmic}[1]
     \STATE Update the velocity using Eq.~\eqref{eq: CL velocity} and update the population,
     \STATE Evaluate the fitness of each individual,
     \IF{$f_{\mathbf{x}_i^d}<f_{\mathbf{p}_{Best_i}^d}$}
     \STATE $\mathbf{p}_{Best_i}^d\gets\mathbf{x}_i^d$, $g_i=0$,
     \ELSE \STATE $g_i\gets g_i+1$,
     \ENDIF
     \IF{$g_i>R_g$}
     \STATE Update $\mathbf{p}_{Best_{f_i^d}}^d$
     \ENDIF
     \end{algorithmic}
 \end{algorithm}
To further enhance adaptability, CLPSO incorporates a refreshing gap \(R_g\), defining a stagnation threshold. If a particle's \(\mathbf{p}_{Best}\) does not improve for \(R_g\) consecutive evaluations, its learning exemplar is reassigned, and a new \(\mathbf{p}_{Best_{f_i^d}}\) is generated. This mechanism avoids wasting time on poor search directions and encourages continual improvement in particle movement. The procedure for updating \(\mathbf{p}_{Best}\) is illustrated in Algorithm \ref{algo: pbest in CL}.

\begin{algorithm}[H]
\caption{Exemplar $\mathbf{p}_{Best_{f_i^d}}^d$.}\label{algo: pbest in CL}
\begin{algorithmic}[1]
\FOR{each dimension \( j \)}
    \IF{$\texttt{rand} < P_{r_{c_i}} $}
        \STATE \( a \gets \texttt{RANDOM}(1, N_s) \)
        \STATE \( b \gets \texttt{RANDOM}(1, N_s) \)
        \IF{\( f_{\mathbf{p}_{\text{Best}_a}} \leq f_{\mathbf{p}_{\text{Best}_b}} \)}
            \STATE \( \mathbf{x}^j_i \gets \mathbf{p}_{\text{Best}_a}^j \)
        \ELSE
            \STATE \( \mathbf{x}^j_i \gets \mathbf{p}_{\text{Best}_b}^j \)
        \ENDIF
    \ELSE
        \STATE \( \mathbf{x}^j_i \gets \mathbf{p}_{\text{Best}_i}^j \)
    \ENDIF
\ENDFOR
\end{algorithmic}
\end{algorithm}
CLPSO has several differences with PSO, such as: (i) Unlike PSO, which uses the particle's \(\mathbf{p}_{Best}\) and \(\mathbf{g}_{Best}\) as fixed exemplars, CLPSO allows every particle to learn from the \(\mathbf{p}_{Best}\) positions of any other particle. This flexibility provides a richer source of guidance, especially in complex multimodal problems. (ii) Each dimension of a particle can learn from different exemplars, providing independent and diverse updates to each coordinate. This contrasts with PSO, where the entire particle learns from the same exemplars for all dimensions. (iii) Instead of learning from both \(\mathbf{p}_{Best}\) and \(\mathbf{g}_{Best}\) simultaneously, each dimension in CLPSO learns from a single exemplar. This reduces the risk of conflicting updates and ensures focused learning.


By allowing each dimension to learn from different \(\mathbf{p}_{Best}\) sources, CLPSO enhances population diversity and effectively mitigates premature convergence. The algorithm dynamically balances exploration (learning from others' \(\mathbf{p}_{Best}\)) and exploitation (refining its own \(\mathbf{p}_{Best}\)) through the learning probability \(P_{r_{c, i}}\) and the refreshing gap \(R_g\). This dynamic adaptability enables CLPSO to excel in multimodal optimization tasks, as it promotes the exploration of multiple optima and avoids entrapment in suboptimal solutions.

\begin{figure}
    \centering
    \includegraphics[width=1\linewidth]{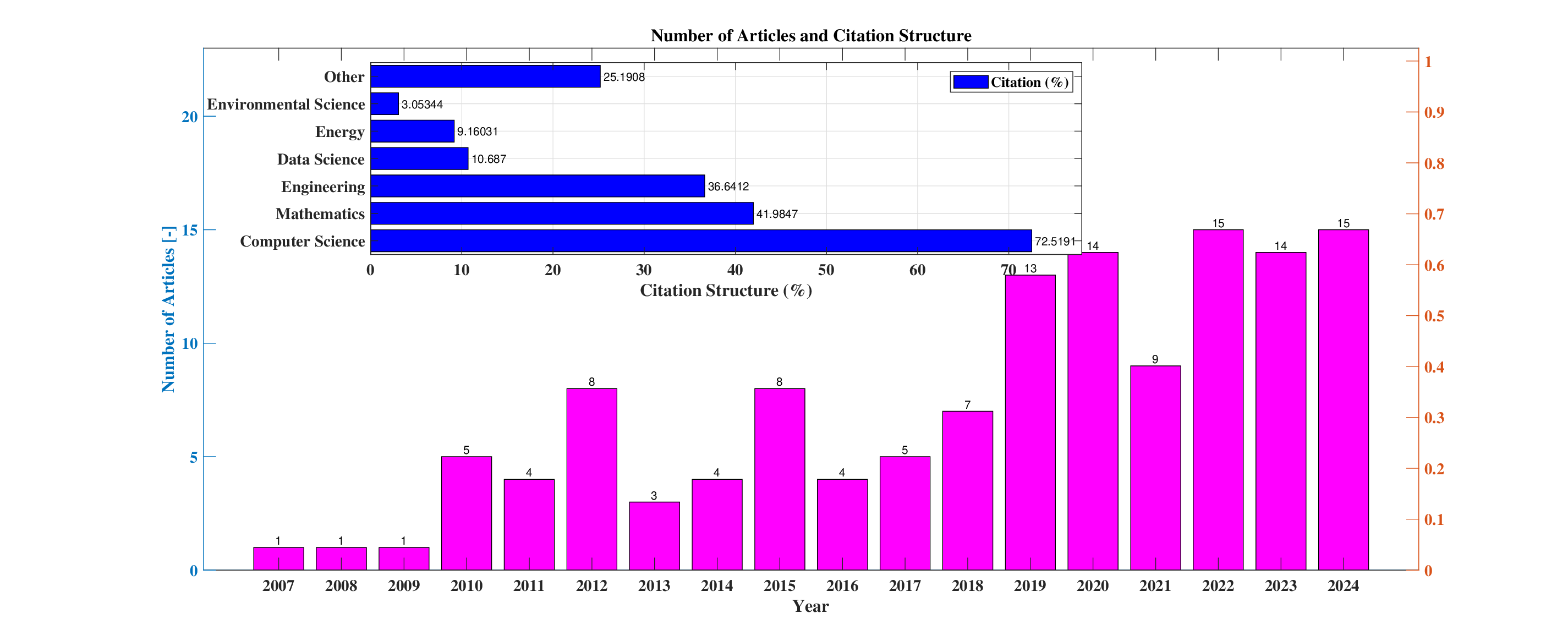}  
    \caption{Number of annually published documents on CL and application in the different fields ($\%$) based on the Scopus platform.}
    \label{fig: CL articles}
\end{figure}
\begin{figure}
    \centering 
       \includegraphics[width=0.6\linewidth]{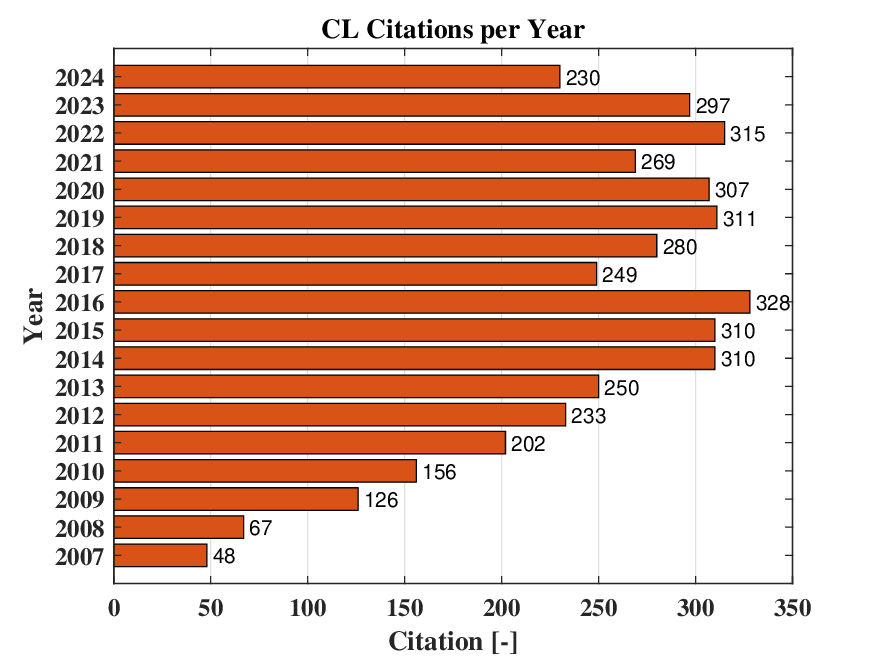} 
    \caption{CL citations per year based on the Google Scholar platform.}
    \label{fig: CL citations}
\end{figure}

Fig.~\ref{fig: CL articles} depicts the annual growth of research articles on the CL mechanism and its applications across various fields, along with the citation structure ($\%$). Between 2007 and 2018, the average number of publications remained around 4 per year, indicating limited early activity. Notable peaks occurred in 2012 and 2015, with 8 articles each year, followed by a steady phase from 2014 to 2018, with 4–7 articles annually. A significant surge began in 2019, averaging 13 articles per year from 2019 to 2024, underscoring the growing importance of CL in optimization and machine learning. Computer Science dominates with 72.52$\%$ of citations, highlighting its primary role in machine learning and optimization. Engineering (38.64$\%$), Mathematics (41.98$\%$), and Data Science (10.69$\%$) also show significant adoption, emphasizing the interdisciplinary nature of CL. Fields like Energy (9.16$\%$), Environmental Science (3.05$\%$), and others (25.19$\%$) have lower citation percentages, suggesting emerging but less impactful applications. The annual citations of the original CL paper \cite{liang2006comprehensive}, as shown in Fig.~\ref{fig: CL citations}, average 238 citations per year, showcasing its widespread recognition and global influence.

Some notable implementations of the CL mechanism are: Lynn et al. \cite{lynn2015heterogeneous} integrated the CL mechanism into PSO (HCLPSO) by dividing the population into two group sizes, enhancing PSO's diversity. This approach was further extended into C-HCLPSO \cite{yousri2019static,van2023chaotic}, incorporating chaotic strategies for photovoltaic parameter identification. Additional extensions include HCLDMSPSO \cite{wang2020heterogeneous,yousri2021parameters}, combining CL for one subpopulation and a dynamic multi-swarm strategy (discussed in Section \ref{sec: DMS}) for another, boosting diversity. Similarly, Cao et al. \cite{cao2018comprehensive} integrated local search and CL in CLPSO-LS to improve PSO's performance. Maitra and Chatterjee \cite{maitra2008hybrid} combined cooperative learning and CL in PSO for multi-level image segmentation. Gulcu and Kodaz \cite{gulcu2015novel} proposed a parallel CLPSO (PCLPSO) using a master-slave paradigm based on multi-swarm collaboration, enhancing performance through parallelism. Zhong et al. \cite{zhong2018discrete} tailored CLPSO for the Traveling Salesman Problem in D-CLPSO by introducing a simulated annealing acceptance criterion, novel flight equations, and lazy velocity computation to boost efficiency and prevent premature convergence. The CL mechanism has also been implemented in JAYA (CLJAYA) \cite{zhang2020comprehensive,zhang2022comprehensive} to optimize photovoltaic parameters.

Zhang et al. \cite{zhang2019enhancing} proposed a local optima topology (CLPSO-LOT), wherein local optima identified during iterations form a new topology, expanding the search space and accelerating convergence. Yu et al. \cite{yu2014enhanced} introduced two enhancements in ECLPSO: a perturbation term in velocity updates activated by personal best bounds to improve exploitation and adaptive learning probabilities based on personal best rankings to enhance convergence. Tan et al. \cite{tan2015adaptive} proposed an adaptive CL bacterial foraging optimization (ALCBFO) algorithm with a time-varying chemotaxis step length and CL strategy. A non-linearly decreasing modulation model balanced exploration and exploitation, while the learning mechanism enhanced diversity and prevented premature convergence. Lin et al. \cite{lin2019adaptive} combined an adaptive mechanism for dynamic CL probability adjustment and a cooperative archive in ACLPSO-CA. This mechanism dynamically adjusted each particle's CL probability across three levels based on performance, while the archive provided additional promising information through swarm cooperation. These studies collectively highlight the versatility and adaptability of the CL mechanism across various optimization contexts, showcasing its potential for solving complex and interdisciplinary problems.

The introduction of the learning strategy in CLPSO, where a particle's velocity is updated either using its own \(\mathbf{p}_{Best}\) position or the better one from two randomly selected particles, is an innovative idea. This process is guided by the learning probability factor \(P_{r_{c, i}}\), which varies exponentially with the particle's index. However, the random selection of exemplars in the learning process may not always provide optimal guidance, as the exemplars chosen might lead the particle toward incorrect directions in the search space.  

The challenge arises when the momentum gained by a particle from exemplars at different stages causes significant deviation, mainly if the exemplars are located in distant and unconnected regions of the search space. This lack of interconnection between the exemplars and the particle can reduce the algorithm's effectiveness \cite{lin2019adaptive,abd2024fractional}. Consequently, refining the learning strategy is essential to provide more reliable and directed guidance in the search process. The other limitation of the CL mechanism is finding the best results on unimodal problems \cite{lynn2015heterogeneous}.
 To address the drawbacks of CLPSO, several enhanced strategies have been proposed, as discussed above. The dynamic neighborhood learning PSO (DNLPSO) \cite{nasir2012dynamic} introduces a learning strategy (Eq. \eqref{eq: CL velocity with gbest}, where parameters $w_g$ varies 0.9 to 0.4 and $c_{e_1}=c_{e_2}=1.49445$ are selected) where a particle's velocity is updated using the historical best information of particles within a dynamic neighborhood, rather than the entire population. These neighborhoods are reformed periodically, allowing participants to learn from their neighborhood or historical best practices, thus enhancing guidance and maintaining diversity. 
 \begin{equation}\label{eq: CL velocity with gbest}
\mathbf{vel}_i^d = w_g \cdot \mathbf{vel}_i^d + c_{e_1} \cdot rand_i^d \cdot \left( \mathbf{p}_{Best_{f_i^d}}^d - \mathbf{x}_i^d \right)+ c_{e_2} \cdot rand_i^d \cdot \left( \mathbf{g}_{Best}^d - \mathbf{x}_i^d \right),
\end{equation}
Additionally, $P_{r_{c, i}}$ adopts a linear variation within the same range (Eq. \eqref{eq: DNL probability}), with the intent of enhancing each particle's learning tendency compared to the exponential variation. 
\begin{equation}\label{eq: DNL probability}
P_{r_{c, i}} = \alpha + \beta \cdot \left( \frac{10\cdot (i - 1)}{N_s - 1} \right).
\end{equation}

 Similarly, the heterogeneous CLPSO (HCLPSO) \cite{lynn2015heterogeneous,wang2020heterogeneous} divides the population into two heterogeneous subpopulations: the first subpopulation generates exemplars using \(\mathbf{p}_{Best}\) experiences from within its subpopulation (Eq. \eqref{eq: CL velocity} where $w_g$ varies from 0.99 to 0.2 and $c_{e}$ varies from 3 to 1.5), preserving diversity, while the second subpopulation uses \(\mathbf{p}_{Best}\) experiences from the entire population to focus on global exploration (Eq. \eqref{eq: CL velocity with gbest} uses parameters where $w_g$ varies from 0.99 to 0.2, $c_{e_1}$ varies from 2.5 to 0.5, and $c_{e_2}$ varies from 0.5 to 2.5.). The $P_{r_{c, i}}$ values for each particle in the population are calculated using Eq. \eqref{eq: CL probability} with $\alpha=0$, and $\beta=0.25$. This division ensures that diversity is retained even if one subpopulation converges prematurely. Furthermore, the parallel CLPSO (PCLPSO) \cite{gulcu2015novel} leverages a master-slave parallel paradigm with multiple swarms working cooperatively and concurrently, improving exploration and computational efficiency. To enhance local search capabilities, a Broyden-Fletcher-Goldfarb-Shanno (BFGS) based local search operator was incorporated into CLPSO \cite{cao2018comprehensive}, combining the global exploration strengths of CLPSO with the precision of BFGS optimization. A $p_{Best}$-guided CLPSO (HPBPSO) \cite{meng2024heterogeneous} has also been proposed, which organizes the population similarly to HCLPSO. The first subpopulation is updated using Eq. \eqref{eq: CL velocity with gbest} with dynamic crossover learning, while the second subpopulation is further divided into two groups and updated using Eq. \eqref{eq: HPBPSO velocity}. \begin{equation}\label{eq: HPBPSO velocity}
\mathbf{vel}_i^d = w_g \cdot \mathbf{vel}_i^d + c_{e_1} \cdot rand_i^d \cdot \left( \mathbf{p}_{Best_{f_i^d}}^d - \mathbf{x}_i^d \right)+ c_{e_2} \cdot rand_i^d \cdot \left( \mathbf{g}_{Best}^d - \mathbf{x}_i^d \right) + (rand_i^d)^\lambda \cdot \left( \mathbf{g}_{Best}^d - \mathbf{p}_{Best_i}^d \right).
\end{equation}

 These advancements collectively enhance CLPSO's robustness and adaptability for complex optimization problems by mitigating its inherent limitations and introducing innovative learning strategies.

\subsection{Orthogonal learning}

In the orthogonal learning PSO (OLPSO) \cite{zhan2009orthogonal}, the orthogonal learning (OL) strategy combines the personal best position \(\mathbf{p}_{Best}\) and the neighborhood best position \(\mathbf{n}_{Best}\) to form a guidance vector \(\mathbf{o}_{Best}\). The particle updates its velocity as:

\begin{equation}\label{eq:OL_velocity}
\mathbf{vel}_i^d = w_g \cdot \mathbf{vel}_i^d + c_e \cdot rand_1^d \cdot \left( \mathbf{o}_{Best}^d - \mathbf{x}_i^d \right),
\end{equation}

where \(c_e = 2.0\) is the acceleration coefficient, and \(rand_1^d\) is a random value in \([0, 1]\). The guidance vector \(\mathbf{o}_{Best}\), constructed using the orthogonal experimental design (OED) method, is given by:

\begin{equation}
    \mathbf{o}_{Best} = \mathbf{p}_{Best} \oplus \mathbf{n}_{Best},
\end{equation}

where \(\oplus\) denotes the OED operation. Each element of \(\mathbf{o}_{Best}\) is selected from \(\mathbf{p}_{Best}\) or \(\mathbf{n}_{Best}\) based on the OED process. This vector serves as a dynamic learning exemplar, guiding the particle's velocity and position updates while refining its personal best \(\mathbf{p}_{Best}\).

To maintain stability, \(\mathbf{o}_{Best}\) is reused for multiple generations until it ceases to improve the particle’s exploration. If \(\mathbf{p}_{Best}\) remains unchanged for \(t\) generations, a new \(\mathbf{o}_{Best}\) is reconstructed. To incorporate dynamic updates, \(\mathbf{o}_{Best}\) stores only the indices of \(\mathbf{p}_{Best}\) and \(\mathbf{n}_{Best}\), enabling immediate integration of new information when these positions improve.

The construction of \(\mathbf{o}_{Best}\) begins with generating an orthogonal array (OA), \(L_M(2^d)\), where \(L\) represents a Latin square, \(M\) is the number of test cases, and \(d\) is the problem's dimensionality. Each dimension is treated as a factor with two levels: \(\mathbf{p}_{Best}\) or \(\mathbf{n}_{Best}\). Using the OA, \(M\) candidate solutions \(\mathbf{x}_j\) (\(1 \leq j \leq M\)) are generated. For each factor, if the OA value is 1, the dimension is selected from \(\mathbf{p}_{Best}\); otherwise, it is chosen from \(\mathbf{n}_{Best}\). Candidate solutions are evaluated for fitness, and the best solution, \(best_x\), is recorded. Factor effects are analyzed to determine the optimal level for each factor. A predictive solution \(x_p\) is constructed and evaluated using these levels. The fitness values \(f(best_x)\) and \(f(x_p)\) are compared, and the better solution’s level combination is used to construct \(\mathbf{o}_{Best}\).

This process ensures that \(\mathbf{o}_{Best}\) efficiently combines the strengths of \(\mathbf{p}_{Best}\) and \(\mathbf{n}_{Best}\), balancing exploration and exploitation. By leveraging the OED method, OLPSO significantly enhances the particle swarm optimization framework's learning capability.

\begin{figure}
    \centering 
       \includegraphics[width=0.6\linewidth]{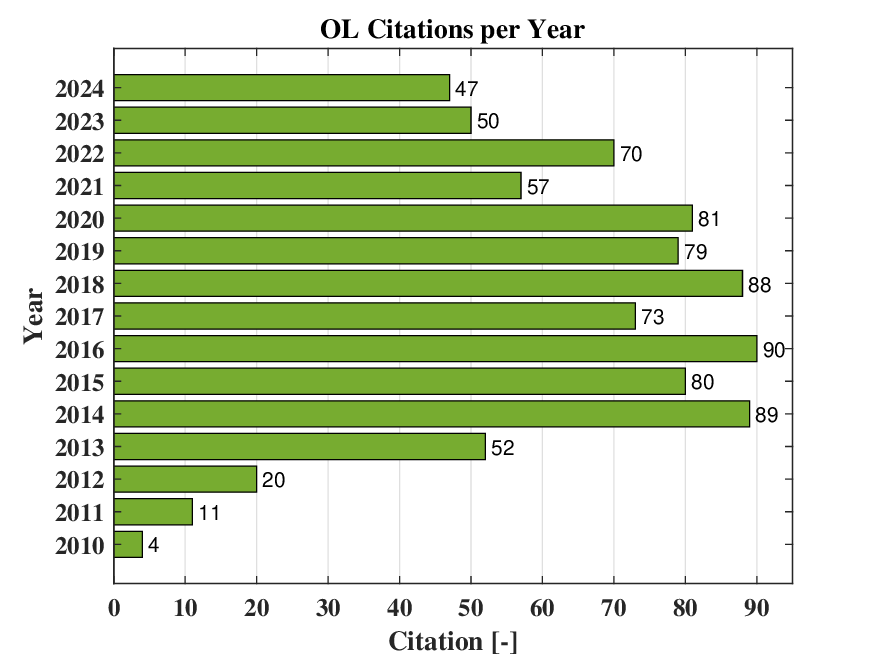} 
    \caption{OL citations per year based on the Google Scholar platform.}
    \label{fig: OL citations}
\end{figure}

A novel PSO variant \cite{wang2022novel}, integrating Lévy flight and OL, enhances exploitation and accelerates search efficiency. Lévy flight boosts exploration, while OL strengthens exploitation and enables dynamic, adaptive learning by incorporating the L\'{e}vy term based on flight probability. Parallel computation in OL further improves efficiency. Unlike PSO, this variant employs a novel strategy where particles learn from a single optimal experience, avoiding conflicts between historical and global best guides. Ho et al. \cite{ho2008opso} introduced an intelligent move mechanism (IMM) into OLPSO, utilizing OED to adjust particle velocities systematically. This divide-and-conquer approach addresses the curse of dimensionality, ensuring efficient particle movement. OLPSO has been applied to plant disease diagnosis by integrating it into a CNN with a decaying learning rate \cite{darwish2020optimized}.

For dynamic optimization problems, OLPSO with a variable relocation strategy (VRS) \cite{wang2016orthogonal} balances fast convergence and population diversity. VRS uses historical data for particle relocation, enabling quick adaptation to changing environments. Bai et al. \cite{bai2024differential} proposed an OLPSO integrated with DE, enhancing exploitation through deeper searches and preventing premature convergence with a stochastic star topology. OL has also improved optimization in various algorithms, including CSO \cite{xiong2018orthogonal}, ABC \cite{bai2017improved,gao2013novel}, AEFA \cite{chauhan2024archive}, DE \cite{li2022efficient,kumar2022differential}, CS \cite{li2014enhancing}, AOA \cite{houssein2021enhanced}, and SCA \cite{chen2020advanced}. These applications highlight OL's versatility in enhancing metaheuristic optimization. Fig. \ref{fig: OL citations} validated its applicability in various years, which depicts the annual citation trends for OL from 2010 to 2024, based on data from Google Scholar. Citations began modestly in 2010 with only 4 citations and experienced a consistent rise, reaching a peak of 90 citations in 2016. After this peak, there was a slight decline, with citations dropping to 73 in 2018 before rebounding to 89 in 2019. The years following 2019 show a gradual decrease, with 81 citations in 2020, 70 in 2022, and 47 in 2024. This trend highlights the increasing recognition of OL in the early years, particularly between 2013 and 2016, followed by fluctuations in recent years. The decline in citations after 2020 may indicate either a shift in research focus or the stabilization of interest as the field matures.

\subsection{Dimensional learning strategy}
Inspired by CLPSO and OLPSO, Xu et al. \cite{xu2019particle} proposed a dimensional learning strategy (DLS) for PSO to effectively discover and preserve the promising information of the population's best solution and construct a learning exemplar for each particle. The learning exemplar is constructed by allowing each dimension of the $\mathbf{p}_{Best}$ to learn from the corresponding $\mathbf{g}_{Best}$ on a dimension-by-dimension basis to construct a learning exemplar $\mathbf{x}_{dl}$. This process ensures the excellent information of $\mathbf{g}_{Best}$ is propagated to $\mathbf{x}_{dl}$, thereby promoting the dissemination and preservation of the coding pattern of $\mathbf{g}_{Best}$. During the DLS process, DLS guarantees that $\mathbf{x}_{dl}$ is no worse than $\mathbf{p}_{Best}$.

The velocity of PSO has been changed by integrating the DLS, DLPSO, by substituting $\mathbf{p}_{Best}$ with $\mathbf{x}_{dl}$ in the PSO velocity update formula (Eq. \eqref{eq: DLS velocity}).
\begin{equation}\label{eq: DLS velocity}
\mathbf{vel}_i^d = w_g \cdot \mathbf{vel}_i^d + c_{e_1} \cdot rand_i^d \cdot \left( \mathbf{x}_{dl_i}^d - \mathbf{x}_i^d \right)+ c_{e_2} \cdot rand_i^d \cdot \left( \mathbf{g}_{Best}^d - \mathbf{x}_i^d \right).
\end{equation}

\begin{figure}
	\centering
\resizebox{0.8\linewidth}{0.6\linewidth}{	\begin{tikzpicture}[node distance=1.5cm,auto]
	\tikzset{->-/.style={decoration={
				markings,
				mark=at position #1 with {\arrow{>}}},postaction={decorate}}}
         
            \draw [help lines,line width=2pt] (4,11) grid (9,12);
     	\node[blue] at (4.5,11.5) {2};\node[blue] at (5.5,11.5) {2};\node[blue] at (6.5,11.5) {2};\node[blue] at (7.5,11.5) {4};\node[blue] at (8.5,11.5) {0};
     \node at (3.5,11.5) {$\mathbf{g}_{Best}$};\node at (10,11.5) {$f_{\mathbf{g}_{Best}}=28$};
        \node at (1.5,10.2) {$\mathbf{p}_{Best}$};
	\draw [help lines,line width=2pt] (1,10) grid (2,5);
	\node[blue] at (1.5,9.5) {1};\node[blue] at (1.5,8.5) {0};\node[blue] at (1.5,7.5) {3};\node[blue] at (1.5,6.5) {4};\node[blue] at (1.5,5.5) {4};

       \node at (3.3,10.2) {$\mathbf{x}_{tp}$};
    \draw [help lines,line width=2pt] (3,10) grid (4,5);
	\node[red] at (3.5,9.5) { \bf 2};\node[blue] at (3.5,8.5) {0};\node[blue] at (3.5,7.5) {3};\node[blue] at (3.5,6.5) {2};\node[blue] at (3.5,5.5) {4};
         \node[black] at (3.5,4.5) {\small $f_{\mathbf{x}_{tp}}$=33};
        \node[black] at (3.5,4) {\small $f_{\mathbf{x}_{tp}}>f_{\mathbf{x}_{dl}}$};

\node at (5.2,10.2) {$\mathbf{x}_{tp}$};
    \draw [help lines,line width=2pt] (5,10) grid (6,5);
	\node[blue] at (5.5,9.5) {1};\node[red] at (5.5,8.5) {\bf 2};\node[blue] at (5.5,7.5) {3};\node[blue] at (5.5,6.5) {2};\node[blue] at (5.5,5.5) {4};
         \node[black] at (5.5,4.5) {\small $f_{\mathbf{x}_{tp}}$=34};
        \node[black] at (5.5,4) {\small $f_{\mathbf{x}_{tp}}>f_{\mathbf{x}_{dl}}$};

    \node at (7.7,10.2) {$\mathbf{x}_{tp}$};
    \draw [help lines,line width=2pt] (7,10) grid (8,5);
	\node[blue] at (7.5,9.5) {1};\node[blue] at (7.5,8.5) {0};\node[red] at (7.5,7.5) {\bf 2};\node[blue] at (7.5,6.5) {2};\node[blue] at (7.5,5.5) {4};
         \node[black,fill=yellow] at (7.5,4.5) {\small $f_{\mathbf{x}_{tp}}$=25};
        \node[black,fill=yellow] at (7.5,3.95) {\small $f_{\mathbf{x}_{tp}}<f_{\mathbf{x}_{dl}}$};

    \node at (9.7,10.2) {$\mathbf{x}_{tp}$};
    \draw [help lines,line width=2pt] (9,10) grid (10,5);
	\node[blue] at (9.5,9.5) {1};\node[blue] at (9.5,8.5) {0};\node[blue] at (9.5,7.5) {2};\node[red] at (9.5,6.5) {\bf 4};\node[blue] at (9.5,5.5) {4};
         \node[black] at (9.5,4.5) {\small $f_{\mathbf{x}_{tp}}$=35};
        \node[black] at (9.5,4) {\small $f_{\mathbf{x}_{tp}}>f_{\mathbf{x}_{dl}}$};
    \node at (11.7,10.2) {$\mathbf{x}_{tp}$};
    
\draw [help lines,line width=2pt] (11,10) grid (12,5);
	\node[blue] at (11.5,9.5) {1};\node[blue] at (11.5,8.5) {0};\node[blue] at (11.5,7.5) {2};\node[blue] at (11.5,6.5) {2};\node[red] at (11.5,5.5) {\bf 0};
     \node[black,fill=yellow] at (11.5,4.5) {\small $f_{\mathbf{x}_{tp}}$=9};
        \node[black,fill=yellow] at (11.5,3.95) {\small $f_{\mathbf{x}_{tp}}<f_{\mathbf{x}_{dl}}$};
    \draw [help lines,line width=2pt] (1,3) grid (2,-2);
	\node[blue] at (1.5,2.5) {1};\node[blue] at (1.5,1.5) {0};\node[blue] at (1.5,0.5) {3};\node[blue] at (1.5,-0.5) {2};\node[blue] at (1.5,-1.5) {4};
        \node[black] at (1.5,-2.3) {\small $\mathbf{x}_{dl}$};
        \node[black] at (1.5,-2.7) {\small $f_{\mathbf{x}_{dl}}=33$};

    \draw [help lines,line width=2pt] (3,3) grid (4,-2);
	\node[blue] at (3.5,2.5) {1};\node[blue] at (3.5,1.5) {0};\node[blue] at (3.5,0.5) {3};\node[blue] at (3.5,-0.5) {2};\node[blue] at (3.5,-1.5) {4};
        \node[black] at (3.5,-2.3) {\small $\mathbf{x}_{dl}$};
        \node[black] at (3.5,-2.7) {\small $f_{\mathbf{x}_{dl}}=30$};

    \draw [help lines,line width=2pt] (5,3) grid (6,-2);
	\node[blue] at (5.5,2.5) {1};\node[blue] at (5.5,1.5) {0};\node[blue] at (5.5,0.5) {3};\node[blue] at (5.5,-0.5) {2};\node[blue] at (5.5,-1.5) {4};
        \node[black] at (5.5,-2.3) {\small $\mathbf{x}_{dl}$};
        \node[black] at (5.5,-2.7) {\small $f_{\mathbf{x}_{dl}}=30$};
        
    \draw [help lines,line width=2pt] (7,3) grid (8,-2);
	\node[violet] at (7.5,2.5) {\bf 1};\node[violet] at (7.5,1.5) {\bf 0};\node[violet] at (7.5,0.5) {\bf 2};\node[violet] at (7.5,-0.5) {\bf 2};\node[violet] at (7.5,-1.5) {\bf 4};
        \node[black] at (7.5,-2.3) {\small $\mathbf{x}_{dl}$};
        \node[black] at (7.5,-2.7) {\small $f_{\mathbf{x}_{dl}}=25$};
        
    \draw [help lines,line width=2pt] (9,3) grid (10,-2);
	\node[blue] at (9.5,2.5) {1};\node[blue] at (9.5,1.5) {0};\node[blue] at (9.5,0.5) {2};\node[blue] at (9.5,-0.5) {2};\node[blue] at (9.5,-1.5) {4};
        \node[black] at (9.5,-2.3) {\small $\mathbf{x}_{dl}$};
        \node[black] at (9.5,-2.7) {\small $f_{\mathbf{x}_{dl}}=25$};
        
     \draw [help lines,line width=2pt] (11,3) grid (12,-2);
	\node[red] at (11.5,2.5) {\bf 1};\node[red] at (11.5,1.5) {\bf 0};\node[red] at (11.5,0.5) {\bf 2};\node[red] at (11.5,-0.5) {\bf 2};\node[red] at (11.5,-1.5) {\bf 0};
        \node[black] at (11.5,-2.3) {\small $\mathbf{x}_{dl}$};
        \node[black] at (11.5,-2.7) {\small $f_{\mathbf{x}_{dl}}=9$};

    \draw [help lines, line width=2pt] (13,6.5) grid (14,2.5);
	\node at (13.5,6.25) {1};\node at (13.5,5.5) {0};\node at (13.5,4.5) {2};\node at (13.5,3.5) {2};\node at (13.5,2.75) {0};
    \draw[line width=2pt,gray]  (13,6.5)--(14,6.5);\draw[line width=2pt,gray]  (13,2.5)--(14,2.5);
        \node[black] at (13.5,6.8) {\small Final exemplar};
        \node[black] at (13.5,2.2) {\small $\mathbf{x}_{dl}$};
        \node[black] at (13.5,1.8) {\small $f_{\mathbf{x}_{dl}}=9$};

        \draw [arrow, line width=1pt] (4.5,11) -- (3.5,10);\draw [arrow, line width=1pt] (5.5,11) -- (5.5,10);\draw [arrow, line width=1pt] (6.5,11) -- (7.5,10);\draw [arrow, line width=1pt] (7.5,11) -- (9.5,10);\draw [arrow, line width=1pt] (8.5,11) -- (11.5,10);

        \draw [arrow, line width=1.5pt] (1.5,5) -- (1.5,3);\draw [arrow, line width=1.5pt] (3.5,3.9) -- (3.5,3);\draw [arrow, line width=1.5pt] (5.5,3.9) -- (5.5,3);\draw [arrow, line width=1.5pt] (7.5,3.7) -- (7.5,3);\draw [arrow, line width=1.5pt] (9.5,3.9) -- (9.5,3);\draw [arrow, line width=1.5pt] (11.5,3.7) -- (11.5,3);

        \draw [dashed, line width=0.5pt] (0.7,-3) -- (0.7,10.5)--(2.3,10.5)--(2.3,-3)--(0.7,-3);
        \draw [arrow, line width=1.5pt, orange] (2.3,4) -- (2.7,4);
        \draw [dashed, line width=0.5pt] (2.7,-3) -- (2.7,10.5)--(4.3,10.5)--(4.3,-3)--(2.7,-3);
         \draw [arrow, line width=1.5pt, orange] (4.3,4) -- (4.7,4);
        \draw [dashed, line width=0.5pt] (4.7,-3) -- (4.7,10.5)--(6.3,10.5)--(6.3,-3)--(4.7,-3);
         \draw [arrow, line width=1.5pt, orange] (6.3,4) -- (6.7,4);
        \draw [dashed, line width=0.5pt] (6.7,-3) -- (6.7,10.5)--(8.3,10.5)--(8.3,-3)--(6.7,-3);
         \draw [arrow, line width=1.5pt, orange] (8.3,4) -- (8.7,4);
        \draw [dashed, line width=0.5pt] (8.7,-3) -- (8.7,10.5)--(10.3,10.5)--(10.3,-3)--(8.7,-3);
         \draw [arrow, line width=1.5pt, orange] (10.3,4) -- (10.7,4);
        \draw [dashed, line width=0.5pt] (10.7,-3) -- (10.7,10.5)--(12.3,10.5)--(12.3,-3)--(10.7,-3);
         \draw [arrow, line width=2.5pt, orange] (12.3,4) -- (13,4);
	\end{tikzpicture}}
	\caption{An illustration of creating $\mathbf{x}_{dl}$ in DLS \cite{xu2019particle}.}
	\label{fig: Illustration DLS}
\end{figure}
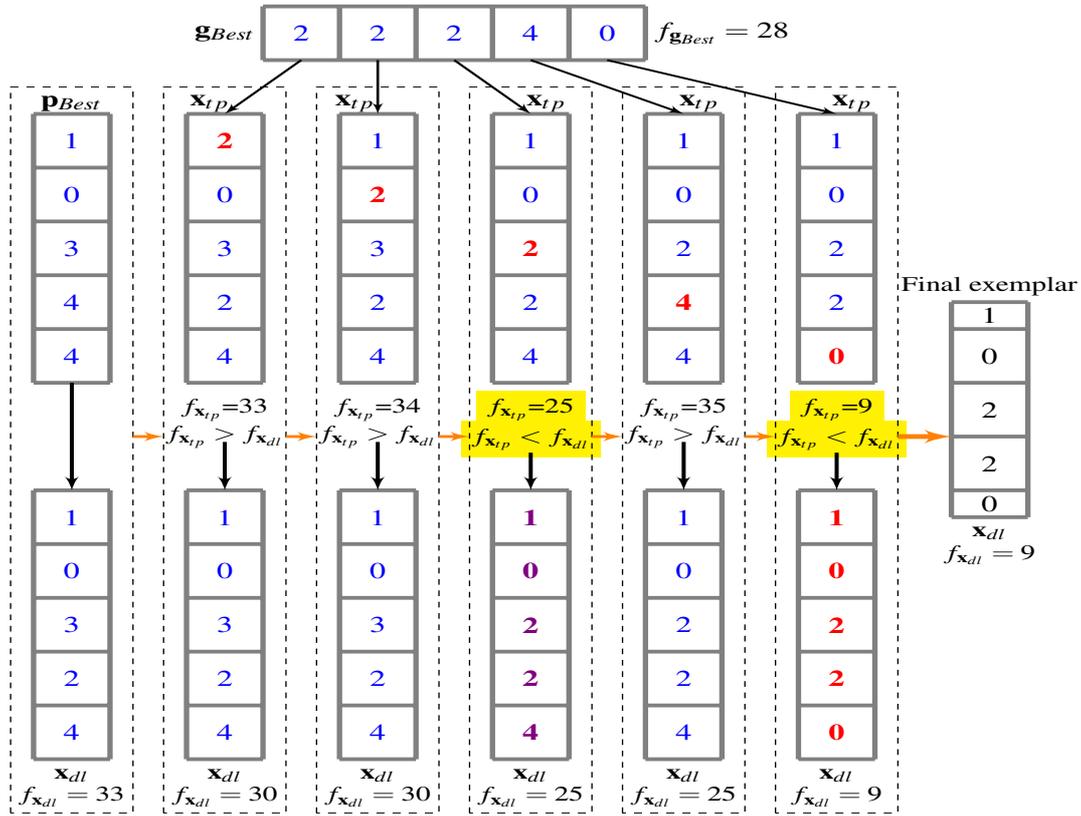

\begin{algorithm}[h]
\caption{Exemplar $\mathbf{x}_{dl}$.}\label{algo: exemplar DLS}
\label{alg:dls}
\begin{algorithmic}[1]
\STATE $\mathbf{x}_{dl} \gets \mathbf{p}_{Best}$  \COMMENT{Initialize the learning exemplar with the personal best position}
\FOR{each dimension $j$}
    \STATE $\mathbf{x}_{tp} \gets \mathbf{x}_{dl}$  \COMMENT{Copy the current learning exemplar}
    \IF{$\mathbf{x}_{tp}^j == \mathbf{g}_{Best}^j$}
        \STATE \textbf{continue}  \COMMENT{Skip if the current dimension already matches the global best}
    \ENDIF
    \STATE $\mathbf{x}_{tp}^j \gets \mathbf{g}_{Best}^j$  \COMMENT{Update the current dimension with the global best value}
    \IF{$f_{\mathbf{x}_{tp}^j} < f_{\mathbf{x}_{dl}^j}$}
        \STATE $\mathbf{x}_{dl}^j \gets \mathbf{g}_{Best}$  \COMMENT{Accept the update if it improves fitness}
    \ENDIF
\ENDFOR
\end{algorithmic}
\end{algorithm}

As shown in Fig. \ref{fig: Illustration DLS}, the temporary exemplar $\mathbf{x}_{dl}$ is compared to the current exemplar each time a dimension of the former is updated. If $\mathbf{x}_{dl}$ demonstrates better fitness than the current learning exemplar, the latter is replaced with $\mathbf{x}_{dl}$. Otherwise, the current learning exemplar remains unchanged and proceeds to the next dimension. This process ensures that the learning exemplar only incorporates dimensions of $\mathbf{g}_{Best}$ that enhance its fitness, guaranteeing no degradation in its quality. The detailed step-by-step procedure is outlined in Algorithm \ref{algo: exemplar DLS}.

Building on the concepts of OLPSO and HCLPSO, the same paper \cite{xu2019particle} introduces a two-swarm learning PSO (TSLPSO) algorithm to balance exploration and exploitation. Like HCLPSO, the population is divided into two sub-populations: one sub-population employs the DLS, while the other adopts the CL mechanism.

\begin{figure}
    \centering 
       \includegraphics[width=0.6\linewidth]{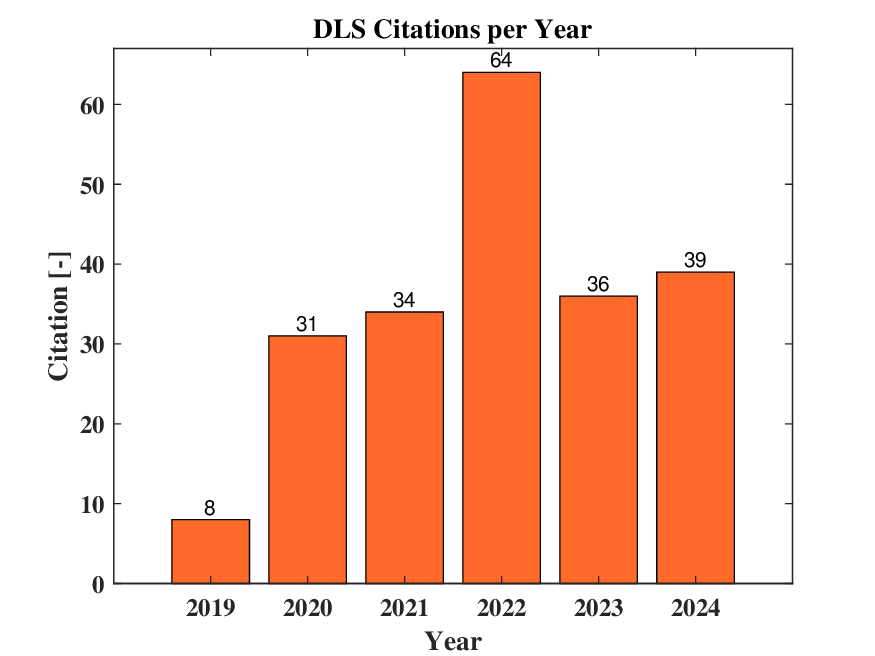} 
    \caption{DLS citations per year based on the Google Scholar platform.}
    \label{fig: DLS citations}
\end{figure}

Based on Google Scholar data, Fig.~\ref{fig: DLS citations} presents the annual citation trends for the DLS from 2019 to 2024. Citations began modestly in 2019 with 8 citations, followed by a steady rise, peaking at 64 in 2022. After this peak, a slight decline was observed, with citations dropping to 36 in 2023 before rebounding to 39 in 2024. This trend highlights the growing recognition of DLS, particularly between 2020 and 2024, with some fluctuations in recent years.

To enhance optimization performance, DLS has been widely integrated into various algorithms. For instance, the DLS mechanism was incorporated into GWO \cite{liu2022dimensional}, where three dominant wolves collaboratively construct an exemplar wolf through DLS to guide the swarm. A multi-strategy DLS approach was proposed to dynamically optimize population structure and enhance diversity during the solving process \cite{wang2025multi}. Additionally, a DLS-based squirrel search algorithm utilizing a roulette strategy was introduced \cite{wen2022dimensional}, where dimensional learning guided squirrels' location updates on oak trees and accurately captured information from hickory trees, improving the efficiency of information flow transmission. In power systems, a DLPSO \cite{wang2021stability} was applied and tested on 13 classical benchmark problems, demonstrating its effectiveness in solving complex optimization tasks. 

\subsection{Adaptive learning strategy}
Li et al. \cite{li2009adaptive,li2011self} proposed an adaptive learning (AL) for PSO, incorporating a variant learning strategy. Each particle learns from four sources: its own best position (\(pbest\)), the closest neighbor’s \(\mathbf{p}_{Best_{nearest}}\), the global best (\(gbest\)), and a random nearby position. This adaptive approach balances exploration and exploitation. The particle update rules are:
\begin{equation*}\begin{cases}
\mathbf{vel}_i^d = \omega_g\cdot \mathbf{vel}_i^d + \eta \cdot r_i^d \cdot (\mathbf{p}_{Best_i}^d - \mathbf{x}_i^d),~ \text{(self-learning)}\\
\mathbf{vel}_i^d = \omega_g\cdot \mathbf{vel}_i^d + \eta \cdot r_i^d \cdot (\mathbf{p}_{Best_{\text{nearest}}}^d - \mathbf{x}_i^d),~\text{(neighbor-learning)} \\
\mathbf{x}_i^d = \mathbf{x}_i^d + \mathbf{vel}_{\text{avg}}^d \cdot N(0,1),~\text{(random search)},~~
\mathbf{vel}_i^d= \omega_g\cdot \mathbf{vel}_i^d + \eta \cdot r_i^d \cdot (\mathbf{g}_{Best}^d - \mathbf{x}_i^d), ~ \text{(global-learning)},
\end{cases}
\end{equation*}
where \(\mathbf{p}_{Best_{\text{nearest}}}\) is the best position of the closest particle, \(\mathbf{vel}_{\text{avg}}\) is the average velocity, and \(N(0,1)\) is a standard normal random variable.

AL employs probability matching to dynamically adjust the selection ratios of the four learning strategies. Initially, each operator has an equal probability (\(1/4\)), updated based on performance. The progress value for operator \(i\) is:
\begin{equation}
\text{prog}_i = \sum_{j=1}^{M_i} f_{\mathbf{x}_{ij}} - \min(f_{\mathbf{x}_{ij}}, f_{\mathbf{x}'_{ij}}),
\end{equation}
where \(\mathbf{x}_{ij}\) and \(\mathbf{x}'_{ij}(t)\) are the parent and child particles, respectively. $M_i$ is the operator's selection times. The reward value is:
\begin{equation}
\text{reward}_i = \exp\left( \alpha_w\cdot\frac{\text{prog}_i}{\sum_{j=1}^{N_s} \text{prog}_j} +  (1 - \alpha_w)\cdot\frac{s_i}{M_i} \right) + c_i\cdot p_i - 1,
\end{equation}
where \(s_i\) counts successful offspring, \(p_i\) is the selection ratio, \(\alpha_w\) is a weight factor, and \(c_i\) is a penalty term. The selection ratio ($p$) is updated every \(g\) generations:
\begin{equation}
p_i = (1 - N_s \cdot \gamma)\cdot \frac{\text{reward}_i}{\sum_{j=1}^{N_s} \text{reward}_j} + \gamma,
\end{equation}
where \(\gamma\) is a minimum selection ratio set to 0.01 in experiments. This framework dynamically optimizes particle learning for improved search efficiency. The AL strategy is further used \cite{wang2011self} to overcome the drawbacks of PSO by incorporating four learning strategies, such as CL, the difference-based velocity updating strategy (DbV), the estimation-based velocity updating strategy (EbV), and CL with $\mathbf{p}_{Best}$. 

Wang et al. \cite{wang2018hybrid} proposed a hybrid PSO with adaptive learning (AL), combining self-learning for exploration and competitive learning for exploitation. A tolerance-based search direction adjustment mechanism (TSDM) is introduced to maintain balance. At each iteration, a particle's current position \(\mathbf{x}_i\) is compared with its best position \(\mathbf{p}_{Best_i}\), updating \(\mathbf{p}_{Best_i}\) if an improvement is found. If no particle improves across the swarm, the condition $\sum_{i=1}^{N_s} (f_{\mathbf{p}_{Best_i}}^t - f_{\mathbf{p}_{Best_i}}^{t-1}) = 0$ holds. To prevent premature convergence, TSDM introduces a tolerance counter \( T \), initialized to zero and incremented by one ($T = T + 1$) when no improvements occur. As \( T \) increases, the likelihood of stagnation rises, prompting an adaptive search direction adjustment. The probability of adjustment, \( prob_{\text{adjust}}= \frac{\exp(T) - 1}{\exp(10) - 1} \), is defined. This mechanism dynamically updates \( prob_{\text{adjust}} \) each iteration, aiding the swarm in escaping local optima. In competitive learning, particles learn from both \(\mathbf{g}_{Best} \) and/or a \textit{Candidate}, generated via Gaussian mutation by selecting two random particles, \(\mathbf{x_k}\) and \(\mathbf{x_m}\).

\subsection{Aging leader and challengers}
Chen et al. \cite{chen2012particle} proposed the aging leader and challengers learning (ALC) for PSO (ALC-PSO), inspired by the natural phenomenon that ``\textit{aging weakens the leader of a colony, providing opportunities for others to challenge the leadership.}" The central concept of ALC is to assign the swarm leader an age and a lifespan, enabling other particles to challenge the leadership when the leader ages. This aging mechanism is a dynamic strategy for promoting leadership transitions, ensuring that the most suitable leader guides the swarm.

The leader's lifespan in ALC is adaptively tuned based on its leading power. Leaders with strong leadership power attract the swarm toward better positions and are granted longer lifespans. Conversely, when a leader fails to improve the swarm, it becomes aged and is replaced by challengers, which introduces diversity and helps the algorithm avoid stagnation and local optima. The influence of the leader (or challenger) on particle velocity is governed by:  
\begin{equation}\label{eq: ALC velocity}
\mathbf{vel}_i^d = \omega \cdot \mathbf{vel}_i^d + c_{e_1} \cdot rand_1^d \cdot (\mathbf{p}_{Best_i}^d - \mathbf{x}_i^d) + c_{e_2} \cdot rand_2^d \cdot (\mathbf{L/C}^d - \mathbf{x}_i^d),
\end{equation}  
where $\mathbf{L/C}$ refers to either the leader ($\mathbf{Leader}_i = \{leader^1_i, leader^2_i, \ldots, leader^d_i\}$) or a challenger ($\mathbf{Challenger}_i$ $ = \{challenger^1_i, challenger^2_i, \ldots, challenger^d_i\}$), depending on their role in the swarm. 

Chaudhary et al. \cite{chaudhary2023modified} applied ALC-PSO to solve task scheduling problems, while Singh et al. used it to address power flow problems with FACTS devices \cite{singh2015particle,singh2015optimal,singh2016particle} and optimal reactive power dispatch. Furthermore, the ALC mechanism has been integrated into differential evolution (DE) algorithms, as demonstrated in \cite{fu2017adaptive,moharam2016design}.

\subsection{Elitist learning strategy}
Lim and Isa \cite{lim2014adaptive} proposed an elitist learning strategy (ELS) for PSO, which incorporates two main components: the OED-based learning strategy (OEDLS), inspired by OLPSO \cite{zhan2009orthogonal,wang2016orthogonal,xiong2018orthogonal}, and the stochastic perturbation-based learning strategy (SPLS), which evolves $\mathbf{g}_{Best}$ when predefined conditions are satisfied. In OEDLS, each $d$th dimension is treated as a factor in a $D$-dimensional problem, resulting in $D$ factors. Two levels ($Q=2$) are defined: global best ($\mathbf{g}_{Best}$) as level 1 and improved personal best ($\mathbf{p}_{Best_i}$) as level 2. Using the $L_M(2^D)$ OA, $M$ candidate solutions $\mathbf{x}_j \, (1 \leq j \leq M)$ are constructed. For each dimension, $\mathbf{g}_{Best}^d$ is selected if the OA level is 1; otherwise, the improved $\mathbf{p}_{Best_i}^d$ is chosen. Factor analysis (FA) is then applied to derive a predictive solution $\mathbf{x}_p$, whose fitness $f_{\mathbf{x}_p}$ is compared to $f_{\mathbf{g}_{Best}}$. If $f_{\mathbf{x}_p} < f_{\mathbf{g}_{Best}}$, $\mathbf{x}_p$ replaces $\mathbf{g}_{Best}$. 

While OEDLS draws inspiration from OLPSO \cite{zhan2009orthogonal}, it differs in its execution. OLPSO activates OL only when a particle's exemplar fails to guide the search, whereas OEDLS is triggered as soon as fitness improvement is detected within the population. This immediate activation allows OEDLS to leverage information from non-global best particles more effectively, further enhancing $\mathbf{g}_{Best}$'s fitness. To prevent $\mathbf{g}_{Best}$ from stagnating in local optima, SPLS introduces a stochastic perturbation to $\mathbf{g}_{Best}$ if its fitness remains unchanged for $m$ consecutive function evaluations (FEs). In SPLS, a randomly selected dimension of $\mathbf{g}_{Best}$, $\mathbf{g}_{Best}^d$, is perturbed using the following equation: \begin{equation}
   \mathbf{g}_{Best,pert}^d = \mathbf{g}_{Best}^d + \text{sgn}(rand_1) \cdot rand_2 \cdot (\mathbf{x}_{\max}^d - \mathbf{x}_{\min}^d), 
\end{equation}where $\text{sgn}()$ is the sign function, $rand_1 \sim U[-1, 1]$, and $rand_2 \sim N(0, R^2)$. Here, $R$ represents the perturbation range, which decreases linearly with the number of FEs. If the fitness of the perturbed particle $\mathbf{g}_{Best, pert}$ is better than $\mathbf{g}_{Best}$, it replaces the latter, ensuring continued improvement. The same author uses the ODELS mechanism in \cite{lim2014bidirectional}.

\subsection{Example-based learning}
An example-based learning (ExL) mechanism \cite{huang2012example} for PSO (ExLPSO) is proposed to address the shortcomings of traditional PSOs by balancing convergence speed and particle diversity. The ExL mechanism is inspired by a common social phenomenon in which learning from multiple elite examples enhances individuals' abilities more effectively than learning from a single example or non-elite individuals. Based on this concept, a PSO utilizing multiple diverse $\mathbf{g}_{Best}$ values can outperform single-$\mathbf{g}_{Best}$ and non-$\mathbf{g}_{Best}$ PSOs in terms of both diversity and efficiency. 

To implement this, a new velocity updating rule is introduced, similar to Eq.~\eqref{eq: CL velocity with gbest}, but with updates to $\mathbf{g}_{Best}$ that allow learning from multiple diverse $\mathbf{g}_{Best}$ values. The velocity update is defined as:

\begin{equation}\label{eq: EL velocity}
\mathbf{vel}_i^d = w_g \cdot \mathbf{vel}_i^d + c_{e_1} \cdot rand_i^d \cdot \left( \mathbf{p}_{Best_{f_i^d}}^d - \mathbf{x}_i^d \right)+ c_{e_2} \cdot rand_i^d \cdot \left( \mathbf{g}_{Best_{g_i^d}}^d - \mathbf{x}_i^d \right),
\end{equation}
Unlike CLPSO, ExLPSO always selected $\mathbf{p}_{Best_{f_i^d}}^d$ with probability $1$, instead of using a learning probability $P_{r_{c}}$.  Furthermore, in contrast to $\mathbf{g}_{Best}$ in PSO, $\mathbf{g}_{Best_{g_i^d}}^d$ in ExLPSO is not always the overall global best. Instead, it is randomly chosen from an example set $\mathbf{E}_s$ of size $B_m$, where $g_i^d$ is a uniformly random integer from $1$ to $B_m$ for each $d$. The construction of the example set $\mathbf{E}_s$ is described in Algorithm~\ref{alg: update_example_set}.

\begin{algorithm}
\caption{Process of updating the example set} \label{alg: update_example_set}
\begin{algorithmic}[1]

\STATE $\mathbf{E}_s \gets \emptyset$ \COMMENT{Initialize the example set}
\IF{$|\mathbf{E}_s| == 0$} 
    \STATE Add the current $\mathbf{g}_{Best}$ to $\mathbf{E}_s$ \COMMENT{Add the first global best to the example set}
\ELSIF{$0 < |\mathbf{E}_s| < B_m$}
    \FOR{each $g \in \mathbf{E}_s$}
        \IF{$f_{\mathbf{g}} > f_{\mathbf{g}_{Best}}$}
            \STATE Add the current $\mathbf{g}_{Best}$ to $\mathbf{E}_s$ \COMMENT{Add $\mathbf{g}_{Best}$ if space is available, and it's better than all examples}
        \ENDIF
    \ENDFOR
\ELSIF{$|\mathbf{E}_s| == B_m$}
    \FOR{each $g \in \mathbf{E}_s$}
        \IF{$f_{\mathbf{g}} > f_{\mathbf{g}_{Best}}$}
            \STATE Replace the oldest $g \in \mathbf{E}_s$ with $\mathbf{g}_{Best}$ \COMMENT{Replace the oldest example using FIFO order if $\mathbf{E}_s$ is full}
        \ENDIF
    \ENDFOR
\ENDIF
\IF{$\exists g \in \mathbf{E}_s$ such that $f_{\mathbf{g}} \leq f_{\mathbf{g}_{Best}}$}
    \STATE $\mathbf{E}_s \gets G$ \COMMENT{No changes made if a better or equal example already exists in $\mathbf{E}_s$}
\ENDIF
\end{algorithmic}
\end{algorithm}

\subsection{Fitness-distance ratio-based learning}\label{subsec: FDR}
Peram et al. \cite{peram2003fitness} introduced a fitness-distance ratio (FDR) mechanism for PSO to address issues caused by crosstalk between particles. In FDR-PSO, each velocity dimension is updated by selecting only one particle at a time. The selected particle satisfies two criteria: (i) it must be near the particle being updated, and (ii) it should have visited a position with higher fitness.

For the \(d\)th dimension of the \(i\)th particle's velocity, the ``nearest best" particle, denoted as \(\mathbf{n}_{Best}\), is chosen based on the best position of another particle (\(\mathbf{p}_{Best_j}\)), maximizing the FDR: \begin{equation}
   \mathbf{FDR}_j^d = \frac{f_{\mathbf{p}_{Best_j}^d} - f_{\mathbf{x}_i^d}}{|\mathbf{p}_{Best_j}^d - \mathbf{x}_i^d|}, 
\end{equation}
 where \(|\cdot|\) denotes the absolute value. The FDR-PSO velocity update is influenced by three factors: (\(\mathbf{p}_{Best}\)), (\(\mathbf{g}_{Best}\)), and (\(\mathbf{n}_{Best}\)). The velocity update equation is given by:

\begin{equation}\label{eq: FDR velocity}
    \mathbf{vel}_i^d = w_{g_0} \cdot \mathbf{vel}_i^d + w_{g_1} \cdot (\mathbf{p}_{Best_i}^d - \mathbf{x}_i^d) + w_{g_2} \cdot (\mathbf{g}_{Best}^d - \mathbf{x}_i^d) + w_{g_3} \cdot (\mathbf{n}_{Best}^d - \mathbf{x}_i^d),
\end{equation}

The coefficients \(w_{g_0}, w_{g_1}, w_{g_2},\) and \(w_{g_3}\) are weights in the velocity update equation. Cleghorn and Engelbrecht \cite{cleghorn2017fitness} provided a theoretical algorithm analysis, offering insights into suitable parameter values. Xiaodong Li \cite{li2007multimodal} highlighted key limitations in FDR calculation: reliance on a particle's frequently changing current position can cause unstable convergence, and using differences between other particles' personal bests and the current position often leads to suboptimal behavior.

\begin{figure}
\centering
\begin{tikzpicture}[>=stealth, scale=1]

\draw[->] (0,1) -- (7,1) node[right] {$x_1$};
\draw[->] (0,1) -- (0,7) node[above] {$x_2$};

\foreach \x in {1,2,...,6} {
    \draw[dashed, gray!50] (\x,1) -- (\x,7);
}
\foreach \y in {1,2,...,6} {
    \draw[dashed, gray!50] (0,\y) -- (7,\y);
}
\foreach \x in {1,2,...,6} {\draw (\x,1) -- (\x,1) node[below] {\x}; 
}
\foreach \y in {1,2,...,6} {
    \draw (0,\y) -- (-0.2,\y) node[left] {\y}; 
}
\coordinate (p1) at (3.3,6.5);
\coordinate (p2) at (1,4);
\coordinate (p3) at (6,2);
\coordinate (pn) at (6,4);
\coordinate (x) at (4,6);
\node[draw, circle, inner sep=2pt, fill=red] at (x) {};
\node[draw, circle, inner sep=2pt, fill=blue] at (p2) {};
\node[draw, circle, inner sep=2pt, fill=black] at (p3) {};
\node at (pn) {\textbullet};

\node[above right] at (p1) {$\mathbf{p}_{Best_1}(4,6)$};
\node[below] at (p2) {$\mathbf{p}_{Best_2}(1,4)$};
\node[below right] at (p3) {$\mathbf{p}_{Best_3}(6,2)$};
\node[above right] at (pn) {$(6,4)$};
\node[above] at (4,6) {$\mathbf{x}_{1}(4,6)$};

\draw[->, red, thick,line width=1.5pt] (x) -- (p2);
\draw[->, blue, thick, dashed,line width=1.5pt] (x) -- (pn);

\draw[dashed,line width=1.5pt] (p2) -- (pn);
\draw[dashed,line width=1.5pt] (p3) -- (pn);
\draw[dashed] (x) -- (pn);

\end{tikzpicture}
\caption{An illustration of computing $\mathbf{n}_{Best}$ using FDR \cite{li2007multimodal}.}\label{Fig: FDR}
\end{figure}
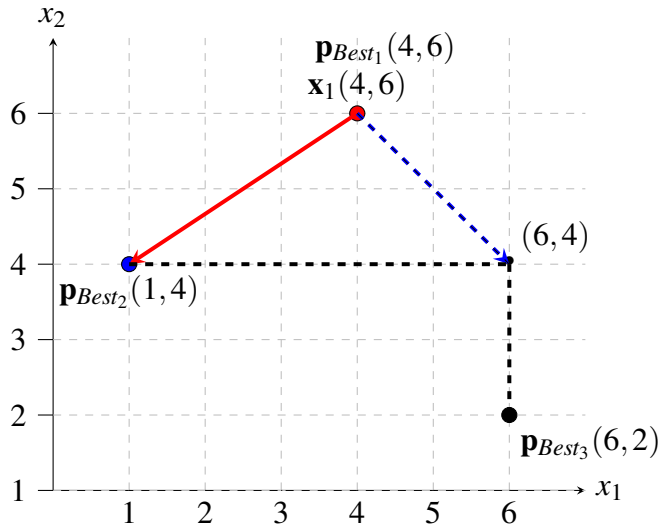

As illustrated in Fig. \ref{Fig: FDR}, these limitations are evident in a 2D search space. Here, \(\mathbf{n}_{Best}\) for \(\mathbf{x}_1\) is calculated using personal bests \(\mathbf{p}_{Best_2}\) and \(\mathbf{p}_{Best_3}\). On dimension 1: 
$
\mathbf{FDR}_2^1 = \frac{1}{|1-4|} = \frac{1}{3}, \quad \mathbf{FDR}_3^1 = \frac{1}{|6-4|} = \frac{1}{2}.
$ On dimension 2: 
$
\mathbf{FDR}_2^2 = \frac{1}{|4-6|} = \frac{1}{2}, \quad \mathbf{FDR}_3^2 = \frac{1}{|2-6|} = \frac{1}{4}.
$
Maximizing the FDR values yields \(\mathbf{n}_{Best}(6,4)\), which is not an ideal attraction point. Despite \(\mathbf{p}_{Best_2}\) and \(\mathbf{p}_{Best_3}\) being equally fit, \(\mathbf{x}_1\) converges to \(\mathbf{n}_{Best}(6,4)\) instead of either promising point.

To address this issue, Li \cite{li2007multimodal} introduced the fitness-Euclidean distance ratio (FER), defined as:
\[
\mathbf{FER}_j^d = \alpha_{fer} \cdot \frac{f_{\mathbf{p}_{Best_j}^d} - f_{\mathbf{p}_{Best_i}^d}}{||\mathbf{p}_{Best_j}^d - \mathbf{p}_{Best_i}^d||},
\]
where \(\alpha_{fer}\) is a scaling factor, and \(||\cdot||\) represents Euclidean distance. The particle velocity update follows Eq. \eqref{eq: FDR velocity} but excludes the global best term or FIPS \cite{mendes2004fully} procedure. The FER mechanism is further used in DE \cite{liang2014differential} and ABC \cite{ming2020balanced}.

\subsection{Scatter learning}
Ren et al. \cite{ren2013scatter} proposed a scatter learning (ScL) mechanism for PSO, which introduces an exemplar pool ($\mathbf{EP}$) containing high-quality, diverse solutions scattered across the search space. Particles select their exemplars from EP using the roulette wheel rule. The $\mathbf{EP}$ is defined as:

\begin{equation}\label{eq: exemplar_pool}
\mathbf{EP} = \{\mathbf{x}_1, \mathbf{x}_2, \dots, \mathbf{x}_m \, | \, f_{\mathbf{x}_1} \geq f_{\mathbf{x}_2} \geq \dots \geq f_{\mathbf{x}_m};~ \forall \mathbf{x}_i, ~\mathbf{x}_j \neq i,~ (\mathbf{x}_i, \mathbf{x}_j) \geq \gamma\},
\end{equation}
where \(m\) is the cardinality of $\mathbf{EP}$, \((\cdot, \cdot)\) denotes the distance between solutions, and \(\gamma\) ensures a minimum distance to maintain diversity. Solutions are added to $\mathbf{EP}$ based on fitness and spatial separation: \(\mathbf{x}_1\) is the best solution, \(\mathbf{x}_2\) is the second-best satisfying \((\mathbf{x}_1, \mathbf{x}_2) \geq \gamma\), and so on. Any solution \(\mathbf{x}'\) violating \((\mathbf{x}_1, \mathbf{x}') < \gamma\) is excluded. This ensures EP includes diverse and high-quality exemplars.

Particles in ScL employ one exemplar at a time to avoid oscillations. The velocity update rule is given by:

\begin{equation}\label{eq: velocity_update}
\mathbf{vel}_i^d= \omega _g \cdot \mathbf{vel}_i^d + c\cdot rand^d \cdot (\mathbf{x}_{e_i}^d - \mathbf{x}_i^d),
\end{equation}
where \(\mathbf{x}_{e_i}^d\) is the \(d\)th element of the current exemplar employed by particle \(i\). Each exemplar in $\mathbf{EP}$ is exploited based on a selection probability, ensuring better exemplars have a greater influence on particle behavior while maintaining diverse search patterns.

\subsection{Informed-learning}

Clerc and Kennedy \cite{clerc2002particle} simplified the PSO velocity update equation, i.e., Eq. \eqref{eq: pso_velocity}, by normalizing the personal and global best positions ($\mathbf{p}_{Best}$ and $\mathbf{g}_{Best}$), yielding:
\begin{equation}
    \mathbf{vel}_i^d = \mathcal{X} \cdot \left(\mathbf{vel}_i^d + c_{e}'\cdot (\overline{\mathbf{p}_{Best_i}^d} - \mathbf{x}_i^d) \right),
\end{equation}
where $c_{e}' = c_{e_1} + c_{e_2}$, $\mathcal{X}$ is the constriction weight, and $\overline{\mathbf{p}_{Best}}$ is the normalized best position given as: $ \overline{\mathbf{p}_{Best_i}^d} = \frac{c_{e_1} \cdot \mathbf{p}_{Best_i}^d + c_{e_2} \cdot \mathbf{g}_{Best}^d}{c_{e_1} + c_{e_2}}.$ This method eliminates the need to balance velocity adjustments between two terms, ensuring proper convergence behavior.

Mendes et al. \cite{mendes2004fully} introduced an alternative approach by incorporating a fully informed learning strategy for PSO (FIPS), where each particle utilizes information from all neighboring particles instead of a single best neighbor. The updated $\overline{\mathbf{p}_{Best}}$ is computed as:
\begin{equation}
    \overline{\mathbf{p}_{Best_i}} = \frac{\sum_{j=1}^{n_g} \psi_k \cdot c_{e_k}' \cdot \mathbf{p}_{Best_j}}{\sum_{j=1}^{n_g} \psi_k \cdot c_{e_k}'}, \quad c_{e_k}' = U[0, c_{e_{max}}'/n_g],
\end{equation}
where $\mathbf{p}_{Best_j}$ represents the best position of neighborhood particle $j$, and $n_g$ is the number of neighboring particles. FIPS has been tested with five PSO topologies, \textit{all}, \textit{ring}, \textit{four clusters}, \textit{pyramid}, and \textit{square} \cite{lynn2018population}, demonstrating improved performance in global optimization by enhancing convergence speed and accuracy.

However, FIPS struggles in multimodal optimization since its topology-based neighborhood selection may cause particles to converge toward a single peak instead of exploring multiple optima. For instance, in a ring topology, particles interact only with adjacent indices, making it likely that neighbors belong to different niches, thereby reducing the effectiveness of peak identification \cite{qu2012distance}.

To address this limitation, Qu et al. \cite{qu2012distance} proposed locally informed learning for PSO (LIPS), which, like FIPS, incorporates neighbor information but selects neighbors based on Euclidean distance rather than fixed topology. The $\overline{\mathbf{p}_{Best}}$ in LIPS is computed as:
\begin{equation}
    \overline{\mathbf{p}_{Best_i}} = \frac{\sum_{j=1}^{n_g} \psi_k \cdot \mathbf{n}_{Best_j}/n_g}{\sum_{j=1}^{n_g} \psi_k},
\end{equation}
where $\psi_k$ is a random number in the range $[0, 4.1/n_g]$. Unlike topology-based PSO, LIPS enhances exploration by adapting to the spatial distribution of particles while preserving niche integrity during exploitation. This allows LIPS to effectively locate multiple peaks with high accuracy, making it well-suited for multimodal optimization \cite{qu2012distance}.

\subsection{Other learning strategies}

Li et al.~\cite{li2023reinforcement} proposed an RL-based PSO incorporating a dynamic oscillation inertia weight for adaptive adjustment, a reinforcement learning-based velocity update for exploration-exploitation balance, cosine similarity-based velocity control for improved solution selection, and a local update strategy to enhance diversity and mitigate premature convergence. Qiu et al.~\cite{qiu2023q} developed a Q-learning-based PSO, where Q-learning adaptively selects the optimal multi-exemplar strategy to maximize long-term rewards. An elite learning strategy improves local search and balances exploration and exploitation. Hein et al.~\cite{hein2017particle} introduced a fuzzy PSO with RL, constructing fuzzy RL policies by training on simulated world models derived from past system transitions. This approach integrates self-organizing fuzzy controllers with model-based batch RL. It is suitable for domains where online learning is impractical, system dynamics are modelable, and interpretable policies are required. Multi-strategy RL has been incorporated into PSO in~\cite{meng2023multi,hou2023multi}.

Tang et al.~\cite{tang2011feedback} proposed a feedback learning mechanism (FLM) for PSO, where learning probabilities are dynamically adjusted based on a fitness feedback mechanism. The acceleration coefficients (\(c_{e_1}\) and \(c_{e_2}\)) are determined by fitness feedback and generation number, enhancing adaptability. To refine the solution further, elite stochastic disturbances are introduced to the globally best particle under adaptive probabilities, ensuring controlled exploration. The learning mechanism focuses on a single dimension of the global best particle, targeting specific refinements while maintaining computational efficiency. Gong et al.~\cite{gong2015genetic} proposed a genetic learning scheme for PSO (GL-PSO), integrating genetic operators to construct exemplars. The crossover operator leverages historical particle information (\(\mathbf{p}_{Best}\) and \(\mathbf{g}_{Best}\)) to generate high-quality offspring, while mutation enhances global exploration by injecting diversity. The selection operator ensures directional evolution across generations, leading to improved search performance.

Sabat et al.~\cite{sabat2011integrated} proposed an integrated learning (IL) approach for PSO to optimize complex multimodal functions. A particle updating strategy based on the hyperspherical coordinate system was introduced, effectively addressing evenly distributed multiple minima. This approach was combined with a CL strategy to enhance exploration. Niu et al.~\cite{niu2013multi} introduced a center learning strategy, selecting the center position or the best-found position within each swarm based on learning probability. Dong et al.~\cite{dong2014autonomous} proposed an autonomous learning adaptation (ALA) method for PSO, where each particle self-adjusts its control parameters based on past successes and failures. For successful movements, parameters increase in the forward direction while others decrease, and the opposite adjustments are applied for unsuccessful movements.

Chen et al.~\cite{chen2017biogeography,chen2020biogeography} developed a biogeography-based learning strategy (BLS) for PSO, where each particle updates using a combination of its own \(\mathbf{p}_{Best}\) and the \(\mathbf{p}_{Best}\) of others via biogeography migration. BLS generates an exemplar vector index \( \mathbf{f}_i = [ f_i^1, f_i^2, \dots, f_i^d] \), similar to the CL mechanism, based on migration and particle ranks. To counter stagnation, Dong et al.~\cite{dong2018reverse} introduced a reverse learning strategy (RLS) for PSO, where, upon \(\mathbf{g}_{Best}\) remaining unchanged for a predefined number of iterations, particle velocities are updated as:
\begin{equation}
\mathbf{vel}_i^d = w_g \cdot \mathbf{vel}_i^d + c_{e_1} \cdot rand_1 \cdot (\mathbf{x}_i^d - \mathbf{x}_{worst_i}^d) + c_{e_2} \cdot rand_2 \cdot (\mathbf{x}_k^d - \mathbf{x}_i^d),
\end{equation}
where \(\mathbf{x}_{worst_i}\) represents the historically worst position of particle \(i\), and \(\mathbf{x}_k\) is a randomly selected particle from a low-fitness niche. Tanweer et al.~\cite{tanweer2015self} introduced two learning strategies: a self-regulating inertia weight for the best particle to enhance exploration and a self-perception mechanism for others to improve exploitation. Inspired by bee-foraging behavior, Chen et al.~\cite{chen2021bee} proposed a bee-foraging learning PSO, incorporating three learning phases similar to those in the artificial bee colony (ABC) algorithm.

\begin{table}[]
    \centering
    \renewcommand{\arraystretch}{1.2}
      \caption{Single population-based learning strategies for PSO.}\label{tab: Single population-based}
\resizebox{1\linewidth}{!}{
\begin{tabular}{llccp{9cm}}\hline
Long name	&	Sort name	&	Year	&	Ref.	&	Comments	\\\hline
Comprehensive learning PSO	&	CLPSO	&	2006	&	\cite{liang2006comprehensive}	&	CLPSO is a PSO variant where particles are not guided by \(\mathbf{g}_{Best}\) but instead choose between their \(\mathbf{p}_{Best}\) or a neighbor’s \(\mathbf{p}_{Best}\) with a probability \(P_{r_c}\), introducing stochasticity that enhances diversity but slows convergence. The CLPSO boundary handling method is retained despite concerns about its effectiveness. Control parameters are set as \(c_{e_1} = 1.49445\), \(\omega\) linearly decreases from 0.9 to 0.2, and $R_g=5$. 	\\
Orthogonal learning PSO	&	OLPSO	&	2009	&	\cite{zhan2009orthogonal}	&	The OL strategy combines the personal best position \(\mathbf{p}_{Best}\) and the neighborhood best position \(\mathbf{n}_{Best}\) to form a guidance vector \(\mathbf{o}_{Best}\). The particle updates its velocity as: $\mathbf{vel}_i^d = w_g \cdot \mathbf{vel}_i^d + c_e \cdot rand_1^d \cdot \left( \mathbf{o}_{Best}^d - \mathbf{x}_i^d \right).$ The guidance vector \(\mathbf{o}_{Best}\), constructed using the OED method, is given by: $\mathbf{o}_{Best} = \mathbf{p}_{Best} \oplus \mathbf{n}_{Best}$.	\\
Dimensional learning strategy PSO	&	DLSPSO	&	2019	&	\cite{xu2019particle}	&	The DLSPSO effectively discovers and preserves the promising information of the population's best solution and constructs a learning exemplar for each particle. The learning exemplar is constructed by allowing each dimension of the $\mathbf{p}_{Best}$ to learn from the corresponding $\mathbf{g}_{Best}$ on a dimension-by-dimension basis to construct a learning exemplar $\mathbf{x}_{dl}$, as: $\mathbf{vel}_i^d = w_g \cdot \mathbf{vel}_i^d + c_{e_1} \cdot rand_i^d \cdot \left( \mathbf{x}_{dl_i}^d - \mathbf{x}_i^d \right)+ c_{e_2} \cdot rand_i^d \cdot \left( \mathbf{g}_{Best}^d - \mathbf{x}_i^d \right).$ This process ensures that the learning exemplar only incorporates dimensions of $\mathbf{g}_{Best}$ that enhance its fitness, guaranteeing no degradation in its quality.	\\
Adaptive learning PSO	&	ALPSO	&	2009	&	\cite{li2009adaptive,li2011self}	&	In ALPSO, each particle learns from four sources: its own best position (\(pbest\)), the closest neighbor’s \(\mathbf{p}_{Best_{nearest}}\), the global best (\(gbest\)), and a random nearby position. This adaptive approach balances exploration and exploitation. ALPSO employs probability matching to dynamically adjust the selection ratios of the four learning strategies.	\\
Aging leader and challengers learning PSO	&	ALCPSO	&	2012	&	\cite{chen2012particle}	&	In this variant, a leader particle replaces \(\mathbf{g}_{Best}\) for velocity updates. Initially, \(\mathbf{g}_{Best}\) serves as the leader, but it ages over time and may be replaced by a competing particle to prevent stagnation in local optima. The central concept of ALC is to assign the swarm leader an age and a lifespan, enabling other particles to challenge the leadership when the leader ages. This aging mechanism is a dynamic strategy for promoting leadership transitions, ensuring that the most suitable leader guides the swarm, as: $\mathbf{vel}_i^d = \omega \cdot \mathbf{vel}_i^d + c_{e_1} \cdot rand_1^d \cdot (\mathbf{p}_{Best_i}^d - \mathbf{x}_i^d) + c_{e_2} \cdot rand_2^d \cdot (\mathbf{L/C}^d - \mathbf{x}_i^d).$ Velocities are initially set to zero and restricted to 50\% of the search bounds. A rebounding technique is applied as in \cite{helwig2012experimental}. Control parameters are \(c_{e_1} = c_{e_2} = 2\), \(\omega = 0.4\), initial leader lifespan = 60, \(T = 2\), and \(pro = 1/d\).	\\
Elitist learning strategy	&	ELPSO	&	2014	&	\cite{lim2014adaptive}	&	EL combines an OED-based learning strategy, inspired by OLPSO \cite{zhan2009orthogonal,wang2016orthogonal,xiong2018orthogonal}, and SPLS, which adapts $\mathbf{g}_{Best}$ under predefined conditions. Unlike OLPSO, OL is triggered only when an exemplar fails, OED activates immediately upon fitness improvement, utilizing non-global best particles to enhance $\mathbf{g}_{Best}$.  To prevent stagnation, SPLS applies a stochastic perturbation to $\mathbf{g}_{Best}$ if its fitness remains unchanged for $m$ consecutive FEs. A randomly selected dimension $\mathbf{g}_{Best}^d$ is perturbed as:  $\mathbf{g}_{Best,pert}^d = \mathbf{g}_{Best}^d + \text{sgn}(rand_1) \cdot rand_2 \cdot (\mathbf{x}_{\max}^d - \mathbf{x}_{\min}^d).$	\\\hline
\end{tabular}}
\end{table}

\begin{table}[]
    \centering
    \renewcommand{\arraystretch}{1.2}
      \caption{Continue Table\ref{tab: Single population-based}.}\label{tab: Single population-based1}
\resizebox{1\linewidth}{!}{
\begin{tabular}{llccp{9cm}}\hline
Long name	&	Sort name	&	Year	&	Ref.	&	Comments	\\\hline
Example-based learning PSO	&	ExLPSO	&	2012	&	\cite{huang2012example}	&	The ExL mechanism mimics the social phenomenon where learning from multiple elite examples enhances performance more effectively than relying on a single or non-elite individual. To achieve this, a velocity update rule is introduced, modifying $\mathbf{g}_{Best}$ to incorporate multiple diverse exemplars. The velocity update is given by:  $\mathbf{vel}_i^d = w_g \cdot \mathbf{vel}_i^d + c_{e_1} \cdot rand_i^d \cdot ( \mathbf{p}_{Best_{f_i^d}}^d - \mathbf{x}_i^d) + c_{e_2} \cdot rand_i^d \cdot ( \mathbf{g}_{Best_{g_i^d}}^d - \mathbf{x}_i^d ).$ This formulation enables particles to learn from multiple elite sources, promoting diverse exploration and improved optimization.  	\\
Fitness-distance ratio-learning based PSO	&	FDR-PSO	&	2003	&	\cite{peram2003fitness}	&	In FDR-PSO, each velocity dimension is updated by selecting a single nearby particle that has visited a higher-fitness position. The velocity update is influenced by three factors: (\(\mathbf{p}_{Best}\)), (\(\mathbf{g}_{Best}\)), and (\(\mathbf{n}_{Best}\)). Xiaodong Li \cite{li2007multimodal} identified key limitations in FDR calculation, including instability due to reliance on frequently changing positions and the potential for suboptimal behavior when using differences between other particles' personal bests and the current position.	\\
Scatter learning PSO	&	ScLPSO	&	2013	&	\cite{ren2013scatter}	&	ScL introduces an exemplar pool (\(\mathbf{EP}\)) with high-quality, diverse solutions distributed across the search space. Particles select exemplars using a roulette wheel rule and employ one at a time to prevent oscillations. The velocity update rule is:  $\mathbf{vel}_i^d= \omega_g \cdot \mathbf{vel}_i^d + c \cdot rand^d \cdot (\mathbf{x}_{e_i}^d - \mathbf{x}_i^d).$	\\
Q-learning PSO	&	QLPSO	&	2023	&	\cite{qiu2023q}	&	Q-learning to optimize multi-exemplar selection for long-term rewards, integrating elite learning for exploration-exploitation balance.	\\
Feedback learning PSO	&	FLPSO	&	2011	&	\cite{tang2011feedback}	&	The acceleration coefficients (\( c_{e_1} \) and \( c_{e_2} \)) adapt based on fitness feedback and generation number. Elite stochastic disturbances refine the global best particle with adaptive probabilities, focusing on one dimension for targeted improvement while maintaining efficiency.	\\
Genetic learning PSO	&	GLPSO	&	2015	&	\cite{gong2015genetic}	&	GLPSO combines PSO with a composite exemplar (\(\mathbf{p}_{Best}\), \(\mathbf{g}_{Best}\)) and genetic operators (crossover, mutation) running in parallel. Crossover exploits historical data for high-quality offspring, mutation enhances diversity, and selection ensures directional evolution. Velocity is capped at 20\% of the bounds span, with randomized initial values. A rebounding technique is applied~\cite{gong2015genetic}. Control parameters follow~\cite{gong2015genetic}: \(c_{e_1} = 1.49618\), \(\omega_g = 0.7298\),$P_m =$ 0.01, and a stopping gap of 7, after which tournament selection (20\% of the population) is used.  	\\
Integrated learning PSO	&	ILPSO	&	2011	&	\cite{sabat2011integrated}	&	A hyperspherical coordinate-based particle update strategy was introduced to efficiently handle evenly distributed multiple minima. It was further integrated with a CL strategy to improve exploration.	\\
Autonomous learning PSO	&	ALPSO	&	2014	&	\cite{dong2014autonomous}	&	Each particle, acting as an intelligent agent, adapts its parameters based on past successes and failures. Successful movements reinforce forward-directed parameters while reducing others, whereas unsuccessful movements trigger the opposite adjustments.	\\
Biogeography-based learning PSO	&	BLSPSO	&	2017	&	\cite{chen2017biogeography,chen2020biogeography}	&	Each particle updates using its own $\mathbf{p}_{Best}$ and those of others via biogeography migration. BLS constructs an exemplar vector \( f_i = [ f_i^1, f_i^2, \dots, f_i^d] \) based on migration and particle ranks, similar to CL.	\\\hline
\end{tabular}}
\end{table}

\begin{table}[]
    \centering
    \renewcommand{\arraystretch}{1.2}
      \caption{Continue Table\ref{tab: Single population-based1}.}\label{tab: Single population-based2}
\resizebox{1\linewidth}{!}{
\begin{tabular}{llccp{10cm}}\hline
Long name	&	Sort name	&	Year	&	Ref.	&	Comments	\\\hline
Reverse learning PSO	&	RLSPSO	&	2018	&	\cite{dong2018reverse}	&	When $\mathbf{g}_{Best}$ stagnates for a predefined number of iterations, a RLS for PSO is proposed, updating particle velocity as: $ \mathbf{vel}_i^d = w_g \cdot \mathbf{vel}_i^d + c_{e_1} \cdot rand_1 \cdot (\mathbf{x}_i^d - \mathbf{x}_{worst_i}^d) + c_{e_2} \cdot rand_2 \cdot (\mathbf{x}_k^d - \mathbf{x}_i^d).$	\\
Self-regulating PSO	&	SRPSO	&	2015	&	\cite{tanweer2015self}	&	Two learning strategies for PSO: a self-regulating inertia weight for the best particle to enhance exploration and a self-perception mechanism for others to improve exploitation.	\\
Bee-foraging learning PSO	&	BFLPSO	&	2021	&	\cite{chen2021bee}	&	Inspired by bee-foraging behavior, Chen et al. \cite{chen2021bee} proposed a bee-foraging learning PSO, incorporating three learning phases similar to those in ABC.	\\
Informed-learning PSO & FIPS & 2004& \cite{mendes2004fully}& Provide an alternate way to calculate $\overline{\mathbf{p}_{Best}}$ by normalized $\mathbf{p}_{Best}$ and $\mathbf{g}_{Best}$. FIPS has been tested with five PSO topologies, \textit{all}, \textit{ring}, \textit{four clusters}, \textit{pyramid}, and \textit{square} \cite{lynn2018population}, demonstrating improved performance in global optimization by enhancing convergence speed and accuracy.\\\hline
\end{tabular}}
\end{table}

\section{Two swarm-based leaning}\label{sec: two-swarm}
In this section, we discuss learning strategies based on a two-population framework for PSO. Unlike single-swarm approaches, where the entire population follows a unified learning strategy, two-population PSO divides the swarm into two distinct groups, each employing single/different learning mechanisms with different parameters. This division enables a more flexible search process, where one population can focus on global exploration while the other prioritizes local exploitation, improving the algorithm’s ability to escape local optima and enhance convergence accuracy.

The interaction between the two populations can be cooperative or competitive. In cooperative strategies, both populations exchange information periodically to enhance search diversity and prevent stagnation. In contrast, competitive approaches maintain independent evolution, where a superior population influences the other through selection mechanisms or dominance-based interactions. Additionally, hybrid methods integrate different learning paradigms, such as reinforcement learning, adaptive parameter control, or exemplar-based knowledge transfer, to optimize search efficiency.

By leveraging complementary learning strategies, two-population PSO offers improved adaptability in dynamic and complex optimization landscapes. The following subsections explore various two-population learning strategies, their mechanisms, and their impact on balancing exploration and exploitation.
\subsection{Social-learning strategy}
Inspired by the social learning mechanism \cite{de2010incremental, daneshyari2010cultural}, Cheng and Jin \cite{cheng2015social} proposed a social learning-based PSO (SLPSO), where followers (particles with lower fitness values) learn from exhibitors (particles with higher fitness values). The position of a follower is updated as:

\begin{equation}\label{eq: SL position}
\mathbf{x}_i^d = 
\begin{cases}
    \mathbf{x}_i^d + \mathbf{vel}_i^d, & \text{if } rand_i \leq p_{r_i}^L, \\
    \mathbf{x}_i^d, & \text{otherwise},
\end{cases}
\end{equation}
where \(p_{r_i}^L\) is the learning probability inspired by natural social learning. If the random value \(rand_i\) satisfies \(0 \leq rand_i \leq p_{r_i}^L\), the follower learns from exhibitors, and its velocity is updated as:

\begin{equation}\label{eq: SL velocity}
\mathbf{vel}_i^d = rand_1^d \cdot \mathbf{vel}_i^d + rand_2^d \cdot \left( \mathbf{x}_k^d - \mathbf{x}_i^d \right) + rand_3^d \cdot \epsilon \cdot \left( \overline{\mathbf{x}}^d - \mathbf{x}_i^d \right),
\end{equation}
where \(rand_1^d, rand_2^d, rand_3^d\) are random values, \(\epsilon\) is a social influence factor, \(\mathbf{x}_k^d\) is the \(d\)th dimension of the exhibitor, and \(\overline{\mathbf{x}}^d\) is the mean position of all particles. 

To update the velocity of each follower, the population is first sorted by their fitness values. Each follower \(i\) learns from a randomly selected exhibitor \(k\), where \(i < k \leq N_s\). For a population of \(N_s\) particles sorted in ascending order of their fitness values as \(\mathbf{x}_1, \mathbf{x}_2, \ldots, \mathbf{x}_{N_s}\), the learning hierarchy is defined as follows: \(\mathbf{x}_1\) (the worst particle) considers \(\mathbf{x}_2, \mathbf{x}_3, \ldots, \mathbf{x}_{N_s}\) as its potential exhibitors. In contrast, \(\mathbf{x}_{N_s-1}\) learns only from \(\mathbf{x}_{N_s}\) (the best particle). The best particle \(\mathbf{x}_{N_s}\) has no exhibitor and is, therefore, not updated. Conversely, the worst particle \(\mathbf{x}_1\) cannot be an exhibitor for any other particle due to its low fitness. The SL strategy has been hybridized with exemplar learning and dynamic mutation \cite{zhang2019differential}. Exemplar learning replaces the follower and social influence components of SLPSO, while a dynamic differential mutation strategy using three \(\mathbf{p}_{Best}\) enhances exploration. Yu et al. \cite{yu2019generation} proposed a generation-based optimal restart strategy using radial basis functions for surrogate-assisted SLPSO. Zhao et al. \cite{zhao2022evolutionary} developed a fuzzy classification-based SLPSO to improve rule initialization and fitness function evaluation. A three-learning PSO \cite{zhang2022three} is proposed to overcome the limitations of SLPSO, such as poor searchability and low search efficiency, by intelligently incorporating three learning strategies. Additionally, the SL strategy was extended by dividing the population into two sub-groups, losers and winners, leading to the competitive swarm optimizer \cite{cheng2014competitive}, where losers learn from winners rather than followers learn from exhibitors.

 \subsection{Interswarm interactive learning}
Inspired by social learning behavior, an interswarm interactive learning (IIL) strategy \cite{qin2015particle} is proposed for PSO. In PSO, the fitness value of $\mathbf{g}_{Best}$ is monitored, and if it remains unchanged for \( k \) consecutive iterations, i.e.,  
\begin{equation}
f_{\mathbf{g}_{Best}^t} - f_{\mathbf{g}_{Best}^{t-k}} = 0.
\end{equation}  
The algorithm is assumed to be trapped in a local optimum, triggering the IIL strategy. Two swarms are then formed similarly to SL: one retains its original learning strategy (learned swarm), while the other (learning swarm) adapts. The softmax function determines the probability of a swarm being selected as the learned swarm.

Each particle in the learning swarm follows a learning probability \( P_{c_i} \), defined as:
\begin{equation}
P_{c_i} = 0.1 + 0.5 \cdot \left(\frac{R_i}{N_s}\right)^5,
\end{equation}
where \( R_i \) is the rank of the \( i \)th particle based on fitness, and \( N_s \) is the swarm size. The learned swarm follows the standard PSO update rule, whereas particles in the learning swarm switch between two learning modes: (1) retaining their original search strategy or (2) learning from the learned swarm. Assuming $Swarm_1$ is the learning swarm and $Swarm_2$ is the learned swarm, the velocity update follows:
\begin{equation}
\mathbf{vel}_{i,1}^d = \omega_g\cdot \mathbf{vel}_{i,1}^d +
\begin{cases}
    c_{e_1}'\cdot r_1 (\mathbf{p}_{Best_{i,1}}^d - \mathbf{x}_{i,1}^{d}) + c_{e_2}'\cdot  r_2(\mathbf{p}_{nBest,1}^{d} - \mathbf{x}_{i,1}^d) + c_{e_3} \cdot r_3 (\mathbf{p}_{gBest,2}^{d} - \mathbf{x}_{i,1}^{d}), & \text{if } r_4 \leq P_{c_i}, \\
    c_{e_1}\cdot r_5 (\mathbf{p}_{Best_{i,1}}^{d} - \mathbf{x}_{i,1}^d) + c_{e_2}\cdot r_6 (\mathbf{p}_{nBest,1}^{d} - \mathbf{x}_{i,1}^d), & \text{otherwise}.
\end{cases}
\end{equation}
Here, \( c'_1, c'_2 \) differ from \( c_1, c_2 \), and \( c_3 \) controls the influence of interswarm learning, $r_j,~j=1,2,\ldots,5$ are randomly generated numbers. This IIL strategy enhances diversity and improves the probability of escaping local optima. 

\subsection{Crisscross and stochastic example learning}
To enhance the performance of PSO and overcome its limitations, Liang et al.~\cite{liang2021hybrid} introduced a stochastic example learning strategy (SEL) integrated with crisscross learning. In this approach, the population is divided into two sub-populations with a ratio \(\lambda_r\): an elite sub-swarm and a following sub-swarm. The elite sub-swarm guides the learning process of the following sub-swarm while simultaneously exploring information from other particles to enhance diversity. 

The learning strategy for the elite sub-swarm is based on crisscross learning~\cite{meng2014crisscross}, where the personal bests of different particles are used as horizontal and vertical crossovers. This mechanism allows the elite sub-swarm to leverage its own and other particles' experiences, improving exploration and diversity. For the following sub-swarm, SEL defines the velocity updating rule as:

\begin{equation} \label{eq: SEL_velocity}
\mathbf{vel}_i^d = w_g \cdot \mathbf{vel}_i^d + c_1 \cdot rand_1^d \cdot \left( rand_2 \cdot \mathbf{p}_{Best_i}^d + (1 - rand_2) \cdot \mathbf{p}_{Best_k}^d - \mathbf{x}_i^d \right),
\end{equation}
where \(k\) is randomly selected from an example pool \([1, i-1]\) of superior particles. The SEL strategy enables particles in the following sub-swarm to learn from superior individuals, improving diversity and enhancing their ability to exploit promising regions. This approach reduces the risk of premature convergence, fosters population diversity, and accelerates convergence. 

\subsection{Teaching and peer-learning} 
A teaching and peer-learning (TPL) strategy is introduced in PSO \cite{lim2014teaching}, consisting of two phases: the teaching phase and the peer-learning phase. In the teaching phase, a particle updates its velocity following the standard PSO rule (Eq. \eqref{eq: pso_velocity}). If the particle fails to improve its fitness in this phase, it enters the peer-learning phase (inspired by TLBO \cite{rao2012teaching}), where it selects an exemplar particle (\(\mathbf{p}_{Best_e}\)) from the personal best positions of peer particles (\(\mathbf{p}_{Best}\)), excluding its own (\(\mathbf{p}_{Best_i}\)) and the global best particle (\(\mathbf{g}_{Best}\)). This dual-phase strategy enhances adaptability, improving the balance between exploration and exploitation. To improve the $\mathbf{g}_{Best}$ by fully exploiting the valuable information of each particle, Lim and Isa \cite{lim2014bidirectional} used the ODELS mechanism in TPL for PSO. Many recent PSO variants lack alternative learning strategies when particles fail to improve fitness. To address this, a teaching–learning-based optimization (TLBO) framework was adapted into PSO, resulting in the teaching and peer-learning PSO (TPLPSO) algorithm. 

\subsection{Neighbor-based learning}
Cao et al. \cite{cao2018neighbor} proposed a neighbor-based learning PSO, where each particle (target particle) learns from two swarm members: a randomly selected neighbor $(\mathbf{x}_k)$ and the global best particle \((\mathbf{g}_{Best} )\), as:
\begin{equation}\label{eq: velocity neighborhood leaning}
\mathbf{vel}_i^d =
\begin{cases}
\omega_g\cdot \mathbf{vel}_i^d + F_{DE}\cdot (\mathbf{x}_k^d - \mathbf{x}_i^d) + c_e \cdot \text{rand} \cdot (\mathbf{g}_{Best}^d - \mathbf{x}_i^d), & \text{if } \text{rand} < L_{p_r}, \\
\mathbf{v}_i^d, & \text{otherwise}.
\end{cases}
\end{equation}
here, \( F_{DE} \) is a scaling factor from DE and \( L_{p_r} \) is the learning probability. Nasir et al. \cite{nasir2012dynamic} proposed DNLPSO introduces a learning strategy (Eq. \eqref{eq: CL velocity with gbest}, where parameters $w_g$ varies 0.9 to 0.4 and $c_{e_1}=c_{e_2}=1.49445$ are selected) where a particle's velocity is updated using the historical best information of particles within a dynamic neighborhood, rather than the entire population. These neighborhoods are reformed periodically, allowing participants to learn from their neighborhood or historical best practices, thus enhancing guidance and maintaining diversity. 
 \begin{equation}\label{eq: CL velocity with gbest}
\mathbf{vel}_i^d = w_g \cdot \mathbf{vel}_i^d + c_{e_1} \cdot rand_1^d \cdot \left( \mathbf{p}_{Best_{f_i^d}}^d - \mathbf{x}_i^d \right)+ c_{e_2} \cdot rand_2^d \cdot \left( \mathbf{g}_{Best}^d - \mathbf{x}_i^d \right),
\end{equation}
Additionally, $P_{r_{c}}$ adopts a linear variation within the same range (Eq. \eqref{eq: DNL probability}), with the intent of enhancing each particle's learning tendency compared to the exponential variation. 

\subsection{Strategy adaptation learning}
An ensemble learning approach is proposed in \cite{lynn2017ensemble} for PSO, termed ensemble PSO (EPSO), which integrates multiple learning strategies into a unified framework. In EPSO, the population is divided into two groups: the first group comprises 40\% of the swarm, while the second group consists of the remaining 60\%. Five learning strategies are incorporated into EPSO:  
(i) CLPSO with $\mathbf{g}_{Best}$ \cite{liang2006comprehensive}, defined in Eq. \eqref{eq: CL velocity with gbest} with the CLPSO probability $P_{c_r}$, (ii) linearly decreasing $\omega_g$ PSO (LDWPSO) \cite{kennedy1995particle},  
(iii) distance-based locally informed PSO (LIPS) \cite{qu2012distance}, (iv) PSO with time-varying acceleration coefficients (TVAC-PSO) \cite{ratnaweera2004self}, (v) fitness distance ratio-based PSO \cite{peram2003fitness}, discussed in Section \ref{subsec: FDR}.  

The first group consistently employs CLPSO with $\mathbf{g}_{Best}$, while the second group dynamically selects one of the five strategies based on past performance. A self-adaptive selection mechanism is used to determine the most effective learning strategy by tracking the success and failure of each approach over a defined learning period $L_p$. The selection probability for each strategy is computed as:

\begin{equation}
    p_{k}(g) = \frac{S_{k}(g)}{\sum_{k=1}^{K} S_{k}(g)},\quad S_{k}(g) = \frac{\sum_{g=G-L_p}^{G-1} n_{s_k}(g)}{ \left( \sum_{g=G-L_p}^{G-1} (n_{s_k}(g) + n_{f_k}(g)) + \varepsilon \right)},
\end{equation}
here, $k = 1, 2, \ldots, K$ ($K$ is number of strategies), $G > L_p$, and $S_{k}(g)$ represents the success rate of the $k$th learning strategy. The small constant $\varepsilon = 0.01$ prevents division by zero and ensures numerical stability. At each generation after $L_p$, the success rate of each learning strategy is updated, and strategies are selected proportionally based on their performance. This self-adaptive mechanism allows EPSO to dynamically adjust the learning strategy for each particle, enhancing its ability to balance exploration and exploitation in different optimization landscapes.

Inspired by EPSO \cite{lynn2017ensemble}, Liu and Nishi \cite{liu2022strategy} proposed a strategy dynamics learning (SDL) approach for PSO, where the population is also divided into two groups: 30\% and 70\% of the population. Four learning strategies, such as CLPSO \cite{liang2006comprehensive}, unified PSO (UPSO) \cite{parsopoulos2019upso}, LDWPSO \cite{kennedy1995particle}, and LIPS \cite{qu2012distance}, are integrated into a single framework with probabilistic selection.  The first group consistently employs CL learning, while the second group adopts one of the four strategies based on a roulette wheel selection mechanism. However, the selection process may lead to rapid convergence, causing one strategy to dominate others. A restart operator is introduced to maintain strategy diversity, dividing the optimization process into \( r \) stages based on the maximum number of fitness evaluations. When entering a new stage, the probability distribution is reset as \( \mathbf{x}_h = 1/k \) for \( h \in \{1,2,\ldots,k\} \).

\subsection{Other learning strategies}
 Wang et al. \cite{wang2020adaptive} introduced an adaptive granularity learning (AGL) distributed PSO that incorporates machine learning techniques, such as locality-sensitive hashing and logistic regression for clustering and adaptive granularity control, respectively. The method uses a master-slave multi-subpopulation model, where subpopulations co-evolve and share evolutionary information. This dynamic adjustment of subpopulation sizes balances exploration and exploitation in large search spaces, enhancing optimization performance. Panda et al. \cite{panda2016static} proposed a static learning strategy (SLS) to vary swarm size in PSO adaptively. The population is divided into two sub-populations, where particles interact within their neighborhoods, and survival probability ($S_{p_i}=f_i/\sum_i f_i$) determines the existence of better particles. PSO is divided into two phases: the first enhances exploration, while the second optimizes the utilization of explored information. 
 
 A cooperative-based difference learning (CDL) strategy \cite{li2023cooperative} for PSO is proposed, where particles are divided into elite and common subswarms based on fitness, dynamically adapting learning strategies to enhance exploration. The elite subswarm selectively updates particles using superior information while preventing premature convergence. The velocity update rule is:  
\begin{equation}
    \mathbf{vel}_i^d =\omega_g \cdot \mathbf{vel}_i^d +c_{e_1} \cdot rand_1 \cdot (\mathbf{p}_{Best_{r_m}} - \mathbf{x}_i^d)+ c_{e_2} \cdot rand_2 \cdot (\mathbf{p}_{Best_{r_s}} - \mathbf{x}_i^d),
\end{equation}  
where $\mathbf{p}_{Best_{r_m}}$ and \(\mathbf{p}_{Best_{r_s}}\) represent the best historical positions of randomly selected particles from the elite and common subswarms, respectively. The common subswarm adopts stage-based strategies to improve convergence accuracy over time. Xia et al. \cite{xia2020expanded} proposed multi-exemplar-based forgetting learning (MFL) for PSO, extending the social learning process from one to two exemplars: the local best (\(\mathbf{l}_{Best_i}\)) and global best (\(\mathbf{g}_{Best}\)). Additionally, each particle is assigned a forgetting ability vector \(\mathbf{f}_{g_i} = \{f_{g_i}^1, f_{g_i}^2, \dots, f_{g_i}^d\} \) to regulate information retention, mimicking human forgetting behavior. The velocity update rule is:  
\begin{equation}
\mathbf{vel}_i = w_g \cdot \mathbf{vel}_i + c_{e_1} \cdot r_{1} \cdot (\mathbf{p}_{Best_i} - \mathbf{x}_i) + c_{e_2} \cdot r_{2} \cdot ((1 - f_{g_i}) \cdot \mathbf{l}_{Best_i} - \mathbf{x}_i) + c_{e_3} \cdot r_{3} \cdot ((1 - f_{g_i}) \cdot \mathbf{g}_{Best} - \mathbf{x}_i),
\end{equation}
where \( c_{e_1}, c_{e_2},\) and \(c_{e_3} \) are acceleration coefficients for personal, local, and global bests, respectively, and \( f_{g_i} \) controls information loss when learning from \(\mathbf{l}_{Best_i}\) and \(\mathbf{g}_{Best}\). The forgetting effect is distance-dependent, meaning particles farther apart experience greater information loss. Multilevel and multi-sample learning strategies are proposed in \cite{li2022pyramid,tian2024diversity,yang2023multi}.

\begin{table}[]
    \centering
    \renewcommand{\arraystretch}{1.2}
      \caption{Two-population-based learning strategies for PSO.}\label{tab: Two population-based}
\resizebox{1\linewidth}{!}{
\begin{tabular}{llccp{9cm}}\hline
Long name	&	Sort name	&	Year	&	Ref.	&	Comments	\\\hline
Social learning PSO	&	SLPSO	&	2015	&	\cite{cheng2015social}	&	Social learning occurs when followers (particles with lower fitness) learn from exhibitors (particles with higher fitness). The population is sorted by fitness, and each follower \(i\) learns from a randomly selected exhibitor \(k\), where \(i < k \leq N_s\). For a population of \(N_s\) particles sorted in ascending fitness order as \(\mathbf{x}_1, \mathbf{x}_2, \dots, \mathbf{x}_{N_s}\), the hierarchy is:  The worst particle \(\mathbf{x}_1\) learns from any \(\mathbf{x}_k\) where \(2 \leq k \leq N_s\).  The second-worst \(\mathbf{x}_2\) has fewer exhibitor choices, and so on.  The best particle \(\mathbf{x}_{N_s}\) has no exhibitor and remains unchanged.  The worst particle \(\mathbf{x}_1\) cannot be an exhibitor. SL has poor searchability and low search efficiency.  	\\
Interswarm interactive learning PSO	&	IILPSO	&	2015	&	\cite{qin2015particle}	&	In IIL, two swarms are formed similarly to SL: one retains its original learning strategy (learned swarm), while the other (learning swarm) adapts. The softmax function determines the probability of a swarm being selected as the learned swarm. The learned swarm follows the standard PSO update rule, whereas particles in the learning swarm switch between two learning modes: (1) retaining their original search strategy or (2) learning from the learned swarm. The IIL strategy enhances diversity and improves the probability of escaping local optima. 	\\
Adaptive granularity learning PSO	&	AGLPSO	&	2020	&	\cite{wang2020adaptive}	&	AGL incorporates machine learning, including locality-sensitive hashing for clustering and logistic regression for adaptive granularity. A master-slave multi-subpopulation model enables co-evolution and information sharing, dynamically balancing exploration and exploitation for improved optimization.	\\
Stochastic example learning PSO	&	SELPSO	&	2021	&	\cite{liang2021hybrid}	&	The population is split into an elite and a following sub-swarm in a ratio \(\lambda_r\). The elite guides the followers while exploring diverse information using crisscross learning~\cite{meng2014crisscross}, leveraging personal bests as horizontal and vertical crossovers. The following sub-swarm updates velocity using: $\mathbf{vel}_i^d = w_g \cdot \mathbf{vel}_i^d + c_1 \cdot rand_1^d \cdot \left( rand_2 \cdot \mathbf{p}_{Best_i}^d + (1 - rand_2) \cdot \mathbf{p}_{Best_k}^d - \mathbf{x}_i^d \right).$ This approach reduces the risk of premature convergence, fosters population diversity, and accelerates convergence. 	\\
Teaching and peer-learning PSO	&	TPLPSO	&	2014	&	\cite{lim2014teaching}	&	It has teaching and peer-learning phases. In the teaching phase, a particle updates velocity using the standard PSO rule (Eq.~\eqref{PSO velocity}). If fitness fails to improve, it enters the peer-learning phase~\cite{rao2012teaching}, selecting an exemplar (\(\mathbf{p}_{Best_e}\)) from peer personal bests (\(\mathbf{p}_{Best}\)), excluding its own (\(\mathbf{p}_{Best_i}\)) and the global best (\(\mathbf{g}_{Best}\)). This dual-phase strategy enhances adaptability, improving the balance between exploration and exploitation.	\\
Static learning PSO	&	SLSPSO	&	2016	&	\cite{panda2016static}	&	The population is split into two sub-populations, where particles interact locally, and survival probability ($S_{p_i}=f_i/\sum_i f_i$) determines superior particles. PSO operates in two phases: the first emphasizes exploration, while the second refines the utilization of discovered information.	\\\hline
\end{tabular}}
\end{table}

\begin{table}[]
    \centering
    \renewcommand{\arraystretch}{1.2}
      \caption{Continue Table\ref{tab: Two population-based}.}\label{tab: Two population-based1}
\resizebox{1\linewidth}{!}{
\begin{tabular}{llccp{9cm}}\hline
Long name	&	Sort name	&	Year	&	Ref.	&	Comments	\\\hline
Difference learning PSO	&	DLPSO	&	2023	&	\cite{li2023cooperative}	&	Particles are categorized into elite and common subswarms based on fitness, dynamically adjusting learning strategies for better exploration. The elite subswarm leverages superior information while mitigating premature convergence. The velocity update is:  $ \mathbf{vel}_i^d =\omega_g \cdot \mathbf{vel}_i^d +c_{e_1} \cdot rand_1 \cdot (\mathbf{p}_{Best_{r_m}} - \mathbf{x}_i^d)+ c_{e_2} \cdot rand_2 \cdot (\mathbf{p}_{Best_{r_s}} - \mathbf{x}_i^d).$	\\
Neighbor-based learning PSO	&	NLPSO	&	2018	&	\cite{cao2018neighbor}	&	Each target particle learns from two swarm members: a randomly selected neighbor $(\mathbf{x}_k)$ and the global best particle \((\mathbf{g}_{Best} )\) as mentioned in Eq. \eqref{eq: velocity neighborhood leaning}. Nasir et al. \cite{nasir2012dynamic} proposed dynamic neighborhood learning PSO introduces a learning strategy in Eq. \eqref{eq: CL velocity with gbest}.	\\
Ensemble learning PSO	&	EPSO	&	2017	&	\cite{lynn2017ensemble}	&	The population is divided into 40\% and 60\% groups. A unified framework integrates CLPSO with $\mathbf{g}_{Best}$ \cite{liang2006comprehensive}, TVAC-PSO \cite{ratnaweera2004self}, FDR-PSO \cite{peram2003fitness}, LDWPSO \cite{kennedy1995particle}, and LIPS \cite{qu2012distance} with probabilistic selection \( p_{k} = S_{k}\big/\sum_{k=1}^{K} S_{k} \) for \( k \in \{1,2,\ldots,K\}\).	\\
Strategy dynamic learning PSO	&	SDPSO	&	2022	&	\cite{liu2022strategy}	&	The population is split into 30\% and 70\% groups. A unified framework integrates CLPSO \cite{liang2006comprehensive}, UPSO \cite{parsopoulos2019upso}, LDWPSO \cite{kennedy1995particle}, and LIPS \cite{qu2012distance} with probabilistic selection. At each stage transition, probabilities reset as \( \mathbf{x}_h = 1/k \) for \( h \in \{1,2,\ldots,k\} \).	\\
Forgetting learning PSO	&	MFLPSO	&	2020	&	\cite{xia2020expanded}	&	It extends social learning to two exemplars: \(\mathbf{l}_{Best_i}\) and \(\mathbf{g}_{Best}\), A forgetting vector \(\mathbf{f}_{g_i} = \{f_{g_i}^1, f_{g_i}^2, \dots, f_{g_i}^d\} \) regulates information retention, mimicking human forgetting. The velocity update rule is $\mathbf{vel}_i = w_g \cdot \mathbf{vel}_i + c_{e_1} \cdot r_{1} \cdot (\mathbf{p}_{Best_i} - \mathbf{x}_i) + c_{e_2} \cdot r_{2} \cdot ((1 - f_{g_i}) \cdot \mathbf{l}_{Best_i} - \mathbf{x}_i) + c_{e_3} \cdot r_{3} \cdot ((1 - f_{g_i}) \cdot \mathbf{g}_{Best} - \mathbf{x}_i)$.	\\\hline
\end{tabular}}
\end{table}

\section{Multiple swarm-based leaning}\label{sec: multiple-swarm}
This section discusses multi-population PSO, where the swarm is divided into multiple subpopulations, each with distinct learning strategies. This approach enhances diversity, prevents premature convergence, and improves optimization performance in complex search spaces. Subpopulations may evolve independently, interact periodically to exchange information, or follow hierarchical structures for adaptive learning. Cooperation-based strategies allow knowledge sharing among subpopulations, while competition-based methods promote exploration by maintaining diversity. Additionally, dynamic population control mechanisms adjust subpopulation sizes or strategies based on performance metrics.

Multi-population PSO is particularly effective for multimodal and large-scale optimization problems. The following subsections explore key techniques and their impact on search efficiency.
\subsection{Dynamic multi-swarm strategy}\label{sec: DMS}
The dynamic multi-swarm strategy (DMS), first proposed for PSO \cite{liang2005dynamic}, enhances the standard PSO by addressing its limitations in tackling complex multimodal optimization problems. This method builds on the local version of PSO and introduces a novel neighborhood topology characterized by two key features: small neighborhoods and a randomized regrouping mechanism.

DMS-PSO employs small neighborhoods, in contrast to traditional PSO approaches that often favor larger populations. By dividing the population into smaller swarms, DMS-PSO slows down the convergence velocity and enhances the diversity of solutions. Each swarm searches independently within the solution space using its members' historical information (Eq. \eqref{eq: DMS velocity}, where $\mathbf{l}_{Best_i}=\{l_{Best_i}^1,l_{Best_i}^2,\ldots,l_{Best_i}^d\}$), allowing the algorithm to efficiently explore and exploit the search space. This approach is particularly beneficial for complex problems, where smaller neighborhoods are better suited to navigating intricate landscapes and avoiding premature convergence. 
\begin{equation}\label{eq: DMS velocity}
\mathbf{vel}_i^d = w_g \cdot \mathbf{vel}_i^d + c_{e_1} \cdot rand_i^d \cdot \left( \mathbf{p}_{Best_i}^d - \mathbf{x}_i^d \right)+ c_{e_2} \cdot rand_i^d \cdot \left( \mathbf{l}_{Best_i}^d - \mathbf{x}_i^d \right).
\end{equation}

To further mitigate the risk of stagnation or premature convergence to local optima, DMS-PSO incorporates a randomized regrouping mechanism. After every $R_p$ generation (the regrouping period), the population is randomly reorganized into new configurations of small swarms. This process promotes information exchange across swarms, leveraging the best solutions discovered in different regions of the search space while also increasing diversity by disrupting static neighborhood structures. The value of $R_p$ plays a pivotal role in the effectiveness of this strategy: a well-chosen $R_p$ ensures meaningful interaction between sub-populations without hindering convergence or exploration. This regrouping achieves two critical objectives:
\begin{enumerate}[(i)]
    \item \textit{Information exchange:} It allows swarms to share the valuable information accumulated during their search, thereby promoting collective intelligence across the population.
\item \textit{Increased diversity:} By restructuring neighborhoods, the algorithm disrupts stagnation and provides fresh opportunities for exploration, helping it escape local optima.
\end{enumerate}

\begin{figure}
	\begin{center}
		\begin{tikzpicture}[node distance=1.5cm,auto]
		\tikzset{->-/.style={decoration={
					markings,
					mark=at position #1 with {\arrow{>}}},postaction={decorate}}}
		
		\draw[gray,very thick,fill=gray!20] (0,7) -- (4,7)-- (4,10)--(0,10)--(0,7);

		\filldraw[purple] (0.5,9.5) circle (4pt) node[anchor=center](1)at (0.5,9.5){\textbf{\textcolor{black}{}}};
		\filldraw[purple] (2.5,9.5) circle (4pt) node[anchor=center](2)at (2.5,9.5){\textbf{\textcolor{black}{}}};
		\filldraw[purple] (1,8.5) circle (4pt) node[anchor=center](3)at (1,8.5){\textbf{\textcolor{black}{}}};
		\draw[dashed,black,thick](1)--(2);
		\draw[dashed,black,thick](2)--(3);
		\draw[dashed,black,thick](3)--(1);
		
		\filldraw[blue!70] (1.5,9.2) circle (4pt) node[anchor=center](4)at (1.5,9.2){\textbf{\textcolor{white}{}}};
		\filldraw[blue!70] (3.5,9.2) circle (4pt) node[anchor=center](5)at (3.5,9.2){\textbf{\textcolor{white}{}}};
		\filldraw[blue!70] (3,7.5) circle (4pt) node[anchor=center](6)at (3,7.5){\textbf{\textcolor{white}{}}};
		\draw[dashed,black,thick](4)--(5);
		\draw[dashed,black,thick](5)--(6);
		\draw[dashed,black,thick](6)--(4);
		
		\filldraw[cyan!70] (3,9) circle (4pt) node[anchor=center](7)at (3,9){\textbf{\textcolor{white}{}}};
		\filldraw[cyan!70] (3.6,7.2) circle (4pt) node[anchor=center](8)at (3.6,7.2){\textbf{\textcolor{white}{}}};
		\filldraw[cyan!70] (1,7.2) circle (4pt) node[anchor=center](9)at (1,7.2){\textbf{\textcolor{white}{}}};
		\draw[dashed,black,thick](7)--(8);
		\draw[dashed,black,thick](8)--(9);
		\draw[dashed,black,thick](9)--(7);
		
		\filldraw[green!70] (2,7.5) circle (4pt) node[anchor=center](10)at (2,7.5){\textbf{\textcolor{white}{}}};
		\filldraw[green!70] (0.5,8) circle (4pt) node[anchor=center](11)at (0.5,8){\textbf{\textcolor{white}{}}};
		\filldraw[green!70] (1.6,8.4) circle (4pt) node[anchor=center](12)at (1.6,8.4){\textbf{\textcolor{white}{}}};
		\draw[dashed,black,thick](10)--(11);
		\draw[dashed,black,thick](11)--(12);
		\draw[dashed,black,thick](12)--(10);
\draw[>= triangle 45,->,thick,line width=2pt](4,8.5)--(6,8.5);

\draw[gray,very thick,fill=gray!20] (6,7) -- (6,10)-- (10,10)-- (10,7)-- (6,7);

\filldraw[purple] (6.5,9.5) circle (4pt) node[anchor=center](13)at (6.5,9.5){\textbf{\textcolor{black}{}}};
\filldraw[purple] (7.5,9.5) circle (4pt) node[anchor=center](14)at (7.5,9.5){\textbf{\textcolor{black}{}}};
\filldraw[purple] (7,8.5) circle (4pt) node[anchor=center](15)at (7,8.5){\textbf{\textcolor{black}{}}};
\draw[dashed,black,thick](13)--(14);
\draw[dashed,black,thick](14)--(15);
\draw[dashed,black,thick](15)--(13);

\filldraw[green!70] (8,9) circle (4pt) node[anchor=center](16)at (8,9){\textbf{\textcolor{white}{}}};
\filldraw[green!70] (9,9.5) circle (4pt) node[anchor=center](17)at (9,9.5){\textbf{\textcolor{white}{}}};
\filldraw[green!70] (9.5,8.5) circle (4pt) node[anchor=center](18)at (9.5,8.5){\textbf{\textcolor{white}{}}};
\draw[dashed,black,thick](16)--(17);
\draw[dashed,black,thick](17)--(18);
\draw[dashed,black,thick](18)--(16);

\filldraw[cyan!70] (6.5,8) circle (4pt) node[anchor=center](19)at (6.5,8){\textbf{\textcolor{white}{}}};
\filldraw[cyan!70] (7.5,7.5) circle (4pt) node[anchor=center](20)at (7.5,7.5){\textbf{\textcolor{white}{}}};
\filldraw[cyan!70] (6.2,7.2) circle (4pt) node[anchor=center](21)at (6.2,7.2){\textbf{\textcolor{white}{}}};
\draw[dashed,black,thick](19)--(20);
\draw[dashed,black,thick](20)--(21);
\draw[dashed,black,thick](21)--(19);

\filldraw[blue!70] (8,8) circle (4pt) node[anchor=center](22)at (8,8){\textbf{\textcolor{white}{}}};
\filldraw[blue!70] (8.3,7.4) circle (4pt) node[anchor=center](23)at (8.3,7.4){\textbf{\textcolor{white}{}}};
\filldraw[blue!70] (9.5,7.8) circle (4pt) node[anchor=center](24)at (9.5,7.8){\textbf{\textcolor{white}{}}};
\draw[dashed,black,thick](22)--(23);
\draw[dashed,black,thick](23)--(24);
\draw[dashed,black,thick](24)--(22);
\draw[>= triangle 45,->,thick,line width=2pt](8,7)--(8,5);
\node[arrow,right,brown]at(8.2,6){\textbf{Regrouping}};

\draw[gray,very thick,fill=gray!20] (6,2) --  (6,5) --  (10,5)-- (10,2)-- (6,2);

\filldraw[cyan!70] (6.2,3.3) circle (4pt) node[anchor=center](25)at (6.2,3.3){\textbf{\textcolor{black}{}}};
\filldraw[purple] (6.9,4.3) circle (4pt) node[anchor=center](26)at (6.9,4.3){\textbf{\textcolor{black}{}}};
\filldraw[blue!70] (8.5,4.8) circle (4pt) node[anchor=center](27)at (8.5,4.8){\textbf{\textcolor{black}{}}};
\draw[dashed,black,thick](25)--(26);
\draw[dashed,black,thick](26)--(27);
\draw[dashed,black,thick](27)--(25);

\filldraw[green!70] (6.3,4) circle (4pt) node[anchor=center](28)at (6.3,4){\textbf{\textcolor{white}{}}};
\filldraw[blue!70] (7,3) circle (4pt) node[anchor=center](29)at (7,3){\textbf{\textcolor{white}{}}};
\filldraw[purple] (8,4) circle (4pt) node[anchor=center](30)at (8,4){\textbf{\textcolor{white}{}}};
\draw[dashed,black,thick](28)--(29);
\draw[dashed,black,thick](29)--(30);
\draw[dashed,black,thick](30)--(28);

\filldraw[cyan!70] (7.5,4.8) circle (4pt) node[anchor=center](31)at (7.5,4.8){\textbf{\textcolor{white}{}}};
\filldraw[purple] (7.5,2.5) circle (4pt) node[anchor=center](32)at (7.5,2.5){\textbf{\textcolor{white}{}}};
\filldraw[green!70] (9.5,3.5) circle (4pt) node[anchor=center](33)at (9.5,3.5){\textbf{\textcolor{white}{}}};
\draw[dashed,black,thick](31)--(32);
\draw[dashed,black,thick](32)--(33);
\draw[dashed,black,thick](33)--(31);

\filldraw[blue!70] (8.5,3.5) circle (4pt) node[anchor=center](34)at (8.5,3.5){\textbf{\textcolor{white}{}}};
\filldraw[green!70] (6.3,2.4) circle (4pt) node[anchor=center](35)at (6.3,2.4){\textbf{\textcolor{white}{}}};
\filldraw[cyan!70] (9,2.4) circle (4pt) node[anchor=center](36)at (9,2.4){\textbf{\textcolor{white}{}}};
\draw[dashed,black,thick](34)--(35);
\draw[dashed,black,thick](35)--(36);
\draw[dashed,black,thick](36)--(34);
\draw[>= triangle 45,->,thick, line width=2pt](6,3.5)--(4,3.5);

\draw[gray,very thick,fill=gray!20] (0,2) -- (0,5) --  (4,5)--  (4,2)-- (0,2);

\filldraw[cyan!70] (0.5,4.5) circle (4pt) node[anchor=center](37)at (0.5,4.5){\textbf{\textcolor{black}{}}};
\filldraw[purple] (1.5,4) circle (4pt) node[anchor=center](38)at (1.5,4){\textbf{\textcolor{black}{}}};
\filldraw[green!70] (0.5,3.5) circle (4pt) node[anchor=center](39)at (0.5,3.5){\textbf{\textcolor{black}{}}};
\draw[dashed,black,thick](37)--(38);
\draw[dashed,black,thick](38)--(39);
\draw[dashed,black,thick](39)--(37);

\filldraw[blue!70] (2.5,4.3) circle (4pt) node[anchor=center](40)at (2.5,4.3){\textbf{\textcolor{white}{}}};
\filldraw[green!70] (3,3.5) circle (4pt) node[anchor=center](41)at (3,3.5){\textbf{\textcolor{white}{}}};
\filldraw[cyan!70] (3.5,4) circle (4pt) node[anchor=center](42)at (3.5,4){\textbf{\textcolor{white}{}}};
\draw[dashed,black,thick](40)--(41);
\draw[dashed,black,thick](41)--(42);
\draw[dashed,black,thick](42)--(40);

\filldraw[cyan!70] (1.5,3.2) circle (4pt) node[anchor=center](43)at (1.5,3.2){\textbf{\textcolor{white}{}}};
\filldraw[blue!70] (1,2.2) circle (4pt) node[anchor=center](44)at (1,2.2){\textbf{\textcolor{white}{}}};
\filldraw[purple] (2.5,2.5) circle (4pt) node[anchor=center](45)at (2.5,2.5){\textbf{\textcolor{white}{}}};
\draw[dashed,black,thick](43)--(44);
\draw[dashed,black,thick](44)--(45);
\draw[dashed,black,thick](45)--(43);

\filldraw[green!70] (2.3,3.3) circle (4pt) node[anchor=center](46)at (2.3,3.3){\textbf{\textcolor{white}{}}};
\filldraw[purple] (3.5,3.2) circle (4pt) node[anchor=center](47)at (3.5,3.2){\textbf{\textcolor{white}{}}};
\filldraw[blue!70] (3.4,2.2) circle (4pt) node[anchor=center](48)at (3.4,2.2){\textbf{\textcolor{white}{}}};
\draw[dashed,black,thick](46)--(47);
\draw[dashed,black,thick](47)--(48);
\draw[dashed,black,thick](48)--(46);
		\end{tikzpicture}
	\end{center}
	\caption{The DMS learning.}\label{fig: DMS strategy}
\end{figure}
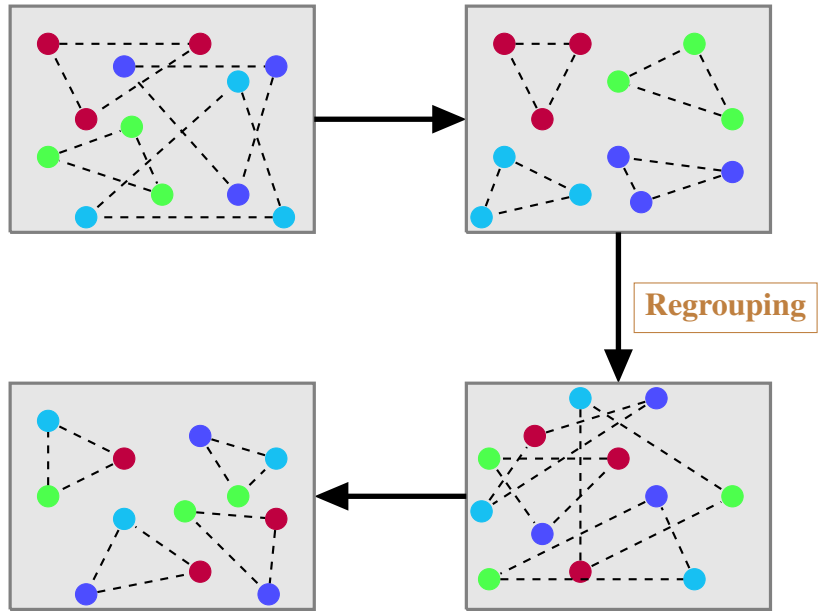

As illustrated in Fig.~\ref{fig: DMS strategy}, consider a population of 12 particles divided into four sub-populations, each with three individuals. Initially, each sub-population focuses on refining its local search. After $R_p$  generations, the population is randomly reorganized into new groups, allowing swarms to exchange valuable information and continue exploring the search space from different configurations.

The dynamic nature of DMS-PSO, with its small-sized swarms and randomized regrouping schedule, grants it greater flexibility and adaptability compared to classical PSO variants. These innovations allow DMS-PSO to efficiently balance exploration and exploitation, making it particularly effective in handling complex, multimodal optimization problems.

\begin{figure}
    \centering
    \includegraphics[width=1\linewidth,height=8cm]{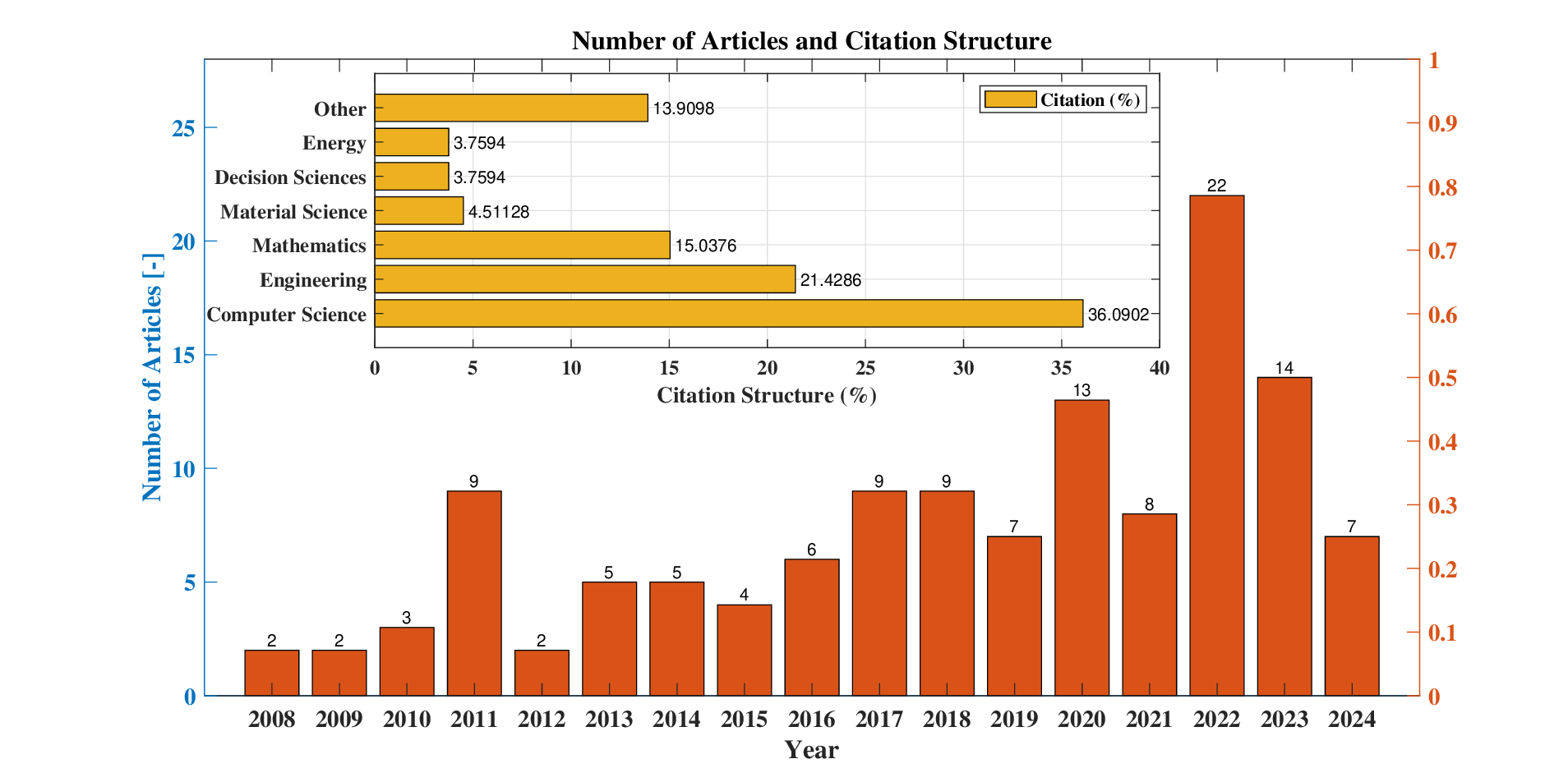}  
    \caption{Number of annually published documents on DMS and application in the different fields ($\%$) based on the Scopus platform.}
    \label{fig: DMS articles}
\end{figure}

\begin{figure}
    \centering 
       \includegraphics[width=0.6\linewidth]{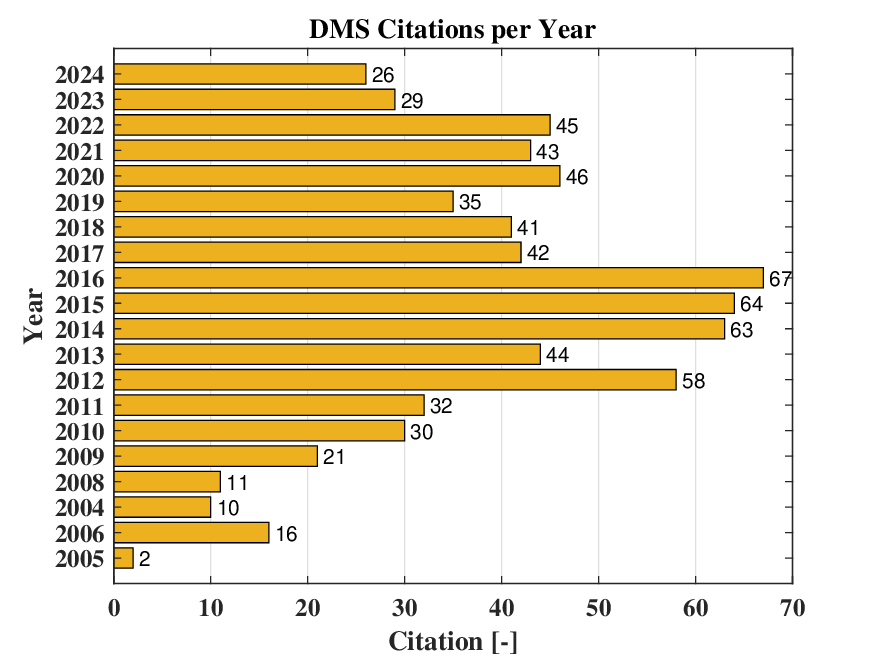} 
    \caption{DMS citations per year based on the Google Scholar platform.}
    \label{fig: DMS citations}
\end{figure}

Fig. \ref{fig: DMS articles} illustrates the annual publication trends and citation distribution of research documents on DMS and its applications across various fields, as analyzed through the Scopus platform. The bottom chart shows the annual published documents from 2008 to 2024. A gradual increase in publications is evident, starting with only two papers in 2008. Notable growth is observed after 2017, peaking in 2022 with 22 publications. This highlights a surge in research interest and developments in DMS during this period. However, a slight decline is visible post-2022, with 14 publications in 2023 and 7 in 2024 (as of now), potentially indicating a stabilization phase or a shift in research focus. The top chart provides insights into the citation structure across various fields. Computer Science leads significantly, contributing 36.09$\%$ of the total citations, showcasing its critical role in developing and implementing DMS algorithms. Engineering follows with 21.43$\%$, reflecting the widespread application of DMS in solving engineering optimization problems. Mathematics accounts for 15.04$\%$ of citations, emphasizing its importance in theoretical advancements and algorithmic foundations. Other fields, including Material Science, Decision Sciences, and Energy, contribute to a smaller but noteworthy share of research, collectively representing the interdisciplinary applications of DMS. The annual citations of the original DMS paper \cite{liang2006comprehensive}, as per Google Scholar, as shown in Fig.~\ref{fig: DMS citations}, average 36 citations per year.

Chen et al. \cite{chen2018dynamic} introduced DMSDL-PSO by integrating differential mutation and the Quasi-Newton method into DMS-PSO, enhancing exploration and exploitation capabilities. The differential mutation was applied to personal historical best particles, and a modified velocity update equation addressed PSO's limitations.  Liu et al. \cite{liu2022dynamic} extended this idea to Harris Hawks optimization for cascade hydropower dispatch. Similarly, local search operators and cooperative learning strategies have been incorporated into DMS-PSO \cite{liang2005local, zhao2008dynamic, xu2015dynamic}, further enhancing its performance. Wang et al. \cite{wang2020heterogeneous} integrated a CL mechanism into DMS-PSO by dividing the population into two groups, a strategy later adapted to the gravitational search algorithm (GSA) \cite{chauhan2023optimizing}. Yousri et al. \cite{yousri2021parameters} combined DMS learning with the marine predator algorithm for parameter identification in fuel cells.

Additional extensions include integrating DMS-PSO with parallel PC cluster systems \cite{fan2010dynamic} to enhance computational efficiency. Tao et al. \cite{tao2022fitness} proposed a fitness peak clustering-based DMS-PSO with a CL strategy, where stagnant particles merge into a global swarm for further evolution. Tang et al. \cite{tang2019dynamic} and Xia et al. \cite{xia2020dynamic} developed DMS-GPSO, where the global sub-swarm focuses on exploitation under the guidance of the optimal particle in the population. At the same time, other sub-swarms prioritize exploration guided by local best positions. Store and reset operators were also introduced to save computational resources and increase diversity. Yang et al. \cite{yang2023adaptive} proposed an adaptive DMS-PSO incorporating a stagnation detection mechanism (SDM) and spatial exclusion strategy. When a sub-swarm's local best solution stagnates before reaching the regrouping period, SDM generates a vitality particle to aid in discovering promising areas.

The DMS learning framework has been integrated into various optimization algorithms, including GSA \cite{chauhan2023optimizing}, bird swarm algorithm \cite{zhang2020dynamic}, fireworks algorithm \cite{lei2023dynamic}, pigeon-inspired algorithm \cite{tang2021dynamic}, K-means clustering \cite{yu2015dynamic}, Nelder-Mead Simplex method \cite{zhong2013hybrid}, firefly algorithm (FA) \cite{chang2023hybrid}, harmonic search (HS) \cite{zhao2011dynamic}, and whale optimization algorithm (WOA) \cite{miao2025dynamic}. These integrations demonstrate the versatility and adaptability of the DMS learning strategy across diverse optimization problems.

Balancing diversity and convergence speed is a persistent challenge in PSO. While DMS  significantly enhances diversity by dividing the population into smaller sub-swarms and regrouping them periodically, it sacrifices convergence speed and exploitation capability due to frequent regrouping. To address these limitations, Liang et al. \cite{liang2005local, zhao2008dynamic} introduced the Quasi-Newton method as a local search algorithm to improve precision in promising regions. Periodically, every $t$ generation, sub-populations are ranked by fitness, and the top 25$\%$ of groups have their $\mathbf{l}_{Best_i}$ refined using the Quasi-Newton method. Additionally, at the end of the search, the globally best solution is optimized with the Quasi-Newton method to ensure maximum accuracy. Further enhancements to DMS have been proposed to improve its performance. In \cite{zhao2011dynamic}, the HS algorithm was incorporated into each sub-swarm (DMS-PSO-HS), allowing the algorithm to effectively utilize historical information from past solutions. Another significant improvement is the hybridization of DMS-PSO with the modified multi-trajectory search (MTS) and sub-regional HS (DMS-PSO-SHS) \cite{zhao2010dynamic}, which expands the search to a larger potential space across sub-populations.

Despite advancements, the cooperative mechanism among sub-swarms in DMS-PSO remains confined to periodic regrouping, which limits its ability to fully exploit global search information. When sub-swarms are trapped in local minima, regrouping primarily exchanges local optima, hindering global optimization and exploration. Xu et al. \cite{xu2015dynamic} introduced a cooperative learning strategy (CLS) in DMS-PSO to address this. Before regrouping, using a tournament selection strategy, the two worst particles in each sub-swarm learn from the better particles of two randomly selected sub-swarms. This approach facilitates better information exchange among sub-swarms, enabling particles to utilize the best knowledge across the population. DMS-PSO-CLS improves the balance between global exploration and local exploitation by enhancing communication and expanding the search space.

\subsection{Dynamic learning strategy}
The dynamic learning strategy (DyLS)~\cite{ye2017novel} introduces an innovative learning mechanism for PSO. Similar to the DMS, DyLS divides the population into multiple sub-populations. Consider a population with \( N_s \) particles, which is divided into \( k \) sub-populations denoted as \( \{h_1, h_2, \ldots, h_k\} \), where \( N_s = \sum_{p=1}^k |h_k| \). The first sub-population consists of particles \( 1 \) to \( h_1 \), the second includes particles \( h_1+1 \) to \( h_2 \), and so on. This partitioning effectively splits a large PSO into several smaller, independent PSOs.

\begin{figure}
\centering
\resizebox{0.65\linewidth}{!}{\begin{tikzpicture}[scale=1.5, every node/.style={font=\small}]

\draw[dashed, thick,fill=gray!20] (0,0) circle (1.5cm);
\draw[dashed, thick,fill=gray!20] (4,0) circle (1.5cm);
\draw[dashed, thick,fill=gray!20] (0,-3.5) circle (1.5cm);
\draw[dashed, thick,fill=gray!20] (4,-3.5) circle (1.5cm);

\node[star, star points=5, fill=blue!70!white, star point ratio=2, draw, scale=1] (lbest1) at (0,0) {};
\node[star, star points=5, fill=blue!70!white, star point ratio=2, draw, scale=1] (lbest2) at (4,0) {};
\node[star, star points=5, fill=blue!70!white, star point ratio=2, draw, scale=1] (lbest3) at (0,-4) {};
\node[star, star points=5, fill=blue!70!white, star point ratio=2, draw, scale=1] (lbest4) at (4,-4) {};

\node[star, star points=5, fill=red!70!white, star point ratio=2, draw, scale=1.5] (u lbest) at (2,-2) {};

\draw[->, line width=1.5pt, red] (lbest1) -- (u lbest);
\draw[->, line width=1.5pt, red] (lbest2) -- (u lbest);
\draw[->, line width=1.5pt, red] (lbest3) -- (u lbest);
\draw[->, line width=1.5pt, red] (lbest4) -- (u lbest);

\node[circle, fill=blue, scale=0.8] (cp1) at (1,0) {};
\node[circle, fill=blue, scale=0.8] (cp2) at (3,0) {};
\node[circle, fill=blue, scale=0.8] (cp3) at (1,-4) {};
\node[circle, fill=blue, scale=0.8] (cp4) at (3,-4) {};


\node[circle, fill=blue, scale=0.8] (op1) at (0.5,0.7) {};
\node[circle, fill=blue, scale=0.8] (op2) at (-0.5,0.6) {};
\node[circle, fill=blue, scale=0.8] (op3) at (-1,-0.8) {};
\node[circle, fill=blue, scale=0.8] (op4) at (1,-0.6) {};
\draw[->, blue] (0.5,0.7) -- (0.3,0.3);
\draw[->, blue] (-0.5,0.6) -- (-0.2,0.3);
\draw[->, blue] (-1,-0.8) -- (-0.6,-0.5);
\draw[->, blue] (1,-0.6) -- (0.5,-0.3);

\node[circle, fill=blue, scale=0.8] (op5) at (0.5,-0.7) {};
\node[circle, fill=blue, scale=0.8] (op6) at (0.1,-1) {};
\node[circle, fill=blue, scale=0.8] (op7) at (-0.9,0) {};
\node[circle, fill=red, scale=0.8] (op8) at (-0.3,-1.3) {};
\draw[->, dashed, line width=1.5pt, purple] (op8) to[out=10,in=200] (u lbest);

\node[circle, fill=blue, scale=0.8] (op21) at (4.5,0.7) {};
\node[circle, fill=blue, scale=0.8] (op22) at (3.5,0.6) {};
\node[circle, fill=blue, scale=0.8] (op23) at (3,-0.8) {};
\node[circle, fill=blue, scale=0.8] (op24) at (5,-0.6) {};
\draw[->, blue] (4.5,0.7) -- (4.3,0.3);
\draw[->, blue] (3.5,0.6) -- (3.8,0.3);
\draw[->, blue] (3,-0.8) -- (3.3,-0.5);
\draw[->, blue] (5,-0.6) -- (4.5,-0.3);

\node[circle, fill=blue, scale=0.8] (op25) at (4.5,-0.7) {};
\node[circle, fill=blue, scale=0.8] (op26) at (4.1,-1) {};
\node[circle, fill=blue, scale=0.8] (op27) at (4.9,0) {};
\node[circle, fill=red, scale=0.8] (op28) at (4.3,-1.3) {};
\draw[->, dashed, line width=1.5pt, purple] (op28) to[out=150,in=0] (u lbest);

\node[circle, fill=blue, scale=0.8] (op31) at (0.5,-4.7) {};
\node[circle, fill=blue, scale=0.8] (op32) at (-0.5,-4.6) {};
\node[circle, fill=blue, scale=0.8] (op33) at (-1,-3.8) {};
\node[circle, fill=blue, scale=0.8] (op34) at (1,-3.6) {};
\draw[->, blue] (0.5,-4.7) -- (0.3,-4.3);
\draw[->, blue] (-0.5,-4.6) -- (-0.2,-4.3);
\draw[->, blue] (-1,-3.8) -- (-0.6,-3.9);
\draw[->, blue] (1,-3.6) -- (0.5,-3.9);

\node[circle, fill=blue, scale=0.8] (op35) at (-1,-3) {};
\node[circle, fill=blue, scale=0.8] (op36) at (0.1,-3) {};
\node[circle, fill=blue, scale=0.8] (op37) at (1,-2.7) {};
\node[circle, fill=red, scale=0.8] (op38) at (-0.3,-2.5) {};
\draw[->, dashed, line width=1.5pt, purple] (op38) to[out=10,in=200] (u lbest);

\node[circle, fill=blue, scale=0.8] (op41) at (4.5,-4.7) {};
\node[circle, fill=blue, scale=0.8] (op42) at (3.5,-4.6) {};
\node[circle, fill=blue, scale=0.8] (op43) at (3,-3.8) {};
\node[circle, fill=blue, scale=0.8] (op44) at (5,-3.6) {};
\draw[->, blue] (4.5,-4.7) -- (4.3,-4.3);
\draw[->, blue] (3.5,-4.6) -- (3.7,-4.3);
\draw[->, blue] (3,-3.8) -- (3.6,-3.8);
\draw[->, blue] (5,-3.6) -- (4.5,-3.8);

\node[circle, fill=blue, scale=0.8] (op45) at (4.5,-2.5) {};
\node[circle, fill=blue, scale=0.8] (op46) at (3.1,-2.5) {};
\node[circle, fill=blue, scale=0.8] (op47) at (4.9,-4) {};
\node[circle, fill=red, scale=0.8] (op48) at (4.3,-3.3) {};
\draw[->, dashed, line width=1.5pt, purple] (op48) to[in=-100, out=200] (u lbest);

\fill[blue] (6,-1.5) circle (0.1cm);
\node at (6.7,-1.5) {Particles};

\fill[red] (6,-2) circle (0.1cm);
\node at (7.5,-2) {Communication particles};

\node[star, star points=5, fill=blue!70!white, star point ratio=2, draw, scale=0.7] at (6,-2.5) {};
\node at (6.5,-2.5) {$lbest$};

\node[star, star points=5, fill=red!70!white, star point ratio=2, draw, scale=1.2] at (6,-3) {};
\node at (7,-3) {United $lbest$};

\end{tikzpicture}}
 \caption{DLS strategy.} \label{fig: DyLS}
\end{figure}
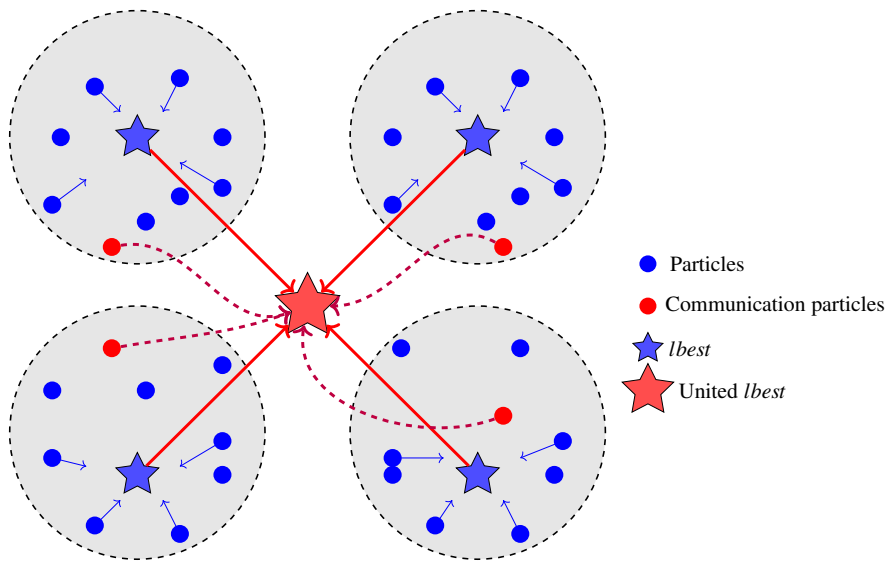

The best-performing particle is designated as \( lbest \) in each subpopulation. DyLS calculates the average of the \( lbest \) values from all sub-populations to update particle velocities using the following equation:
\begin{equation}\label{eq: DyLS velocity}
\mathbf{vel}_i^d = w_g \cdot \mathbf{vel}_i^d + c_{e_1} \cdot rand_i^d \cdot \left( \mathbf{p}_{Best_i}^d - \mathbf{x}_i^d \right)+ c_{e_2} \cdot rand_i^d \cdot \left( \overline{\mathbf{l}_{Best}^d} - \mathbf{x}_i^d \right),
\end{equation}

where \( \overline{\mathbf{l}_{Best}^d} = \frac{1}{k} \sum_{p=1}^k \overline{\mathbf{l}_{Best_p}^d} \), and \( \mathbf{l}_{Best_p}^d \) represents the best position achieved so far in the \( p^{th} \) sub-population. The final term in Eq.~\eqref{eq: DyLS velocity} is the united \( \mathbf{l}_{Best} \). Fig.~\ref{fig: DyLS} illustrates the DyLS framework in the context of a multi-population PSO. The strategy classifies particles in each sub-swarm into ordinary and communication particles. Ordinary particles search for better local bests $\mathbf{l}_{Best}$ within their sub-swarm, following the traditional framework. Communication particles, however, actively interact between sub-swarms, breaking free from $\mathbf{l}_{Best}$'s control and exploring additional useful information (Eq. \eqref{eq: DyLS velocity}). 

\subsection{Multi-exemplar AL}
Wei et al. \cite{wei2020multiple} proposed a multi-exemplar AL for PSO, where particles are divided into multiple subgroups, each utilizing multi-exemplar adaptive learning (MAL) based on particle performance. Based on MALs, each subgroup particle selects distinct exemplars within a generation and across generations. After choosing candidate exemplars, a particle \( i \) updates its velocity using its learning model. In each generation, the \( i \)th particle has two exemplars: \( \mathbf{Exemplar1}_{i}=\{Exemplar1_{i}^1, Exemplar1_{i}^2,\ldots, Exemplar1_{i}^d \}\), the best-so-far position of its swarm, and \( \mathbf{Exemplar2}_{i}=\{Exemplar2_{i}^1, Exemplar2_{i}^2,\ldots, Exemplar2_{i}^d \} \), a randomly selected neighbor from its candidate learning exemplars. The velocity update rule is given by:  
\begin{equation}
\mathbf{vel}_i^d = w_g \cdot \mathbf{vel}_i^d + rand_1 \cdot (\mathbf{Exemplar1}_i^d - \mathbf{x}_i^d) + rand_2 \cdot (\mathbf{Exemplar2}_i^d - \mathbf{x}_i^d).
\end{equation}
MAL removes the acceleration coefficients used in canonical PSO to simplify the velocity update. In this framework, elite particles focus on exploitation due to high-quality exemplars, while inferior particles benefit from increased exploration by following randomly selected exemplars. Mediocre particles balance exploration and exploitation through the combined influence of their exemplars. This approach allows particles within the same subgroup to exhibit diverse learning behaviors. The other AL strategies are presented in \cite{wang2024hybrid,tian2024particle,wu2024novel,liu2020modified}

\subsection{Hierarchical learning}
Wang et al. \cite{wang2023hierarchical} proposed a fuzzy logic-based hierarchical learning (HL) for PSO, where particles are stratified into four layers (10\%, 20\%, 30\%, and 40\%) based on fitness. Each layer classifies particles as winners (high-energy) or losers (low-energy). Winners learn from upper-layer particles, while losers learn from winners within the same layer, promoting diversity and mitigating premature convergence.

\textit{Hierarchical energy gathering point (EM)}: Aggregates all particles' positions within a layer. \textit{High-energy gathering point (EMH)}: Aggregates only high-energy particles' positions. These gathering points enable one-to-many cooperation, enhancing information exchange and search efficiency. The velocity updates for high-energy and low-energy particles are:
\begin{align}
\mathbf{vel}_{MH_{i}}^d =& \omega_g\cdot \mathbf{vel}_{MH_{i}}^d + c_{e_1}\cdot (\mathbf{p}_{Best_{i}}^d - \mathbf{x}_{MH_{i}}^d) + c_{e_2}\cdot \epsilon\cdot (\mathbf{g}_{Best}^d - \mathbf{x}_{MH_{i}}^d) + a \cdot(\mathbf{EM}_{M-1} - \mathbf{x}_{MH_{i}}^d)\\
\mathbf{vel}_{ML_{i}}^d = &\omega_g\cdot \mathbf{vel}_{ML_{i}}^d + c_{e_1}\cdot (\mathbf{p}_{Best_{i}}^d - \mathbf{x}_{ML_{i}}^d) + r\cdot (\mathbf{EMH} - \mathbf{x}_{ML_{i}}^d),
\end{align}
where \(\mathbf{x}_{MH_i}^{d}\) and \(\mathbf{x}_{ML_i}^{d}\) denote the positions of high-energy and low-energy particles, respectively. The energy gathering points are defined as: 
$\mathbf{EM}_{M}=\frac{1}{N_s} \sum_{i=1}^{N_s}\mathbf{x}_{M_i}, \quad
\mathbf{EMH}=\frac{1}{N_s} \sum_{i=1}^{N_s} \mathbf{x}_{MH_i},~
\{\mathbf{x}_{MH_i} \mid f_{\mathbf{x}_{MH_i}} < \text{Average}(f_{\mathbf{x}_{M}})\}$. High-energy particles exploit global and hierarchical knowledge, while low-energy particles rely on local structures, ensuring a balanced exploration-exploitation tradeoff. This learning is also used by Li et al. \cite{li2022pyramid}, who proposed an HL-based pyramid PSO. In pyramid PSO, the population is divided into layers based on fitness values, and each layer is divided into winners and losers, where winners learn from the above-layer individuals or top-most layer individuals, and losers learn from the winners of the same layer.

\subsection{Other learning strategies}
 Tang et al. \cite{tang2021multi} introduced a level-based inter-task learning strategy with a dynamic local topology. Particles are grouped into levels using distinct learning methods for cross-task information sharing. Diverse search preferences enhance exploration and refinement, while systematic updates to the local topology optimize inter-task neighbor selection. A reinforcement learning (RL) level approach is proposed by Wang et al. \cite{wang2022reinforcement} for PSO, where particles are divided into $l$ levels based on their fitness values. RL and competition strategies are used to control the number of levels, enhancing the search efficiency of PSO. 

\begin{table}[]
    \centering
    \renewcommand{\arraystretch}{1.2}
      \caption{multiple-population-based learning strategies for PSO.}\label{tab: multiple population-based}
\resizebox{1\linewidth}{!}{
\begin{tabular}{llccp{9cm}}\hline
Long name	&	Sort name	&	Year	&	Ref.	&	Comments	\\\hline
Dynamic multi-swarm PSO	&	DMS-PSO	&	2005	&	\cite{liang2005dynamic}	&	This method builds on the local version of PSO and introduces a novel neighborhood topology characterized by two key features: small neighborhoods and a randomized regrouping mechanism. DMS-PSO employs small neighborhoods, in contrast to PSO approaches that often favor larger populations. By dividing the population into smaller swarms, DMS-PSO slows down the convergence velocity and enhances the diversity of solutions. Each swarm searches independently within the solution space using its members' historical information as $\mathbf{vel}_i^d = w_g \cdot \mathbf{vel}_i^d + c_{e_1} \cdot rand_i^d \cdot \left( \mathbf{p}_{Best_i}^d - \mathbf{x}_i^d \right)+ c_{e_2} \cdot rand_i^d \cdot \left( \mathbf{l}_{Best_i}^d - \mathbf{x}_i^d \right).$ The parameters $c_{e_1}=c_{e_2}=1.49445$, $\omega_g=0.729$, and $R_p=5$.	\\
Dynamic learning strategy	&	DyLSPSO	&	2017	&	\cite{ye2017novel}	&	Similar to the DMS, DLS divides the population into multiple sub-populations. Consider a population with \( N_s \) particles, which is divided into \( k \) sub-populations denoted as \( \{h_1, h_2, \ldots, h_k\} \), where \( N_s = \sum_{p=1}^k |h_k| \), where \( \overline{\mathbf{l}_{Best}^d} = \frac{1}{k} \sum_{p=1}^k \overline{\mathbf{l}_{Best_p}^d} \).	\\
Hierarchical learning PSO	&	HLPSO	&	2023	&	\cite{wang2023hierarchical}	&	In HL, particles are stratified into four layers (10\%, 20\%, 30\%, and 40\%) based on fitness. Each layer classifies particles as winners (high-energy) or losers (low-energy). Winners learn from upper-layer particles, while losers learn from winners within the same layer, promoting diversity and mitigating premature convergence. High-energy particles exploit global and hierarchical knowledge, while low-energy particles rely on local structures, ensuring a balanced exploration-exploitation tradeoff.	\\
Reinforcement learning PSO	&	RLPSO	&	2022	&	\cite{wang2022reinforcement}	&	In RLPSO, particles are ranked into $l$ levels based on fitness, with reinforcement learning (RL) and competition strategies optimizing level control for enhanced search efficiency. RL-driven velocity updates, cosine similarity for velocity control, and local updates to prevent premature convergence \cite{li2023reinforcement}. 	\\
Inter-task learning PSO	&	ITLPSO	&	2021	&	\cite{tang2021multi}	&	A level-based inter-task learning strategy with dynamic topology groups particles into levels for cross-task learning. Diverse search preferences balance exploration and refinement, while adaptive topology updates optimize neighbor selection.	\\\hline
\end{tabular}}
\end{table}

\section{Theoretical development}\label{sec: theoretical}
This paper reviews learning strategies for PSO and explores theoretical developments from original works. As with most PSO convergence analyses \cite{clerc2002particle,fernandez2010stochastic}, a deterministic implementation is used for theoretical assessment, primarily establishing stability rather than guaranteeing convergence to the global optimum. These analyses focus on whether the swarm reaches a stable point without confirming if it is the best solution.

A well-balanced optimization algorithm must navigate the trade-off between exploration and convergence. Exploration enables broad search space coverage to escape local optima, while convergence ensures particles stabilize around an optimal solution. An imbalance can degrade performance, excessive exploration leads to inefficiency, while premature convergence risks stagnation in suboptimal regions. Most mathematical analyses of PSO address stability but often overlook the distinction between mature and premature convergence. Poli \cite{poli2008dynamics} conducted a moment-based stability analysis of PSO, assuming a stagnation phase where particles stop moving—an unrealistic assumption. This study confirmed order-1 and order-2 stability but did not determine whether the final point is globally optimal.

Building on Poli's work, Nasir et al. \cite{nasir2012dynamic} extended the theoretical analysis to DNLPSO, an improved CLPSO variant. They showed that setting acceleration coefficients \( c_{e_1} = c_{e_2} > 1.2 \) while maximizing \( P_{r_c} \) preserves exploration by adjusting the topological distance between particles, influencing their attraction toward solutions. DNLPSO further enhances search efficiency by dynamically restructuring neighborhoods, ensuring diverse exemplar selection, and preventing stagnation, unlike CLPSO, where exemplars remain static. Cheng and Jin \cite{cheng2015social} provided a convergence proof for SLPSO, modeling its dynamics as:  
\begin{align*}
&\mathbf{x}_i^d(t+1) = \mathbf{x}_i^d(t) + \mathbf{vel}_i^d(t+1),\\&
\mathbf{vel}_i^d(t+1) = \frac{1}{2} \cdot \mathbf{vel}_i^d(t) + \frac{1}{2} \cdot \left( \mathbf{x}_k^d(t) - \mathbf{x}_i^d(t) \right) + \frac{1}{2} \cdot \epsilon \cdot \left( \overline{\mathbf{x}}^d(t) - \mathbf{x}_i^d(t) \right).
\end{align*}
converges to equilibrium if the social influence parameter \( \epsilon > -1 \). Huang et al. \cite{huang2012example} established ELPSO convergence, linking it to PSO and CLPSO. Using Markov process theory, they demonstrated that if PSO/CLPSO converges, ELPSO does as well. Conversely, if ELPSO fails to converge, PSO/CLPSO will fail for the same optimization problem.

Inspired by the above and theoretical development mentioned in \cite{van2006study}, we design the convergence proof for CLPSO \cite{liang2006comprehensive}. For a one-dimensional search space, Eq. \eqref{eq: CL velocity} of CLPSO is written as: 
\begin{align*}
&\mathbf{x}(t+1)=\mathbf{x}(t)+\mathbf{vel}(t+1)
\\&\mathbf{vel}(t+1) = w_g \cdot \mathbf{vel}(t) + \phi \cdot \left( \mathbf{p}(t) - \mathbf{x}(t) \right),
\end{align*}where $\mathbf{p}$ is nothing but $\mathbf{p}_{Best_{f_i}}$ and $\phi=c_e\cdot rand_1$. Substituting velocity into the position equation, the following equation is obtained:
\begin{align}
&\mathbf{x}(t+1)=(1+\omega_g-\phi)\mathbf{x}(t)-w_g \cdot \mathbf{x}(t-1)-\phi \cdot \mathbf{p}(t)
\end{align}
which can be rewritten in the form of the following matrix:
\begin{equation}\label{eq: stochastic form}
\begin{bmatrix}
\mathbf{x}(t+1) \\ \mathbf{x}(t) \\ 1
\end{bmatrix}
=
A
\begin{bmatrix}
\mathbf{x}(t) \\ \mathbf{x}(t-1) \\ 1
\end{bmatrix}, \quad A =
\begin{bmatrix}
1 + \omega_g - \phi & -\omega_g & \phi \mathbf{p} \\
1 & 0 & 0 \\
0 & 0 & 1
\end{bmatrix}. 
\end{equation}
The matrix \( A \) satisfies the eigenvalue equation \( |A - \Lambda I| = 0 \), yielding eigenvalues:
$\Lambda_0 = 1,~~
\Lambda_1 = \frac{1 + \omega_g - \phi + \Gamma}{2},$ $
\Lambda_2 = \frac{1 + \omega_g - \phi - \Gamma}{2},$ where $\Gamma = \sqrt{(1 + \omega_g - \phi)^2 - 4\omega_g}.$
Thus, Eq. \eqref{eq: stochastic form} can be rewritten as:
\begin{equation}\label{eq: stochastic form1}
\mathbf{x}(t) = k_1 + k_2 \Lambda_1^t + k_3 \Lambda_2^t.
\end{equation}
where
\begin{equation*}
\begin{bmatrix}
k_1 \\ k_2 \\ k_3
\end{bmatrix}
=
\begin{bmatrix}
\mathbf{p} \\ 
\frac{\Lambda_2 -(1+\Lambda_2) \mathbf{x}(t) + \mathbf{x}(t+1)}{\Gamma (\Lambda_1 - 1)} \\ 
\frac{ -\Lambda_1+(1+\Lambda_1)\mathbf{x}(t) - \mathbf{x}(t+1)}{\Gamma (\Lambda_2 - 1)}
\end{bmatrix}.
\end{equation*}
In Eq. (\ref{eq: stochastic form1}), when the condition $\max (|\Lambda_1|, |\Lambda_2|) < 1$
is satisfied \cite{van2006study}, it follows that: $\lim_{t \to \infty} k_1 \Lambda_1^t = 0,$ $\lim_{t \to \infty} k_2 \Lambda_2^t = 0.$ Therefore, the particle positions eventually converge to a stable point \( \mathbf{p} \), i.e.,
\begin{equation}
\lim_{t \to \infty} \mathbf{x}(t) = \lim_{t \to \infty} (k_1 + k_2 \Lambda_1^t + k_3 \Lambda_2^t) = \lim_{t \to \infty} (k_1 + 0 + 0) = k_1 = \mathbf{p}.
\end{equation}

\section{Experimental analysis}\label{sec: experimental analysis}
In this section, we compare the optimization performance of selected learning strategies, including multiple adaptive learning (MAL) \cite{wei2020multiple}, strategy dynamics learning (SDL) \cite{liu2022strategy}, dynamic neighborhood learning (DNL) \cite{nasir2012dynamic}, comprehensive learning (CL) \cite{liang2006comprehensive}, dynamic multi-swarm (DMS) \cite{liang2005local}, dimensional learning strategy (DLS) \cite{xu2019particle}, social learning (SL) \cite{cheng2015social}, multi-exemplar forgetting learning (MFL) \cite{xia2020expanded}, and adaptive strategy (AL) \cite{liu2020modified}, fully-informed strategy (FIS) \cite{mendes2004fully}, locally-informed strategy (LIS) \cite{qu2012distance}, ensemble learning (EL) \cite{lynn2017ensemble}, and unified learning (UL) \cite{parsopoulos2019upso}.  

These strategies were selected based on the availability of publicly accessible source codes, ensuring reproducibility. We exclude the original PSO, as their respective authors have already validated these strategies' effectiveness. All parameter settings remain unchanged except for population size and function evaluations. Details are presented in Table~\ref{tab: Learning strategies for experimental}. It is noted that SDL and EL used various learning strategies to improve the impact of PSO. 
\begin{table}[]
    \centering
    \caption{Selected learning strategies for experimental analysis.}\label{tab: Learning strategies for experimental}
\resizebox{0.8\linewidth}{!}{\begin{tabular}{cclclc}\hline
\multicolumn{3}{l}{$N_s=75$, $d=5,~10,~15,~20,~30,~50,~100$, $MaxFE=75,000$}\\\hline
      SR   &   Strategy Name &PSO Name& Year&Author& Ref. \\\hline
        1. &FIS& Fully-informed strategy-based PSO & 2004& Mendes et al.&\cite{mendes2004fully}\\
        2. & DMS& Dynamic multi-swarm-based PSO & 2005 & Liang et al. & \cite{liang2005local}\\
        3.  & CL & Comprehensive learning-based PSO & 2006& Liang et al.&\cite{liang2006comprehensive}\\
        4. & DNL &Dynamic neighborhood learning-based PSO &2012& Nasir et al.& \cite{nasir2012dynamic}\\
        5. & LIS & Locally-informed PSO & 2012 & Qu et al. &\cite{qu2012distance}\\
        6. & SL & Social learning-based PSO & 2015 & Cheng and Jin &\cite{cheng2015social}\\
        7. & EL & Ensemble learning PSO & 2017 & Lynn et al. & \cite{lynn2017ensemble}\\
        8. & DLS&Dimensional learning strategy-based PSO & 2019& Xu et al.& \cite{xu2019particle}\\
        9. & UL & Unified learning PSO & 2019 & Parsopoulos et al. & \cite{parsopoulos2019upso}\\
        10. & MFL & Multi-exemplar forgetting learning-based PSO & 2020 & Xia et al. & \cite{xia2020expanded}\\
        11. & AL & Adaptive learning-based PSO & 2020 & Liu et al. & \cite{liu2020modified}\\
        12. & MAL & Multiple adaptive learning-based PSO & 2020 & Wei et al. & \cite{wei2020multiple}\\
        13. & SDL & Strategy dynamics learning-based PSO & 2022 & Liu et al. & \cite{liu2022strategy}\\\hline
    \end{tabular}}
\end{table}
 We evaluate the optimization performance of the selected learning strategies on 10 unimodal, 16 multimodal, 13 hybrid, and 13 composite benchmark problems. These problems are sourced from basic test functions \cite{cheng2015social}, CEC 2017 \cite{awad2016problem}, and CEC 2020 \cite{yue2019problem}, along with classical benchmark functions from \cite{brest2006self,zhang2009jade}.  

The population size is set to \(N_s = 75\), and the maximum number of function evaluations is 75,000, following the recommendation in \cite{piotrowski2020population}. Each algorithm is executed for 25 independent runs, and the mean objective function values are recorded. The optimization performance is ranked for each strategy, where rank 1 is assigned to the strategy with the lowest mean objective function value and rank 13 to the poorest-performing strategy.  

For each benchmark set (basic test functions, CEC 2017, CEC 2022), the average rank of each strategy is computed by aggregating results across all problems within that set and across different dimensionalities where applicable (e.g., CEC 2013 and CEC 2017). The basic results of all strategies are presented in the Supplementary file. All experiments were conducted on a Windows 11 12th Gen Intel(R) Core(TM) i5 16GB using MATLAB 2024b.

We assessed the statistical significance of pairwise comparisons among the 13 learning strategies across different problem sets and dimensionalities using the Wilcoxon signed-rank test \cite{derrac2011practical} 
at a 5\% significance level, as recommended in \cite{garcia2008extension,ulacs2012cost}. Additionally, ranking evaluations were conducted using the Friedman test \cite{friedman2001elements}.  

We identify learning strategies for each algorithm and problem sets that are statistically significantly better, worse, or similar to the learning mechanism that performs well in the Friedman rank test. 
The statistical test only determines whether differences exist at \(\alpha = 0.05\). However, we interpret a learning strategy as superior (inferior) if its mean rank (evaluated by the Friedman test \cite{friedman2001elements}) is lower (higher) than its competitors and the test confirms statistical significance. Additionally, we evaluated the final \texttt{AvgRanks} values to analyze the evolutionary behavior of each learning strategy. To compute \texttt{AvgRanks}, we first recorded the best fitness values at each function evaluation across all runs. Then, we ranked the learning strategies for each function in every run and averaged these ranks across all functions. Additionally, the analysis of search behavior and diversity is also discussed in this article.  

\begin{figure}
	\centering
	 \begin{subfigure}[b]{0.49\linewidth}
	\includegraphics[width=1\linewidth]{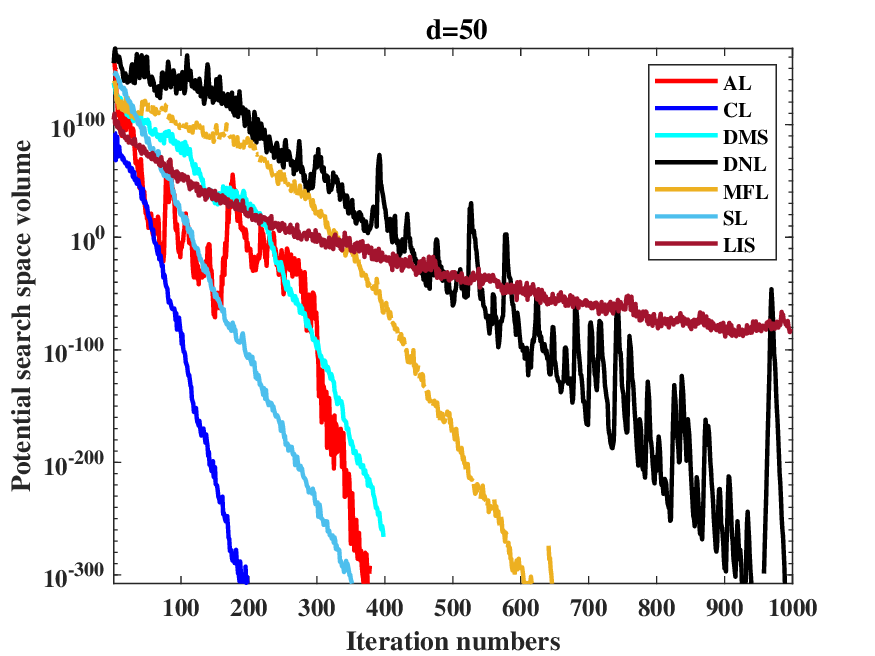}
        \caption{}\label{subfig: potential sphere}
        \end{subfigure}
        \begin{subfigure}[b]{0.49\linewidth}	
        \includegraphics[width=1\linewidth]{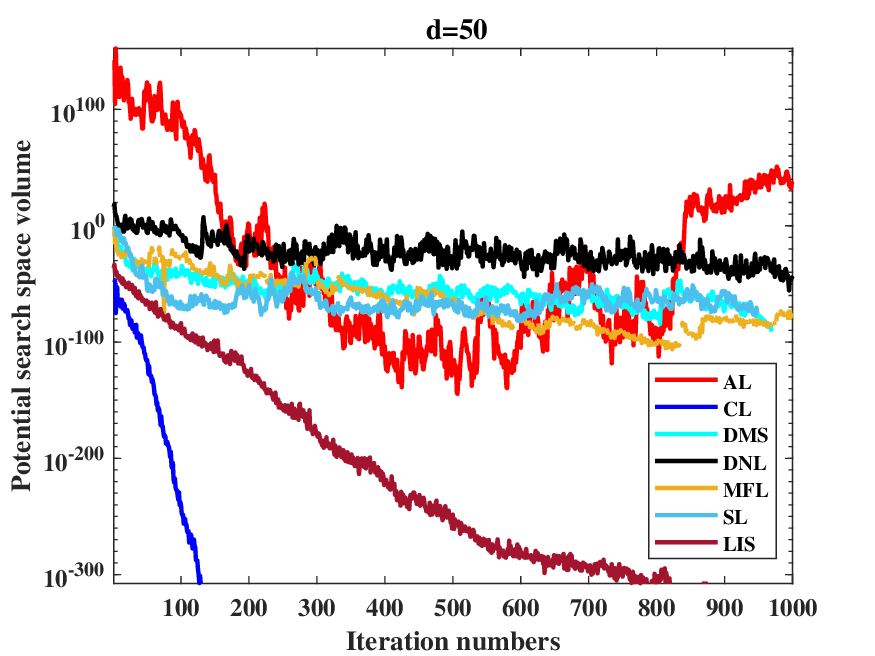}
        \caption{}\label{subfig: potential rosenbrock}
        \end{subfigure}
        \begin{subfigure}[b]{0.49\linewidth}\centering
	\includegraphics[width=1\linewidth]{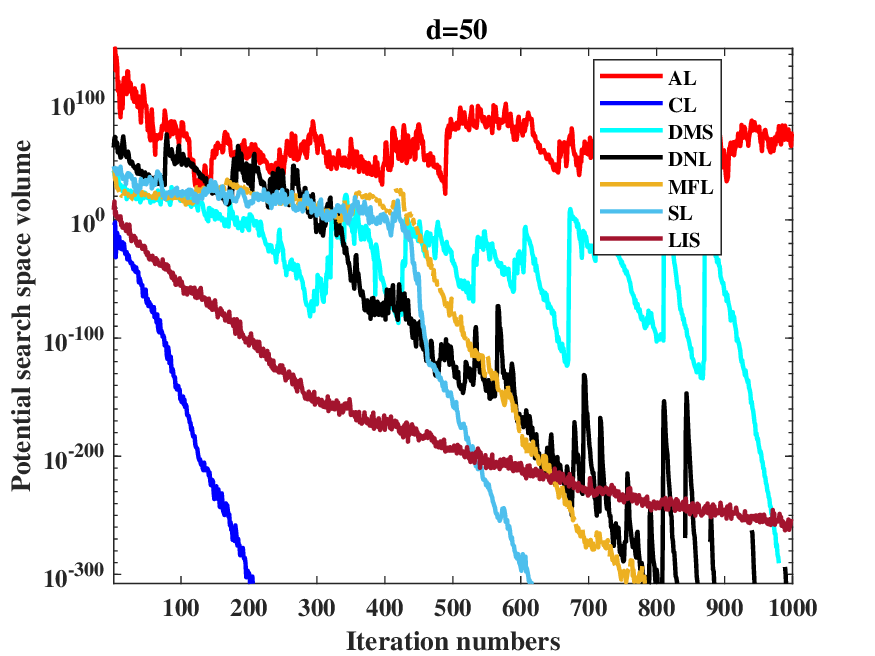}
        \caption{}\label{subfig: potential rastrigin}
	 \end{subfigure}
     \begin{subfigure}[b]{0.49\linewidth}\centering
	\includegraphics[width=1\linewidth]{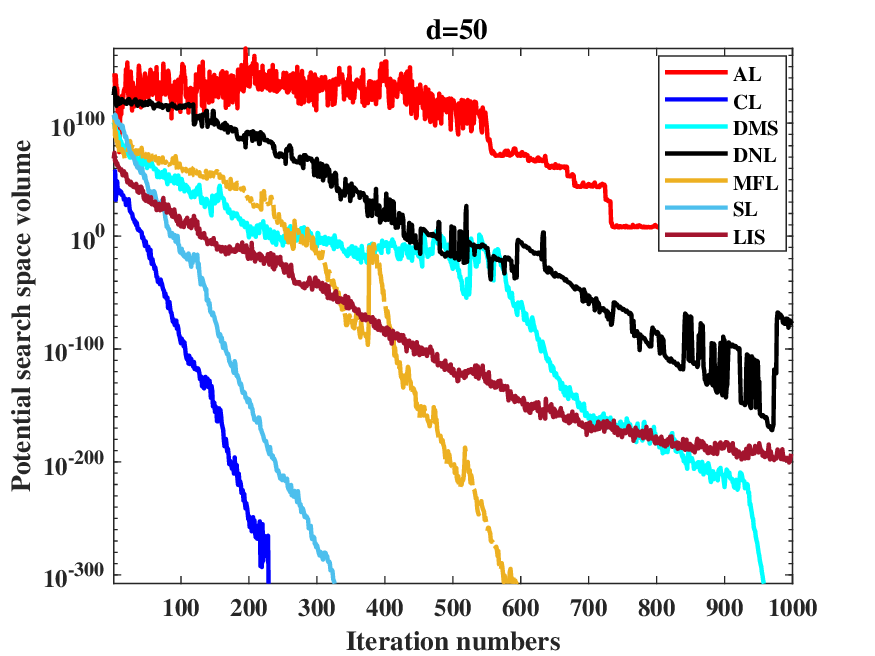}
        \caption{}\label{subfig: potential ackley}
	 \end{subfigure}
	\caption{Comparison of potential search space's volume for two unimodal problems ((a) Sphere and (b) Rosenbrock) and two multimodal problems ((c) Rastrigin and (d) Ackley) at 50 dimensions.}\label{fig: potential search}
\end{figure}

\subsection{Search space behavior analysis}
In this section, we analyze the search space volume of selected learning strategies for two unimodal and two multimodal problems in a 50-dimensional search space. We consider seven learning strategies: AL, DNL, DMS, CL, SL, MFL, and LIS. The length of the potential search space for these strategies is defined in Table~\ref{tab: search space length}. The potential search space volume for the $i$th particle under each learning strategy is given by \cite{liang2006comprehensive}:  

\begin{equation}
    R_{i,j} = \prod_{d=1}^{D} r_{ji}^{d}, \quad j = 1,2,3,4,5.
\end{equation}

Where $\bar{R}_{j}=mean(R_{i,j})$ represents the mean volume of the potential search spaces across the entire swarm for a given learning strategy. To ensure a fair comparison, we execute each strategy 25 times on both unimodal and multimodal functions. This evaluation allows us to assess how different learning strategies influence the search space exploration and convergence behavior across different problem landscapes.

\begin{table}[]
    \centering
     \caption{Mathematical formulations of search space lengths.}
    \label{tab: search space length}
  \resizebox{0.8\linewidth}{!}{\begin{tabular}{ll|ll}\hline
 Name & Search space length & Name & Search space length\\\cline{1-4}
    AL:   & $\mathbf{r}_{1i}^d=|\mathbf{S}_{Best_i}^d-\mathbf{x}_i^d|$ & CL: &$\mathbf{r}_{2i}^d=max(\mathbf{p}_{Best_{f_i^d}^d},\mathbf{x}_i^d)-min(\mathbf{p}_{Best_{f_i^d}^d},\mathbf{x}_i^d)$\\\hline LIS: &$\mathbf{r}_{4i}^d=|\overline{\mathbf{p}_{Best_i}^d} - \mathbf{x}_i^d|$&      
      SL:&$\mathbf{r}_{3i}^d=|\mathbf{x}_{k}^d-\mathbf{x}_i^d|+|\mathbf{\bar{x}}^d-\mathbf{x}_i^d|$ \\\hline
       DNL:  & $\mathbf{r}_{5i}^d=|\mathbf{p}_{Best_{f_i^d}}^d - \mathbf{x}_i^d|+|\mathbf{g}_{Best}^d - \mathbf{x}_i^d|$&DMS:&$\mathbf{r}_{6i}^d=|\mathbf{p}_{Best_i}^d - \mathbf{x}_i^d|+|\mathbf{l}_{Best_i}^d - \mathbf{x}_i^d|$\\\hline
       MFL:& \multicolumn{3}{l}{$\mathbf{r}_{7i}^d=|\mathbf{p}_{Best_i}^d - \mathbf{x}_i^d|+|(1 - f_{g_i})\cdot\mathbf{l}_{Best_i}^d - \mathbf{x}_i^d|+|(1 - f_{g_i})\cdot\mathbf{g}_{Best}^d - \mathbf{x}_i^d|$}\\\hline
    \end{tabular}}
   
\end{table}


Fig. \ref{fig: potential search} presents a comparative analysis of the potential search space's volume for four benchmark optimization problems in a 50-dimensional space. The problems include two unimodal functions: Fig. \ref{subfig: potential sphere} Sphere and Fig. \ref{subfig: potential rosenbrock} Rosenbrock, and two multimodal functions: Fig. \ref{subfig: potential rastrigin} Rastrigin and Fig. \ref{subfig: potential ackley} Ackley. The x-axis represents the number of iterations, while the y-axis, shown in a logarithmic scale, represents the potential search space volume. This metric helps assess how different optimization strategies navigate the search space over time.

For unimodal functions like the Sphere and Rosenbrock functions, the search space volume tends to decrease more smoothly and consistently. In Fig. \ref{subfig: potential sphere}, which corresponds to the Sphere function, most strategies exhibit rapid convergence, indicating that they quickly shrink the search space and focus on exploitation. This is expected, as the Sphere function has a simple structure with a single global minimum, making it easier for optimization methods to converge. In contrast, Fig. \ref{subfig: potential rosenbrock} (Rosenbrock function) shows a more gradual reduction in search space volume. This slower convergence is due to the narrow valley structure of the Rosenbrock function, which makes optimization more challenging, requiring strategies to balance exploration and exploitation carefully.

For multimodal functions, Figs. \ref{subfig: potential rastrigin} and Fig. \ref{subfig: potential ackley} show a significantly different pattern. The search space volume decreases more slowly and exhibits greater fluctuations, reflecting the presence of multiple local minima. In the Rastrigin function Fig. \ref{subfig: potential rastrigin}, the search space volume remains high for longer periods, especially for strategies that focus on exploration, such as AL (red) and DNL (black). These methods maintain diversity in the search space, preventing premature convergence to local optima. Similarly, in Fig. \ref{subfig: potential ackley} (Ackley function), some strategies exhibit high variance in search space volume, suggesting that they explore a wide range of possible solutions before settling into an optimum. The fluctuating behavior observed in certain methods indicates their ability to adapt to the rugged landscape of multimodal functions.

The comparison of strategies reveals distinct behavioral patterns. Methods like CL (blue) and DMS (cyan) tend to reduce search space volume quickly, demonstrating fast convergence. While this is beneficial for unimodal problems, it can be risky for multimodal problems, where premature convergence may lead to suboptimal solutions. On the other hand, AL (red) and DNL (black) maintain higher search space volume throughout the iterations, indicating a strong exploratory capability, which is essential for solving complex multimodal functions. Additionally, some strategies, such as MFL and SL, show intermediate behavior, balancing exploration and exploitation depending on the function's complexity.


\begin{figure}
	\centering
	\begin{subfigure}[b]{1\linewidth}
		\includegraphics[width=0.49\linewidth]{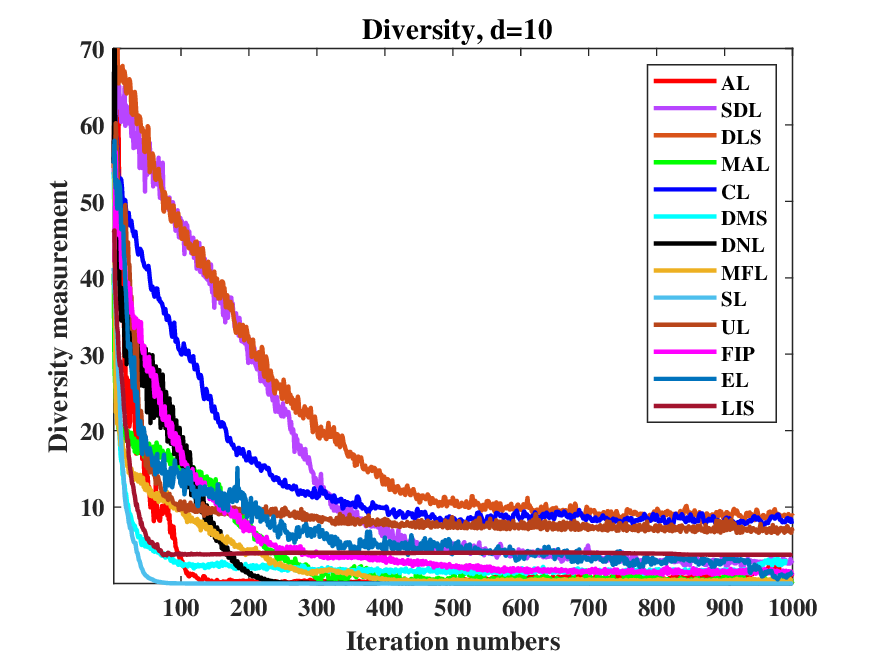}	
        \includegraphics[width=0.49\linewidth]{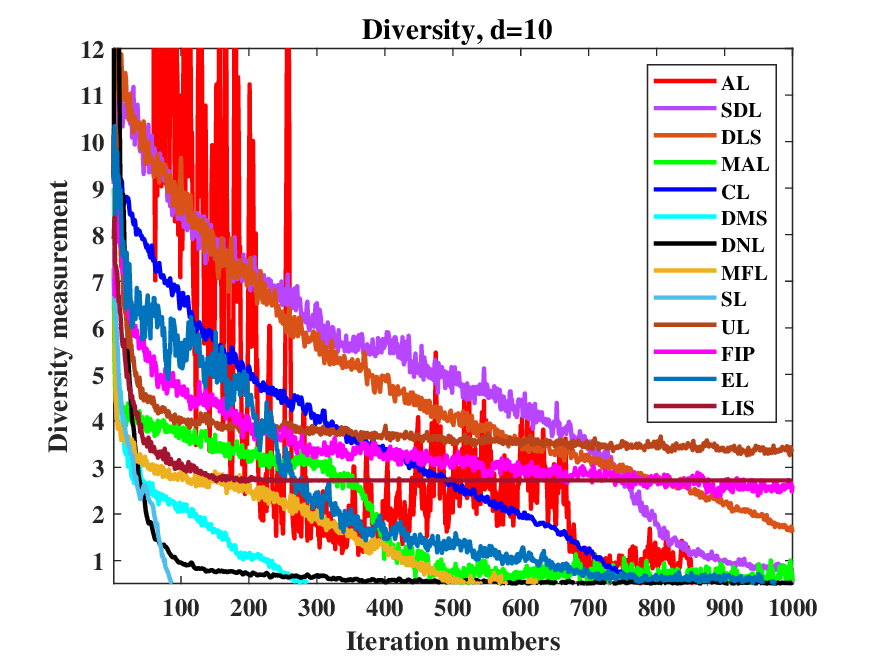}	
		\includegraphics[width=0.49\linewidth]{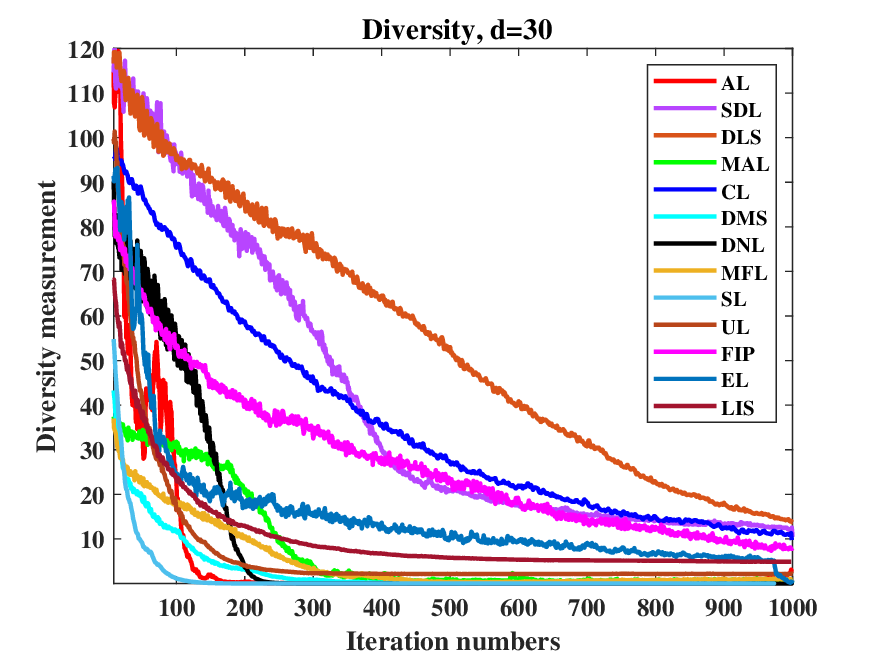}
        \includegraphics[width=0.49\linewidth]{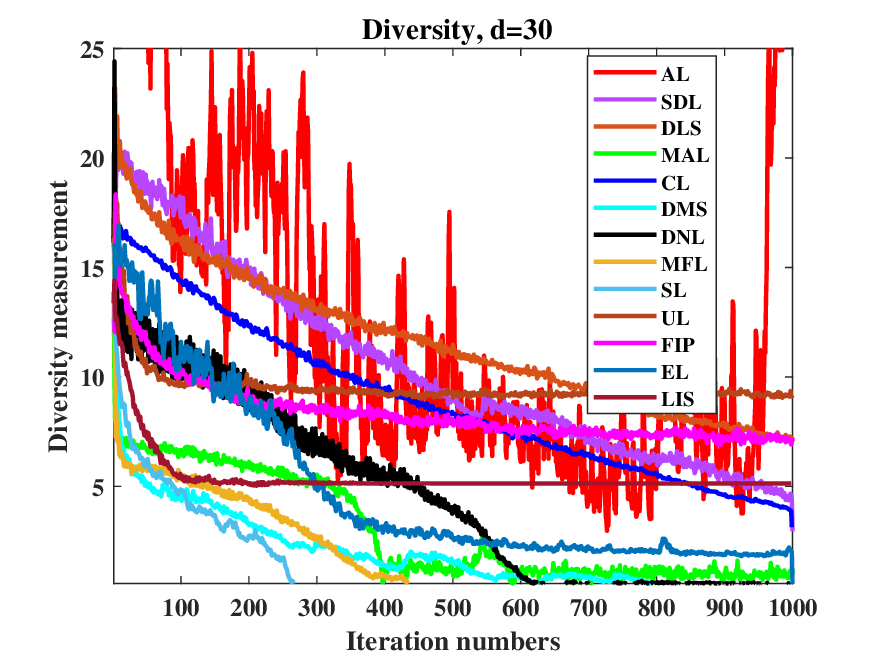}
		\includegraphics[width=0.49\linewidth]{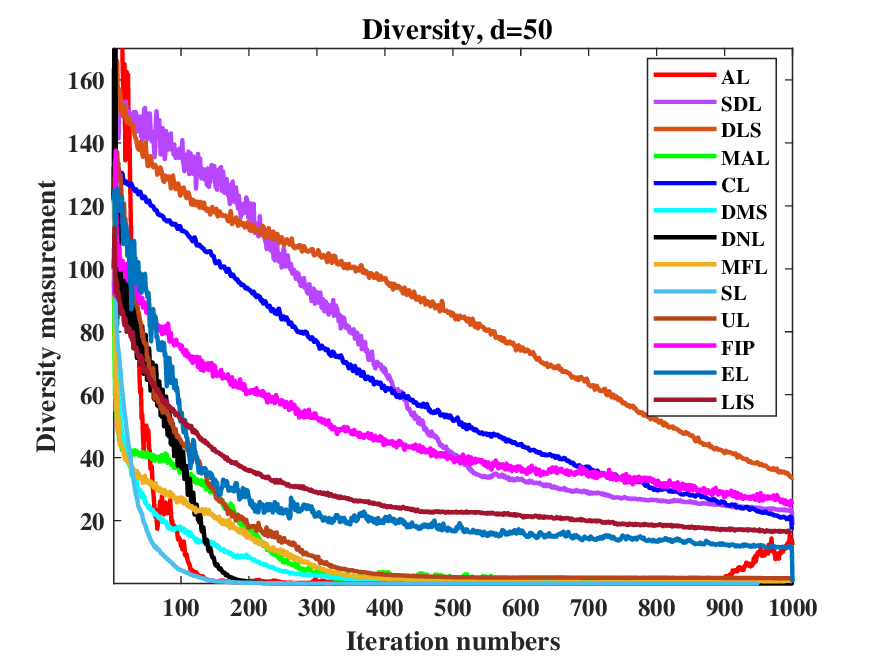}
        \includegraphics[width=0.49\linewidth]{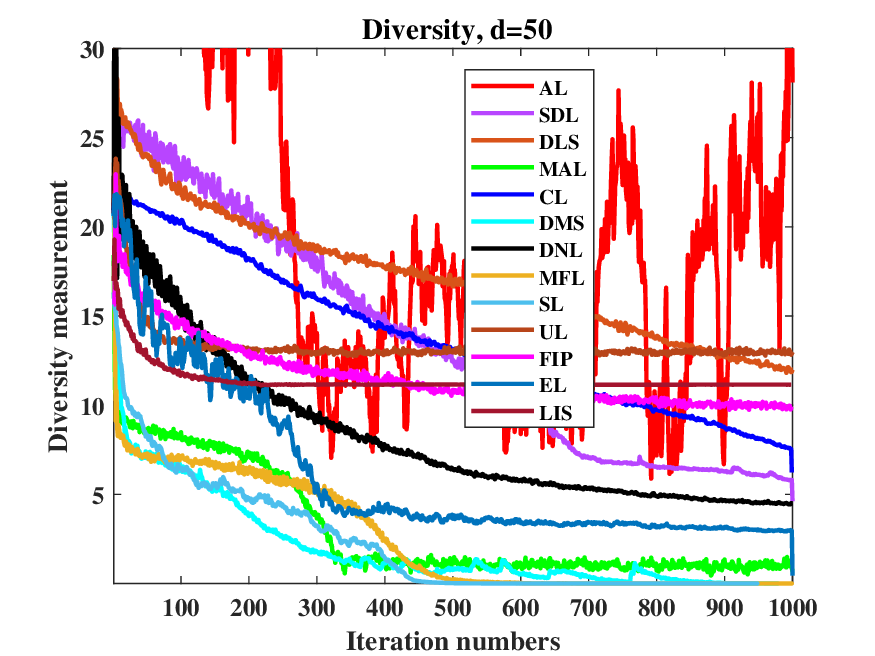}
		\includegraphics[width=0.49\linewidth]{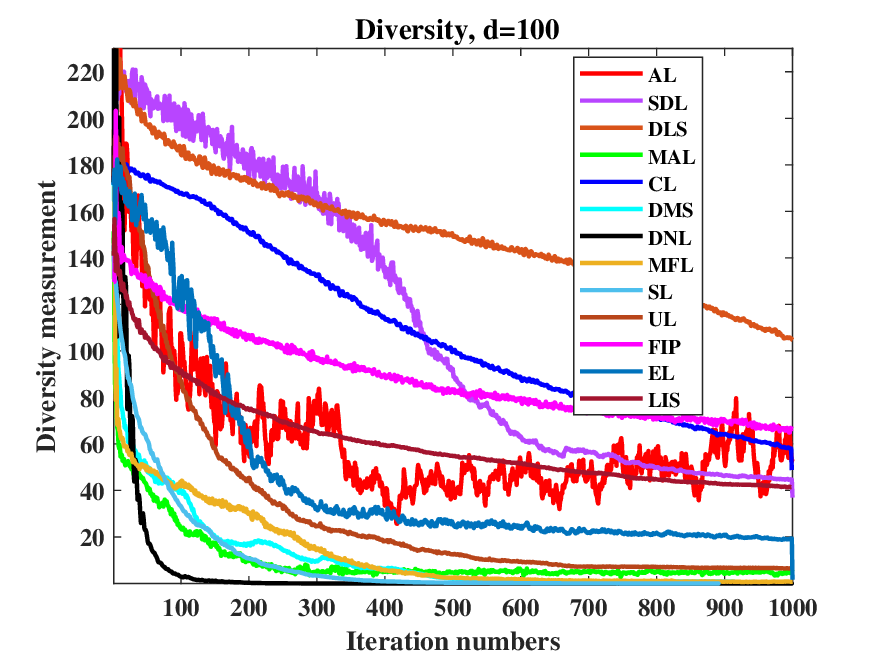}
        \includegraphics[width=0.49\linewidth]{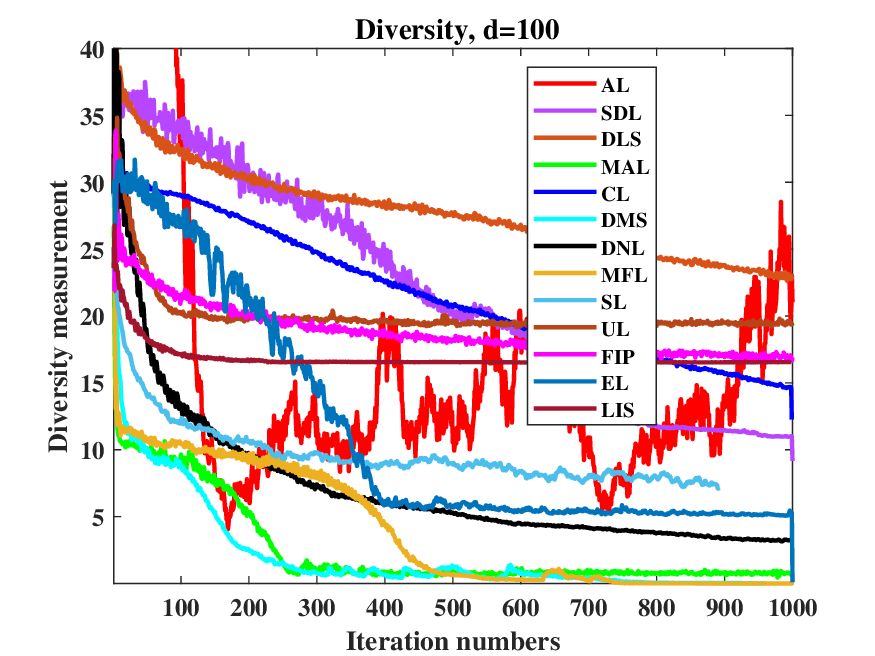}
	\end{subfigure}
	\caption{Analysis of diversity in the Rosenbrock (unimodal, left) and Rastrigin (multimodal, right) functions across different dimensions.}\label{fig: diversity}
\end{figure}

\subsection{Diversity analysis}
In this section, we analyze the diversity factor introduced in our study. We employ the population diversity index to quantify the variation and spread among individuals within the population. This metric provides insight into how effectively each learning strategy enhances diversity and promotes exploration within the solution space. The mathematical formulation of this index is defined as follows \cite{chauhan2023optimizing}:

\begin{align}\label{eq: diversity}
\begin{array}{ll}
Div_{pop}=\frac{1}{N_s}\sum_{i=1}^{N_s}\sqrt{\left(\sum_{d=1}^D (\overline{\mathbf{x}}_{j}-\mathbf{x}_{i,j}^2\right)},&
\overline{\mathbf{x}}_{j}=\frac{1}{N_s}\sum_{i=1}^{N_s}\mathbf{x}_{i},
\end{array}
\end{align}

where $Div_{pop}$ represents the diversity of the entire population. The term $\mathbf{x}_{i,j}$ denotes the position of the $i$th individual in the $d$th dimension, while $\overline{\mathbf{x}}_{j}$ represents the mean position of all individuals. Diversity plays a crucial role in optimization algorithms as it directly impacts both exploration and convergence behavior.  

Fig. \ref{fig: diversity} presents an analysis of diversity for two optimization problems: the Rosenbrock function (unimodal, left column) and the Rastrigin function (multimodal, right column) across different dimensions ($d = 10, 30, 50, 100$). The graphs illustrate how different learning strategies influence population diversity throughout 1000 iterations.

For the Rosenbrock function (left column), diversity generally decreases as iterations progress, indicating a gradual convergence of the search process—an expected behavior for a unimodal function. However, different learning strategies exhibit varying rates of diversity reduction. Some strategies, such as DLS, CL, and SDL, maintain higher diversity for a prolonged period, suggesting enhanced exploration capability. In contrast, strategies like SL, DMS, AL, and LIS demonstrate a rapid decline in diversity, implying faster convergence but a higher likelihood of premature stagnation.

In contrast, the Rastrigin function (right column) exhibits a significantly different diversity pattern due to its multimodal nature. Instead of a smooth decline, diversity fluctuates more prominently, particularly for strategies such as SDL, DLS, CL, UL, FIP, and AL, which exhibit sharp peaks and drops. This behavior suggests that these strategies frequently escape local minima and redistribute solutions across the search space. Unlike the Rosenbrock function, where a steady decline in diversity indicates convergence, maintaining a moderate level of diversity in Rastrigin is essential to avoid premature convergence to suboptimal solutions.

As the dimensionality increases ($d = 10, 30, 50, 100$), notable trends in diversity levels emerge. In the Rosenbrock function, higher dimensions generally lead to a slower decline in diversity, implying that larger search spaces necessitate extended exploration phases. This effect is particularly evident at $d = 100$, where strategies such as DLS, SDL, CL, AL, EL, and FIP maintain significant diversity even in the later iterations.

For the Rastrigin function, the impact of dimensionality is even more pronounced. The fluctuations in diversity become increasingly irregular as the number of dimensions grows, suggesting that higher-dimensional multimodal landscapes introduce greater challenges in maintaining a balance between exploration and exploitation. Strategies such as AL and SDL exhibit larger oscillations, while others, such as DNL, MAL, DMS, and MFL, demonstrate a more stable diversity decline.

Overall, the results indicate that optimal diversity management is highly problem-dependent. While lower diversity is beneficial for unimodal functions (e.g., Rosenbrock), maintaining a certain level of diversity is crucial for multimodal functions (e.g., Rastrigin) to prevent premature convergence and improve solution quality. Strategies that perform well in unimodal landscapes may not be ideal for multimodal problems, and vice versa.

\begin{figure}
	\centering
	 \begin{subfigure}[b]{0.49\linewidth}
	\includegraphics[width=1\linewidth]{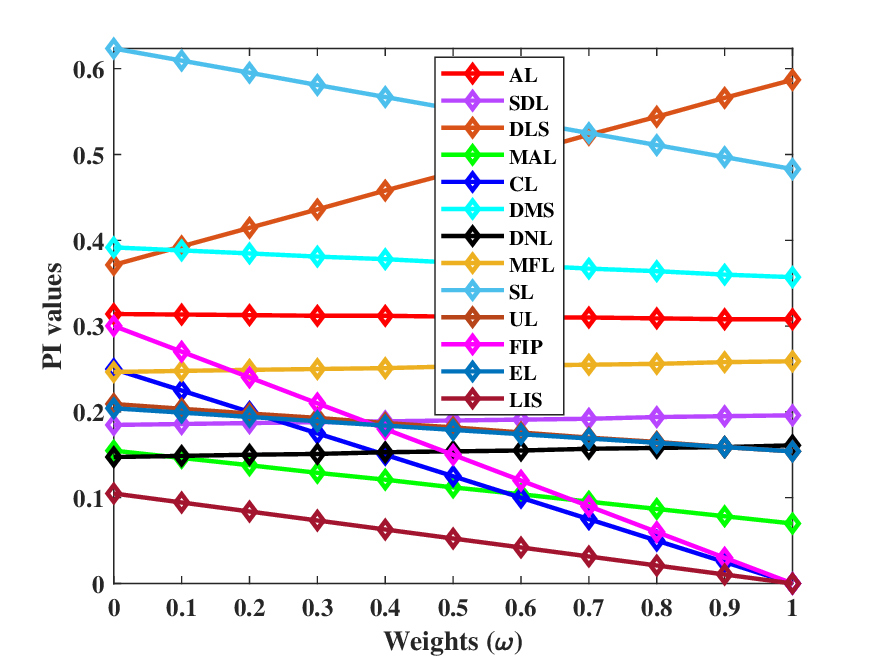}
        \caption{In this case, $k_1=\omega$, $k_2=k_3=(1-\omega)/2$.}\label{subfig: case1}
        \end{subfigure}
        \begin{subfigure}[b]{0.49\linewidth}	
        \includegraphics[width=1\linewidth]{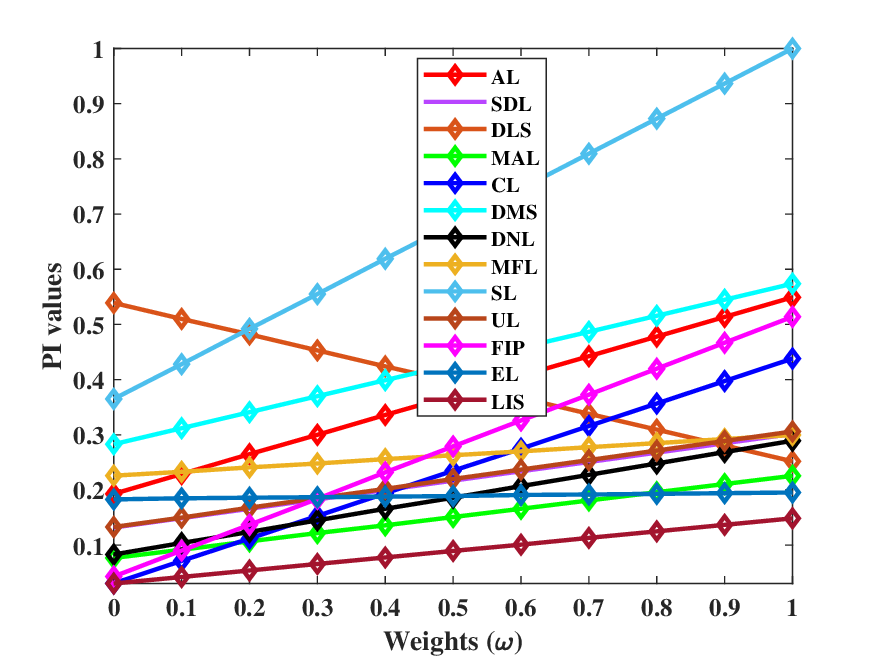}
        \caption{In this case, $k_2=\omega$, $k_1=k_3=(1-\omega)/2$.}\label{subfig: case2}
        \end{subfigure}
        \begin{subfigure}[b]{1\linewidth}\centering
	\includegraphics[width=0.49\linewidth]{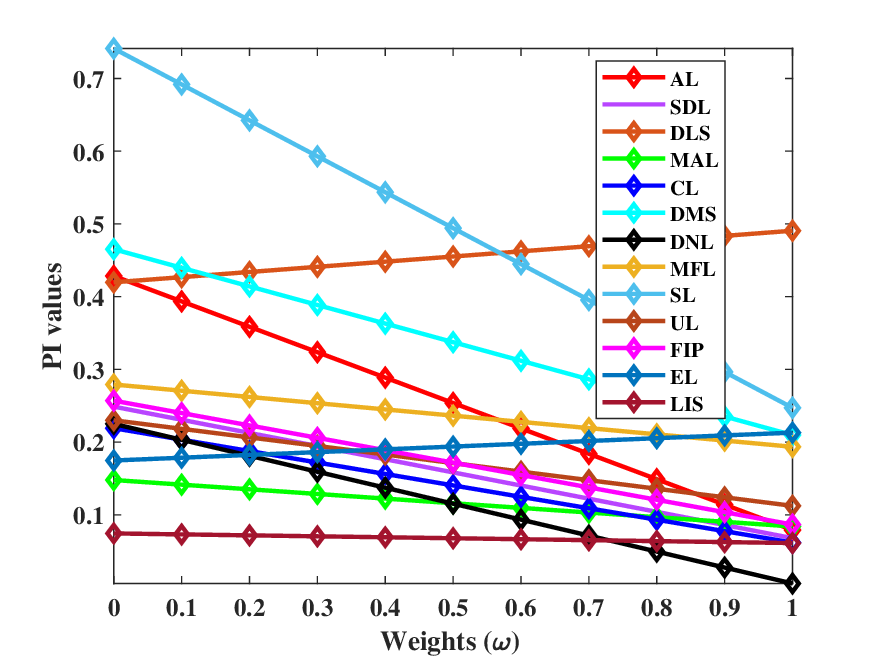}
        \caption{In this case, $k_1=k_2=(1-\omega)/2$, $k_3=\omega$.}\label{subfig: case3}
	 \end{subfigure}
	\caption{Performance index of all learning strategies on basic test functions at 50 dimensions.}\label{fig: PI}
\end{figure}

\subsection{Performance index analysis}
In this section, we evaluate the optimization ability of each learning strategy based on the performance index (PI), which is defined as follows \cite{deep2007new,chauhan2024comprehensive}:

\begin{align}
 & PI=\frac{1}{{N_p}}\sum_{i=1}^{N_p}(k_1\cdot\alpha_{1,i}+k_2\cdot \alpha_{2,i}+k_3\cdot \alpha_{3,i})  ,\\&
 \alpha_{1,i}=\frac{SR_i}{TR},~\alpha_{2,i}=\frac{MT_i}{AT_i},~\alpha_{3,i}=\frac{MF_i}{AF_i},
\end{align}

where $SR_i$ and $TR$ represent the number of successful runs for problem $i$ and the total number of runs, respectively. $MT_i$ and $AT_i$ denote the minimum average time of all strategies and the average time of the respective strategy for problem $i$. Similarly, $MF_i$ and $AF_i$ are the minimum average fitness value and the average fitness value of the strategy for the same problem. The total number of problems, $N_p$, is set to 13 basic test functions. In all three cases, the weight $\omega$ varies between 0 and 1 with a step of 0.1.

Fig. \ref{fig: PI} illustrates the PI of different learning strategies across various weight values (\(\omega\)) in a 50-dimensional search space. Each subplot corresponds to a different weighting scheme, highlighting how different strategies perform under varying priority assignments for success rate (\(SR\)), execution time (\(MT\)), and fitness value (\(MF\))

In subplot \ref{subfig: case1}, where success rate (\(SR\)) is emphasized (\(k_1 = \omega\)) while execution time and fitness value are given equal weight (\(k_2 = k_3 = (1 - \omega)/2\)), DLS, SL, AL, and DMS consistently achieve the highest PI values, indicating their strong ability to balance success rate while maintaining reasonable computational efficiency and solution quality. In contrast, LIS, FIP, and CL perform the worst, showing a decline in PI values as \(\omega\) increases. The other strategies show a moderate change in PT values during $\omega$ increases. 

In subplot \ref{subfig: case2}, where execution time (\(MT\)) is prioritized (\(k_2 = \omega\)), the SL, DMS, DLS, AL, FIP, and CL strategies again perform well, while strategies like UL, LIS, DNL, and MAL exhibit a rising trend, suggesting that these methods are more efficient in reducing execution time. However, EL and UL continue to show poor performance across all weight values, highlighting their inefficiency.  

For subplot \ref{subfig: case3}, where fitness value (\(MF\)) is given the highest priority (\(k_3 = \omega\)), SL and DLS continue to outperform other strategies, maintaining high PI values even as \(\omega\) increases. MFL, UL, and EL demonstrate stable trends, while FIP, SDL, CL, MAL, and DMS show more pronounced declines. Strategies like DNL remain the least effective across all weight conditions.

In general, SL and DLS exhibit superior performance across different weight values, maintaining consistently high PI values. SDL and MAL also demonstrate competitive performance, although their effectiveness varies depending on the test function. Conversely, FIP and LIS consistently report the lowest PI values, indicating limited search efficiency.

As weight values ($\omega$) increase, certain strategies like DMS and MFL exhibit relatively stable performance, whereas UL and FIP show noticeable fluctuations. The divergence in trends across different test cases suggests that some strategies are more sensitive to weight variations, adjusting differently based on function characteristics. Overall, the results emphasize that strategy effectiveness is function-dependent, and an appropriate balance between exploration and exploitation is essential for achieving optimal search performance.

\subsection{Friedman ranks analysis}
The tables present Friedman rank test results on the classical benchmark problems (Table \ref{tab: ranks classical}), the CEC 2017 test suite (Table \ref{tab: ranks CEC 2017}), and the CEC 2020 test suite (Table \ref{tab: ranks CEC 2020}), comparing various learning strategies across different problem dimensions. The Friedman rank test provides a ranking for each algorithm, with lower ranks indicating better performance.

In Table \ref{tab: ranks classical}, the Friedman ranks for the classical benchmark problems show that DNL consistently achieves the lowest rank across all dimensions, indicating its superior performance compared to other learning strategies. For instance, at $d=10$, DNL has a rank of 5.3846, which is the lowest among all methods, while at $d=100$, its rank further improves to 2.1538, reinforcing its robustness in higher-dimensional spaces. Other strategies like SL and DLS also show competitive performance, with relatively lower ranks compared to alternatives such as AL, LIS, and FIP, which tend to have higher ranks, suggesting weaker performance.

In Tables \ref{tab: ranks CEC 2017} and \ref{tab: ranks CEC 2020}, which present the Friedman ranks on the CEC 2017 and CEC 2020 test suites, the results show more variability. DLS emerges as the best-performing strategy, obtaining the lowest rank across all dimensions. For instance, at $d=100$, DLS has a rank of 3.395, which is significantly lower than other strategies. DNL also maintains strong performance, with competitive ranks at lower dimensions, though not as dominant as in Table \ref{tab: ranks classical}. On the other hand, strategies like AL, SDL, SL, and FIP tend to have higher ranks, suggesting that they are less effective compared to the top-performing methods.

In summary, these results indicate that DNL performs exceptionally well on classical benchmark problems, while DLS dominates in the CEC 2017 and CEC 2020 test suites. This highlights the importance of benchmarking different learning strategies across diverse test suites, as the best-performing method may vary depending on the problem set. Additionally, as the problem dimension increases, some strategies like DNL and DLS show greater robustness, reinforcing their effectiveness in high-dimensional multimodal optimization tasks.

\begin{table}[]
    \centering
    \caption{Friedman ranks on the classical benchmark problems.}\label{tab: ranks classical}
\resizebox{1\linewidth}{!}{\begin{tabular}{ccccccccccccccc}\hline
Dimension	&	AL	&	SDL	&		DLS	&	MAL	&	CL	&	DMS	&		DNL	&	MFL	&	SL	&	UL	&	FIP	&	EL	&	LIS	\\\hline
$d=10$	&	6.6608	&	9.1958	&		5.5979	&	6.7378	&	5.5000	&	8.4161	&	\cellcolor{gray}	5.3846	&	6.6608	&	5.5839	&	6.7832	&	10.3776	&	6.1434	&	7.9580	\\\hline
$d=30$	&	7.4965	&	9.5664	&		5.2867	&	8.2972	&	4.7832	&	7.3916	&	\cellcolor{gray}	3.4126	&	7.4965	&	4.3986	&	6.6923	&	11.4476	&	5.6923	&	9.0385	\\\hline
$d=50$	&	8.1888	&	9.4336	&		4.6748	&	8.5420	&	5.0350	&	6.0874	&	\cellcolor{gray}	2.6224	&	8.1888	&	3.6084	&	6.4580	&	11.7832	&	6.0490	&	10.3287	\\\hline
$d=100$	&	9.0035	&	9.0140	&		5.0734	&	8.3147	&	4.7552	&	5.5280	&	\cellcolor{gray}	2.1538	&	9.0035	&	4.1399	&	6.2028	&	11.9301	&	5.6014	&	10.2797	\\\hline
  \end{tabular}}
\end{table}

\begin{table}[]
    \centering
    \caption{Friedman ranks on the CEC 2017 test suite.}
    \label{tab: ranks CEC 2017}
\resizebox{1\linewidth}{!}{\begin{tabular}{ccccccccccccccc}\hline
Dimension	&	AL	&	SDL	&		DLS	&	MAL	&	CL	&	DMS	&		DNL	&	MFL	&	SL	&	UL	&	FIP	&	EL	&	LIS	\\\hline
$d=10$	&	8.1599	&	6.5361	&	\cellcolor{gray}	4.0392	&	8.4436	&	7.4216	&	6.0251	&		4.5925	&	8.1599	&	7.3777	&	7.1489	&	9.5564	&	5.6803	&	7.8589	\\\hline
$d=30$	&	7.6473	&	10.9248	&	\cellcolor{gray}	3.3433	&	8.0219	&	5.5862	&	5.3950	&		6.1113	&	7.6473	&	4.5423	&	7.3730	&	10.6897	&	6.4357	&	7.2821	\\\hline
$d=50$	&	8.3539	&	11.3117	&	\cellcolor{gray}	3.2532	&	7.9740	&	5.3539	&	4.7013	&		5.5097	&	8.3539	&	3.9903	&	7.4286	&	10.9123	&	5.6883	&	8.1688	\\\hline
$d=100$	&	8.8041	&	11.2320	&	\cellcolor{gray}	3.3950	&	7.4671	&	4.8025	&	4.8119	&		4.7586	&	8.8041	&	4.9028	&	6.7994	&	10.8809	&	5.2382	&	9.1034	\\\hline
  \end{tabular}}
\end{table}

\begin{table}[]
    \centering
    \caption{Friedman ranks on the CEC 2020 test suite.}
    \label{tab: ranks CEC 2020}
\resizebox{1\linewidth}{!}{\begin{tabular}{ccccccccccccccc}\hline
Dimension	&	AL	&	SDL	&		DLS	&	MAL	&	CL	&	DMS	&		DNL	&	MFL	&	SL	&	UL	&	FIP	&	EL	&	LIS	\\\hline
$d=5$	&	8.0955	&	5.4000	&	\cellcolor{gray}	4.0682	&	8.7455	&	8.1500	&	5.9136	&		4.6273	&	8.0955	&	9.7227	&	5.9500	&	7.9773	&	5.6227	&	8.6318	\\\hline
$d=10$	&	7.2682	&	6.8364	&	\cellcolor{gray}	4.4273	&	9.5727	&	7.1818	&	6.9500	&		4.4773	&	7.2682	&	8.1773	&	7.0545	&	9.7818	&	6.0182	&	5.9864	\\\hline
$d=15$	&	7.2909	&	8.1455	&	\cellcolor{gray}	3.7545	&	9.0000	&	7.0136	&	7.2636	&		4.7455	&	7.2909	&	6.0773	&	7.8773	&	10.6727	&	5.6955	&	6.1727	\\\hline
$d=20$	&	8.1000	&	8.6182	&	\cellcolor{gray}	3.9409	&	8.5000	&	6.4409	&	6.1091	&		4.5136	&	8.1000	&	5.5773	&	7.8864	&	10.3091	&	6.4955	&	6.4091	\\\hline
  \end{tabular}}
\end{table}

\subsection{Statistical results analysis}
In this section, we statistically compare each learning strategy with the best-performing strategy in the Friedman rank test. Tables \ref{tab: statistical classical} to \ref{tab: statistical CEC 2020} present Wilcoxon Signed-Rank statistical test results comparing different learning strategies across three benchmark sets: classical benchmark problems, the CEC 2017 test suite, and the CEC 2020 test suite. Each table lists results for multiple learning strategies, with performance summarized in terms of wins, draws, and losses across different problem dimensions ($d$). These results provide insights into how various learning strategies perform in different optimization settings.

In Table \ref{tab: statistical classical}, the results on classical benchmark problems compare differential learning strategies against the DNL approach. Across different dimensions ($d=10, 30, 50, 100$), certain strategies such as DLS, DMS, EL, MAL, and UL appear to have strong performance in comparison to DNL. 

Table \ref{tab: statistical CEC 2017} presents statistical comparisons for the CEC 2017 test suite, where the SDL approach is tested against other learning strategies. The results suggest that DLS, DNL, FIS, LIS, MFL, and UL exhibit strong performance, particularly in lower dimensions. As problem complexity increases ($d=30, 50, 100$), strategies like DLS and DMS maintain high win counts, indicating robustness across different problem scales. 

In Table \ref{tab: statistical CEC 2020}, results from the CEC 2020 test suite reveal similar trends. For lower dimensions ($d=5, 10$), SDL performs better than other strategies. However, as the problem dimension increases ($d=15, 20$), the number of draws rises, reinforcing the trend that learning strategies become harder to distinguish in higher-dimensional multimodal optimization tasks. Certain strategies, like UL and SL, appear to perform less favorably in comparison.

Overall, the results indicate that some learning strategies consistently perform well across different benchmark suites. The increasing number of draws in higher dimensions suggests that performance differences between methods become less significant as complexity grows. Additionally, the variation in rankings across different test suites highlights the importance of benchmarking across multiple problem sets to ensure a comprehensive evaluation of learning strategies in differential evolution.

\begin{table}[]
    \centering
    \caption{Statistical test results on the classical benchmark problems, DNL $vs.$}
    \label{tab: statistical classical}
\resizebox{1\linewidth}{!}{ \begin{tabular}{ccccccccccccc}\hline
Learning	&	AL					&	CL					&	DLS					&	DMS					&	EL					&	FIS					&	LIS					&	MAL					&	MFL					&	SDL					&	SL					&	UL					\\\hline
$d=10$	&	4	/	4	/	5	&	6	/	4	/	3	&	2	/	6	/	5	&	1	/	7	/	5	&	1	/	9	/	3	&	7	/	5	/	1	&	1	/	10	/	2	&	3	/	8	/	2	&	1	/	9	/	3	&	1	/	8	/	4	&	1	/	8	/	4	&	0	/	9	/	4	\\\hline
$d=30$	&	3	/	5	/	5	&	4	/	5	/	4	&	1	/	2	/	10	&	1	/	1	/	11	&	1	/	4	/	8	&	5	/	7	/	1	&	2	/	7	/	4	&	1	/	7	/	5	&	1	/	1	/	11	&	1	/	2	/	10	&	1	/	2	/	10	&	1	/	5	/	7	\\\hline
$d=50$	&	4	/	5	/	4	&	3	/	6	/	4	&	1	/	1	/	11	&	1	/	0	/	12	&	1	/	3	/	9	&	6	/	6	/	1	&	3	/	8	/	2	&	1	/	4	/	8	&	1	/	1	/	11	&	1	/	2	/	10	&	1	/	1	/	11	&	1	/	4	/	8	\\\hline
$d=100$	&	8	/	1	/	4	&	4	/	8	/	1	&	1	/	0	/	12	&	1	/	0	/	12	&	1	/	2	/	10	&	12	/	1	/	0	&	8	/	5	/	0	&	1	/	0	/	12	&	1	/	0	/	12	&	1	/	1	/	11	&	3	/	0	/	10	&	2	/	1	/	10	\\\hline
 \end{tabular}}
\end{table}

\begin{table}[]
    \centering
    \caption{Statistical test results on the CEC 2017 test suite, SDL $vs.$}
    \label{tab: statistical CEC 2017}
\resizebox{1\linewidth}{!}{ \begin{tabular}{ccccccccccccc}\hline
Learning	&	AL					&	CL					&	DLS					&	DMS					&	DNL					&	EL					&	FIS					&	LIS					&	MAL					&	MFL					&	SL					&	UL					\\\hline
$d=10$	&	20	/	8	/	1	&	11	/	18	/	0	&	14	/	12	/	3	&	6	/	14	/	9	&	18	/	11	/	0	&	7	/	20	/	2	&	23	/	4	/	2	&	20	/	7	/	2	&	10	/	14	/	5	&	19	/	8	/	2	&	5	/	19	/	5	&	15	/	9	/	5	\\\hline
$d=30$	&	12	/	12	/	5	&	27	/	2	/	0	&	4	/	15	/	10	&	2	/	10	/	17	&	11	/	17	/	1	&	4	/	23	/	2	&	22	/	6	/	1	&	10	/	13	/	6	&	5	/	15	/	9	&	6	/	15	/	8	&	1	/	20	/	8	&	10	/	15	/	4	\\\hline
$d=50$	&	12	/	15	/	2	&	29	/	0	/	0	&	6	/	8	/	15	&	3	/	10	/	16	&	14	/	13	/	2	&	5	/	20	/	4	&	26	/	2	/	1	&	16	/	13	/	0	&	6	/	12	/	11	&	6	/	16	/	7	&	3	/	15	/	11	&	15	/	11	/	3	\\\hline
$d=100$	&	19	/	10	/	0	&	29	/	0	/	0	&	11	/	7	/	11	&	3	/	11	/	15	&	15	/	13	/	1	&	9	/	17	/	3	&	25	/	1	/	3	&	21	/	8	/	0	&	9	/	9	/	11	&	10	/	12	/	7	&	7	/	13	/	9	&	17	/	7	/	5	\\\hline 
  \end{tabular}}
\end{table}

\begin{table}[]
    \centering
    \caption{Statistical test results on the CEC 2020 test suite, SDL $vs.$}
    \label{tab: statistical CEC 2020}
\resizebox{1\linewidth}{!}{ \begin{tabular}{ccccccccccccc}\hline
Learning	&	AL					&	CL					&	DLS					&	DMS					&	DNL					&	EL					&	FIS					&	LIS					&	MAL					&	MFL					&	SL					&	UL					\\\hline
$d=5$	&	6	/	3	/	1	&	2	/	7	/	1	&	8	/	2	/	0	&	2	/	5	/	3	&	7	/	3	/	0	&	2	/	7	/	1	&	9	/	1	/	0	&	6	/	4	/	0	&	4	/	4	/	2	&	7	/	3	/	0	&	3	/	6	/	1	&	4	/	6	/	0	\\\hline
$d=10$	&	4	/	6	/	0	&	3	/	7	/	0	&	7	/	3	/	0	&	1	/	7	/	2	&	8	/	2	/	0	&	3	/	7	/	0	&	8	/	2	/	0	&	5	/	4	/	1	&	4	/	6	/	0	&	6	/	4	/	0	&	1	/	9	/	0	&	4	/	6	/	0	\\\hline
$d=15$	&	5	/	4	/	1	&	7	/	3	/	0	&	3	/	6	/	1	&	2	/	5	/	3	&	8	/	2	/	0	&	1	/	8	/	1	&	8	/	2	/	0	&	4	/	6	/	0	&	6	/	3	/	1	&	2	/	8	/	0	&	1	/	8	/	1	&	4	/	6	/	0	\\\hline
$d=20$	&	8	/	0	/	2	&	8	/	2	/	0	&	4	/	6	/	0	&	3	/	4	/	3	&	6	/	4	/	0	&	4	/	6	/	0	&	8	/	1	/	1	&	3	/	7	/	0	&	3	/	7	/	0	&	4	/	6	/	0	&	1	/	9	/	0	&	5	/	4	/	1	\\\hline
  \end{tabular}}
\end{table}

\begin{figure}
    \centering
    \includegraphics[width=0.8\linewidth,height=8cm]{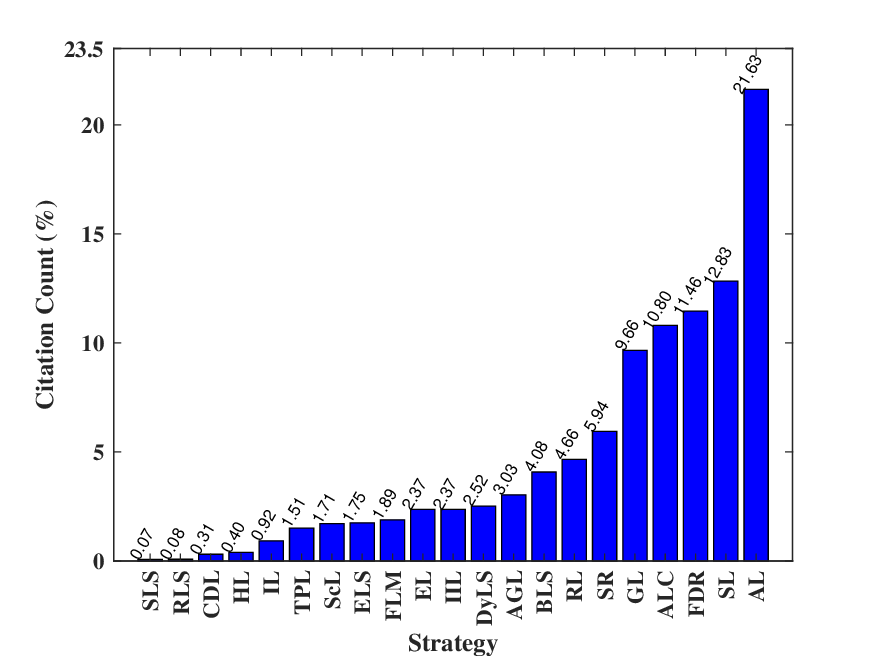}
    \caption{Citations in percentage of learning strategies.}
    \label{fig: citation others}
\end{figure}
\section{Research challenges and open questions}\label{sec: future directions}

Despite the significant advancements and widespread adoption of learning strategies across various domains, several challenges and open research questions remain. Over the past few years, these strategies have attracted growing interest, leading to notable progress in their applications. However, ongoing developments continue to reveal unresolved issues that require further exploration. Addressing these challenges demands sustained efforts from researchers to refine existing approaches and introduce innovative techniques. In the next section, we present key questions and attempt to provide possible answers based on our observations and knowledge.

\subsection{Open challenges and discussion}
Despite the significant advancements in PSO learning strategies (See Fig. \ref{fig: citation others}), several challenges remain unresolved, limiting their effectiveness across diverse optimization problems. While numerous strategies have been proposed, their integration, adaptability, and impact on fundamental PSO behavior require deeper investigation. This section highlights key open questions and research challenges that need to be addressed to further enhance the performance and applicability of PSO.

\textbf{Q.1.} \textit{Which individual strategy or combination of strategies can significantly improve PSO performance across different problem domains?}\\
As we discussed in the previous sections, PSO has been enhanced through various learning strategies, each designed to address specific challenges such as premature convergence, lack of diversity, and slow adaptability to complex landscapes. While no single approach is universally optimal across all problem domains, several strategies have proven effective in different scenarios. CL and DNL introduced a mechanism where particles learn from multiple exemplars rather than a single best neighbor, improving diversity and enhancing exploration in multimodal problems. AL dynamically adjusts key parameters based on swarm behavior, allowing the algorithm to balance exploration and exploitation more effectively across different optimization landscapes. Hybrid learning strategies, e.g., Genetic learning, incorporate evolutionary operators like selection, crossover, and mutation to maintain diversity and prevent stagnation, making it particularly useful in large-scale optimization. Additionally, multi-swarm-based strategies, such as cooperative or hierarchical, divide the population into multiple interacting sub-swarms, allowing for better local and global search capabilities. Hybrid approaches that integrate techniques such as OBL, OL, and SL can further enhance performance by guiding particles toward promising regions of the search space while maintaining a level of randomness necessary for exploration. More recent developments, such as heterogeneous and ensemble, leverage multiple search behaviors simultaneously, making them effective in highly complex and dynamic environments. The strategy dynamics and ranking-based strategies can significantly improve PSO performance. These strategies collectively contribute to making PSO more robust, scalable, and adaptable for a wide range of applications, from engineering optimization to machine learning and real-world decision-making problems.

\textbf{Q.2.} \textit{How can these learning strategies be systematically integrated to maximize their effectiveness across various optimization landscapes?}\\
To systematically integrate learning strategies and maximize their effectiveness across diverse optimization landscapes, an adaptive strategy selection framework can be developed. Such a framework would dynamically adjust the employed strategies based on the characteristics of the problem, ensuring an optimal balance between exploration and exploitation. Multi-swarm and hybrid models offer another promising approach, where different subpopulations utilize distinct learning strategies, enhancing search diversity and adaptability. Additionally, machine learning techniques, such as reinforcement learning and surrogate modeling, can be leveraged to automate strategy selection, allowing the algorithm to switch between strategies in real-time based on performance feedback. Furthermore, evolving strategy ensembles, where different strategies compete and adapt based on historical performance, can provide a more flexible and efficient approach to solving complex optimization problems. By integrating these techniques, PSO can be made more robust, scalable, and capable of handling a wide range of optimization challenges.

\textbf{Q.3.} \textit{How do different learning mechanisms impact fundamental PSO behavior, including convergence speed, robustness, and the exploration-exploitation trade-off?}\\
Different learning mechanisms in PSO significantly influence key aspects of its performance, including convergence speed, robustness, and the exploration-exploitation balance. Comprehensive learning strategies, such as CLPSO, enhance diversity by allowing particles to learn from multiple exemplars, making them effective in multimodal landscapes but sometimes leading to slower convergence. Adaptive and self-learning strategies dynamically adjust search behaviors, enabling PSO to respond to changing landscapes and improving robustness in real-world applications. Two-swarm and multi-swarm models, such as HCLPSO and cooperative PSO, further refine diversity control by allowing different subpopulations to specialize in exploration or exploitation, reducing the likelihood of stagnation in local optima. Genetic learning mechanisms, such as those in GL-PSO, introduce evolutionary operators like crossover and mutation to enhance solution diversity and prevent premature convergence. However, these techniques may come at the cost of increased computational complexity. Overall, selecting an appropriate learning mechanism requires careful consideration of the optimization landscape and the trade-offs between convergence speed, robustness, and exploration-exploitation dynamics.

\textbf{Q.4.} \textit{What methodologies should be used to fairly and systematically assess the effectiveness of different PSO learning strategies?}\\
To fairly and systematically assess the effectiveness of different PSO learning strategies, a multi-faceted evaluation framework is necessary. Benchmark testing on a diverse set of optimization problems, including unimodal, multimodal, high-dimensional, dynamic, and constrained problems, provides a comprehensive assessment of each strategy’s strengths and limitations. Statistical analysis, using non-parametric tests such as the Wilcoxon rank-sum test and the Friedman test, ensures that performance differences are statistically significant rather than due to random variations. Ablation studies further contribute to the understanding of PSO variants by isolating the impact of individual learning components in hybrid or adaptive models. Computational efficiency metrics, including convergence speed, scalability, and memory consumption, may be essential to assess the trade-offs between solution quality and computational cost.

In addition to traditional performance evaluations, explainability methods such as \texttt{SHAP} (Shapley Additive Explanations), \texttt{LIME} (Local Interpretable Model-Agnostic Explanations), and \texttt{PDP} (Partial Dependence Plots) can provide insights into the decision-making process of different PSO strategies. \texttt{SHAP} values can help quantify the influence of different learning mechanisms on particle updates, \texttt{LIME} can offer local interpretability of how specific strategies affect convergence, and \texttt{PDP} can visualize the effect of key parameters on PSO behavior. These explainability techniques not only enhance the interpretability of PSO strategies but also aid in identifying potential biases, improving parameter tuning, and guiding future modifications. Finally, real-world applications in engineering design, machine learning, and decision-making provide practical validation, ensuring that PSO strategies are not only effective in theoretical benchmarks but also applicable to real-world optimization challenges.

\textbf{Q.5.} \textit{How can adaptive learning strategies dynamically adjust based on search dynamics to enhance global optimization performance?}\\
Adaptive learning strategies dynamically adjust based on search dynamics to enhance global optimization performance through various mechanisms. One effective approach is feedback-driven parameter adaptation, where methods like Adaptive PSO (APSO) modify key parameters such as inertia weight and learning coefficients in response to real-time population behavior. By continuously analyzing convergence trends, the algorithm can dynamically shift between exploration and exploitation, improving search efficiency and preventing premature convergence.

Another promising technique is evolutionary control mechanisms, where self-adaptive and ensemble-based strategies enable PSO to transition between different learning behaviors during optimization. These strategies adjust based on the complexity of the problem and variations in the search landscape, reducing the risk of stagnation in local optima. By integrating multiple learning mechanisms, PSO can dynamically choose the most effective approach at different stages of the optimization process.

In addition, reinforcement learning and meta-learning provide a sophisticated way to enhance adaptability. Techniques such as deep reinforcement learning (RL) and evolutionary algorithms can optimize PSO’s learning strategy selection in real time. By framing strategy selection as a learning task, these methods allow PSO to continuously refine its search behavior based on past performance, making it more effective across a wide range of problem domains.

Finally, hierarchical and multi-agent learning introduces a structured approach to swarm intelligence. By assigning different roles to particles, where some prioritize exploration while others focus on exploitation, PSO can maintain a dynamic balance throughout the search process. This structured learning mechanism is particularly beneficial in complex, high-dimensional optimization tasks, ensuring both diversity in search exploration and efficiency in convergence.

\textbf{Q.6.} \textit{What are the limitations of existing PSO learning mechanisms, and how can future research address them to develop self-adaptive, intelligent PSO variants?}\\
Existing PSO learning mechanisms face several limitations that hinder their effectiveness in complex optimization scenarios. One of the primary challenges is premature convergence, which is particularly problematic in high-dimensional search spaces \cite{houssein2021major,zhang2022three,xia2018multi}. Many PSO variants struggle to maintain diversity in the swarm, leading to early stagnation at suboptimal solutions \cite{xu2019particle}. To address this, future research could explore adaptive multi-objective learning strategies that dynamically adjust exploration and exploitation pressures based on real-time feedback from the search process.

Another significant limitation is parameter tuning, which remains a bottleneck in PSO's performance. Many existing approaches require extensive manual fine-tuning of inertia weight, learning mechanism, and cognitive and social parameters, making them less adaptable to different problem landscapes \cite{nasir2012dynamic}. Future advancements in self-adaptive and meta-learning-based PSO variants could automate parameter selection, allowing the algorithm to dynamically adjust its behavior based on search dynamics without extensive human intervention.

PSO also struggles with handling dynamic environments, as many real-world optimization problems involve continuously changing landscapes. Traditional PSO variants cannot adapt to shifting optima over time \cite{li2022ranking,liu2022strategy,tao2022fitness}. Learning-driven PSO models that incorporate memory-based or reinforcement learning techniques could be developed to overcome this, enabling the swarm to retain useful knowledge from past iterations and adapt more effectively to dynamic changes.

A crucial concern is the computational cost associated with advanced learning mechanisms. Many sophisticated PSO variants integrate complex learning strategies, increasing the computational burden and reducing efficiency, especially for large-scale problems. Future research could explore efficient surrogate-assisted or distributed PSO models, which leverage machine learning surrogates or parallel computing to reduce computational overhead while maintaining optimization quality. Moreover, one fundamental limitation of PSO variants lies in their lack of crossover and selection operators, which are integral to many EAs such as DE. These additional operators in DE provide greater flexibility and enhance search dynamics by enabling more diverse exploration and refined exploitation. In contrast, standard swarm algorithms rely solely on velocity and position updates, which may limit their ability to escape local optima or adapt to complex problem landscapes. As a result, PSO variants often underperform the best DE-based algorithms in terms of global search power and solution diversity, as can be observed from the CEC competitions. 

Finally, a key limitation is the lack of explainability in PSO learning mechanisms. Many strategies work well empirically, but there is limited understanding of why certain approaches perform better in specific scenarios. Developing interpretable PSO mechanisms with explainable AI (XAI) techniques could provide deeper insights into swarm behavior, guiding the integration of PSO in real-world decision-making processes. Addressing these challenges through adaptive and intelligent learning strategies would be critical for advancing PSO’s capabilities and ensuring its applicability to increasingly complex optimization problems.
\subsection{Potential research directions}
In the continuation of the previous section, to foster future advancements, drawing inspiration from emerging trends and incorporating more effective methodologies would be essential. Based on our literature review, we highlighted key unresolved issues that warrant deeper investigation within the swarm intelligence community. Additionally, we outline potential areas for enhancement, including refining learning strategies, improving methodologies, and expanding their applications.

Future research could prioritize the development of fully adaptive and self-learning mechanisms, enabling PSO to dynamically select and fine-tune learning strategies based on problem complexity. Integrating reinforcement learning, meta-learning, and transfer learning can facilitate real-time adaptability, enhancing PSO's performance across diverse optimization landscapes. Additionally, hybridizing PSO's learning strategies with deep learning, evolutionary algorithms (e.g., DE, GA, ACO), and ensemble-based techniques can improve search efficiency and solution quality. For instance, incorporating EA's crossover and selection operators into PSO, such as in HCLPSO variants, can improve convergence quality while maintaining PSO's natural advantage of faster convergence speed. This hybridization can potentially balance exploration and exploitation more effectively, enabling PSO to perform competitively across a wider range of optimization problems. Developing principled frameworks for such hybrid algorithms, along with theoretical studies and empirical validations, will be a promising direction to enhance the robustness and generalizability of PSO learning strategies.

Machine learning-assisted PSO can predict optimal learning strategies, while hybrid evolutionary frameworks can leverage multiple algorithms to achieve a better exploration-exploitation balance. Enhancing scalability for large-scale, high-dimensional, and multi-objective optimization problems is another critical direction. Furthermore, the development of context-aware, chaos-based, or quantum-inspired learning strategies can improve robustness in dynamic and uncertain environments, ensuring adaptability to real-world applications.

To advance the field, establishing standardized benchmarks for evaluating PSO learning strategies and conducting theoretical convergence analyses will provide deeper insights into their effectiveness. The XAI methods, such as \texttt{SHAP}, \texttt{LIME}, and \texttt{PDP}, can be used to analyze the insights into the decision-making process of different PSO strategies. Finally, customizing learning mechanisms for real-time optimization in domains such as robotics, energy systems, and healthcare can bridge the gap between theoretical advancements and practical implementations.

\section{Conclusion}\label{sec: conclusion}
This review paper comprehensively explored and classified a wide range of learning strategies developed to enhance the performance of particle swarm optimization (PSO). From single-swarm to multi-swarm learning models, including adaptive, hybrid, and cooperative mechanisms, we analyzed how each approach contributes to improved convergence speed, robustness, and the balance between exploration and exploitation. A key contribution of this work lies in the systematic categorization of these strategies, filling a critical gap in the literature where no unified framework previously existed. Furthermore, our comparative analysis offers valuable insights into the effectiveness of various strategies across diverse optimization scenarios. Looking ahead, we emphasize the importance of developing self-adaptive, explainable, and intelligent PSO variants that can dynamically adjust their behavior based on problem characteristics. Such advancements will pave the way for more resilient and scalable PSO models, capable of addressing the growing complexity of real-world optimization problems.

\section*{Acknowledgment}
	\noindent The author is thankful to the National
University of Singapore for the necessary support for this research.

	\section*{Compliance with ethical standards}
	\noindent \textbf{Conflict of interest:} All the authors declare that they have no conflict of interest.
	
	\noindent \textbf{Ethical approval:} This article does not contain any studies with human participants or animals performed by any of the authors.

	\bibliographystyle{unsrt}
	\small
	\bibliography{LS_PSO}
\end{document}